\documentclass[runningheads]{llncs}
\usepackage{graphicx}
\usepackage{subcaption}
\usepackage{tikz}
\usepackage{comment}
\usepackage{amsmath,amssymb} 
\usepackage{color}

\usepackage[width=122mm,left=12mm,paperwidth=146mm,height=193mm,top=12mm,paperheight=217mm]{geometry}

\begin{document}
\pagestyle{headings}
\mainmatter
\def\ECCVSubNumber{100}  

\title{Hey Human, If your Facial Emotions are Uncertain, You Should Use Bayesian Neural Networks!}

\titlerunning{Hey Human, If your Facial Emotions are Uncertain, You Should Use BNNs!} 

%

\author{Maryam Matin\inst{1} \and
    Matias Valdenegro-Toro\inst{2}}
\authorrunning{Matin et al.}
%
\institute{Hochschule Bonn-Rhein-Sieg, 53757 Sankt Augustin, Germany \email{maryam.matin.1987@gmail.com} \and
German Research Center for Artificial Intelligence, 28359 Bremen, Germany \email{matias.valdenegro@dfki.de}}

\maketitle

\begin{abstract}
Facial emotion recognition is the task to classify human emotions in face images. It is a difficult task due to high aleatoric uncertainty and visual ambiguity. A large part of the literature aims to show progress by increasing accuracy on this task, but this ignores the inherent uncertainty and ambiguity in the task. In this paper we show that Bayesian Neural Networks, as approximated using MC-Dropout, MC-DropConnect, or an Ensemble, are able to model the aleatoric uncertainty in facial emotion recognition, and produce output probabilities that are closer to what a human expects. We also show that calibration metrics show strange behaviors for this task, due to the multiple classes that can be considered correct, which motivates future work. We believe our work will motivate other researchers to move away from Classical and into Bayesian Neural Networks.
\end{abstract}

\section{Introduction}

Emotion recognition in facial images is the task of classifying the face of a person into a set of emotions. One important characteristic of this task is its high degree of aleatoric uncertainty \cite{barsoum2016training}, which presents itself as ambiguity in defining what is the correct emotion class given an image \cite{wang2020suppressing} \cite{ko2018brief}. Most state of the art neural networks used for this task do not model any kind of uncertainty, which makes them ill-posed for emotion recognition.

In this paper we evaluate three scalable methods for uncertainty quantification in neural networks, namely Monte Carlo Dropout/DropConnect, and Deep Ensembles, on the FER+ dataset \cite{barsoum2016training} using three different neural network architectures. This dataset is a variation of the FER dataset \cite{Goodfellow-2013} where each image is labeled with a crowd-sourced distribution of emotion classes, instead of a single class annotation per image. Our results show that Bayesian neural networks are better able to model this kind of problem, even as only one label is used during training.

We believe that our results show that metrics for this task need to be rethought, and that only methods able to model at least aleatoric uncertainty should be used for emotion recognition. It makes little sense to obtain high accuracy on this task, given the visual ambiguity and the multiple correct answers that are possible.

\section{Related Work}

There is a rich literature on emotion recognition from facial images \cite{ko2018brief}. The FER+ dataset \cite{barsoum2016training} is one dataset used to evaluate progress in this task. Two defining characteristics of this dataset are being grayscale images at $64 \times 64$ resolution, and labels might indicate multiple emotions, as defined by a crowd-sourced probability distribution. Classes are 'neutral', 'happiness', 'surprise', 'sadness','anger', 'disgust', 'fear', and 'contempt'. The baseline reported in \cite{barsoum2016training} is 84.7\% accuracy with a custom VGG13 network and a standard cross-entropy loss and data augmentation from \cite{yu2015image}.

Georgescu et al. \cite{georgescu2018local} use CNNs with bag of visual words to obtain 87.7\% accuracy.  Other baselines presented in this paper are 84.4\% accuracy with VGG-face \cite{parkhi2015vggface}, a Bag of Visual Words alone obtaining 79.6\%, and a large ensemble achieving 88\% accuracy. Arriaga et al.\cite{arriaga2019real} reports 78\% and 81\% accuracy with a reduced VGG and a mini-Xception network.

Overall most methods for facial emotion recognition use classical neural networks, and Bayesian neural networks are not commonly used, even more recent work that uses ensembles like Siqueira et al. \cite{siqueira2020efficient} or Surace et al. \cite{surace2017emotion} do not consider the possibility of modeling output uncertainty, despite Lakshminarayanan et al. \cite{lakshminarayanan2017simple} showing that ensembles are able to produce state of the art uncertainty quantification.

\section{Experimental Methodology}

For our experiments we use three of the most common CNN model architectures which have shown outstanding performance in image classification competition on ImageNet, namely AlexNet \cite{krizhevsky2012imagenet}, VGG16 \cite{simonyan2014very} and DenseNet121 \cite{huang2017densely} with some minor modifications. To simplify the models, we reduce the number of neurons in fully connected layers to 256 instead of 4096.
For AlexNet, we add batch normalization \cite{Ioffe-2015} after each layer. DenseNet-121 is modified to integrate dropout layers into the architecture.

We use the dropout/drop rate of 0.5 for AlexNet and 0.2 for the other two models as suggested by their original implementations. The batch size is set to 32 and we use categorical cross-entropy loss and accuracy metric with Adam optimizer \cite{kingma2014adam}. Note that most implementations of the cross-entropy loss use only a single label per class, even as more labels might be available, so in this work we do not explore the use of soft labels \cite{peterson2019human}. We decided to only tune the learning rate in range $10^{-1}$ to $10^{-4}$. For VGG16 models we used Stochastic Gradient Descent instead of Adam. For SGD optimizer we tuned the learning rate decay in range $10^{-1}$ to $10^{-6}$. The actual training after hyper-parameter tuning is done over 80 epochs. We do not perform any kind of data augmentation.

Since full inference in a Bayesian neural network is intractable, we use approximate methods. Due to their scalability and simplicity, we use Monte Carlo Dropout \cite{gal2016dropout}, Monte Carlo DropConnect \cite{mobiny2019dropconnect}, and Deep Ensembles \cite{lakshminarayanan2017simple}. These methods all have a hyper-parameter in common, the number of stochastic forward passes $T$ for Monte Carlo methods, and the number of ensembles $N$ for Deep Ensembles. Note that while MC Dropout/DropConnect are approximations to a BNN, Deep Ensembles is a non-Bayesian method but it outperforms other methods in uncertainty quantification and out of distribution detection \cite{ovadia2019can}.

We evaluate three metrics on the FER+ dataset \cite{barsoum2016training}. We compute classification error, Negative Log Likelihood (NLL), and expected calibration error (ECE) \cite{guo2017calibration}, all as a function of number of stochastic forward samples/ensembles, which are varied between 1 to 15. NLL determines whether the network is assigning high confidence to correct classes, and calibration determines if its confidence estimates are compatible with the true likelihood of the data.

\section{Experimental Results and Analysis}

\begin{figure}[!t]
    \centering
    \begin{subfigure}[b]{0.28\linewidth}
        \includegraphics[width=\linewidth]{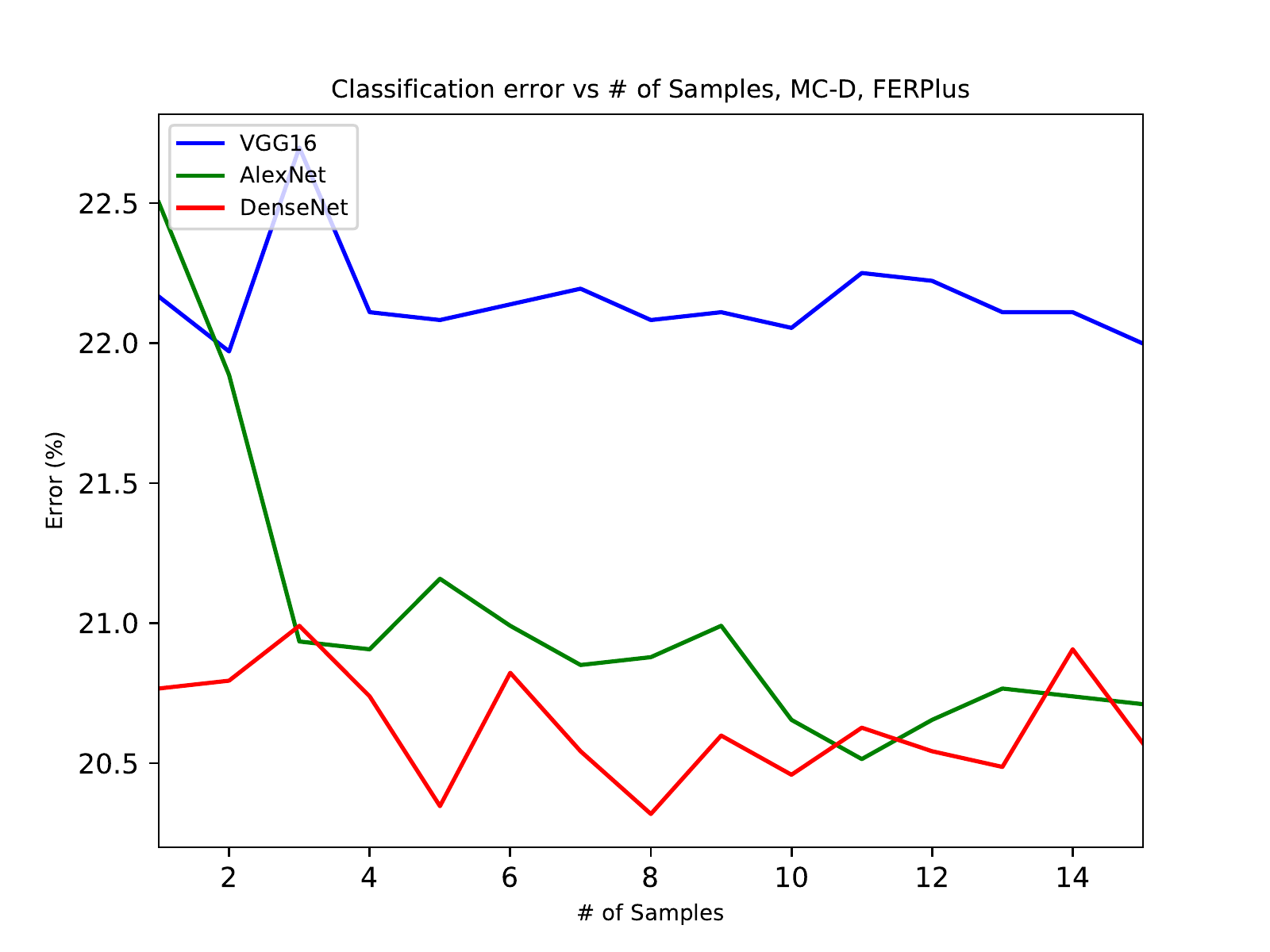}
        \caption{MC-D}
    \end{subfigure}
    \begin{subfigure}[b]{0.28\linewidth}
        \includegraphics[width=\linewidth]{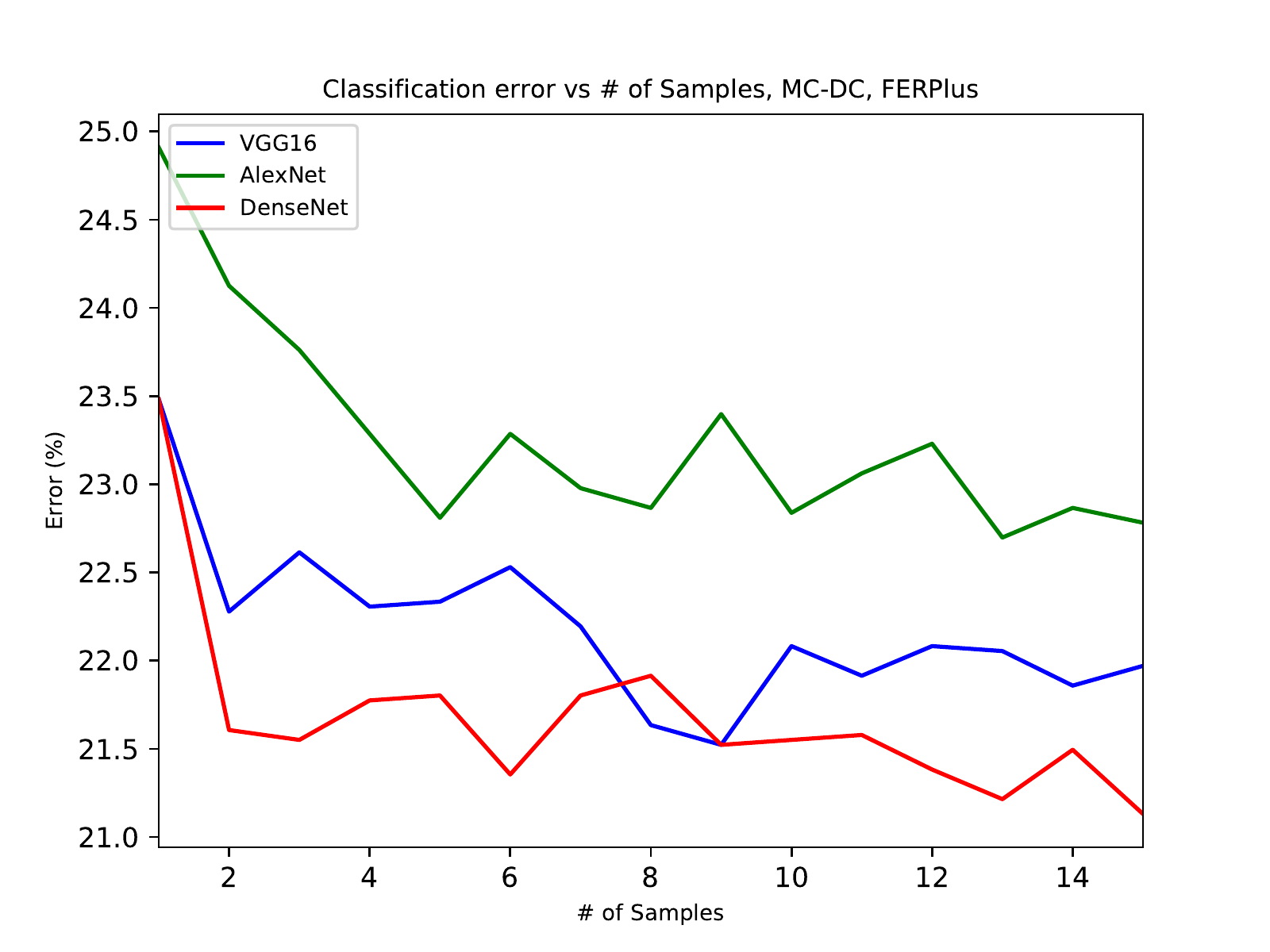}
        \caption{MC-DC}
    \end{subfigure}
    \begin{subfigure}[b]{0.28\linewidth}
        \includegraphics[width=\linewidth]{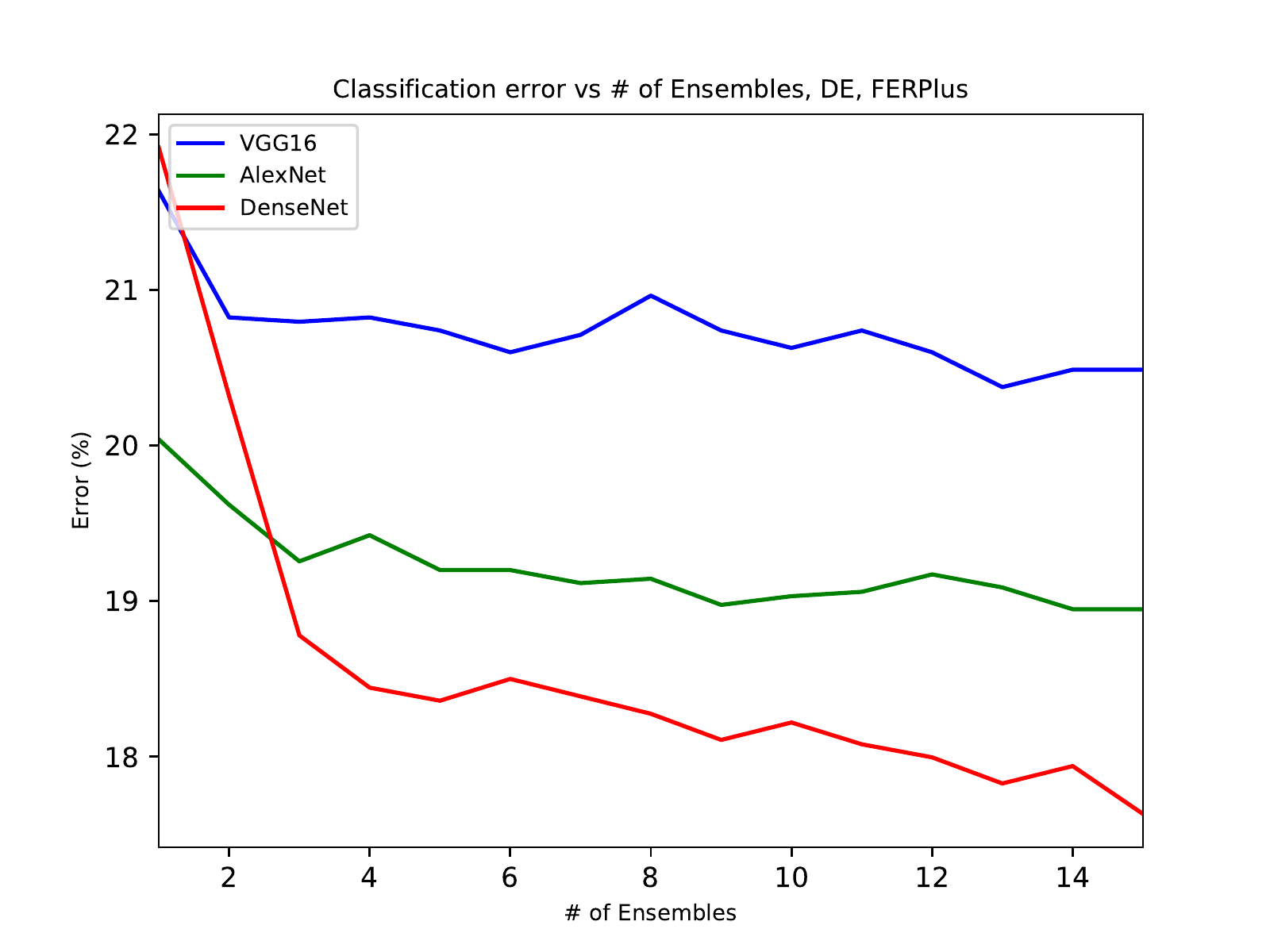}
        \caption{DE}
    \end{subfigure}
    \caption{Classification error as a function of \# of samples/ensembles for all methods in different models on FERPlus dataset.}
    \label{fig:ferplus_classification}
    \vspace{-2em}
\end{figure}

\begin{figure}[!t]
    \centering
    \begin{subfigure}[b]{0.28\linewidth}
        \includegraphics[width=\linewidth]{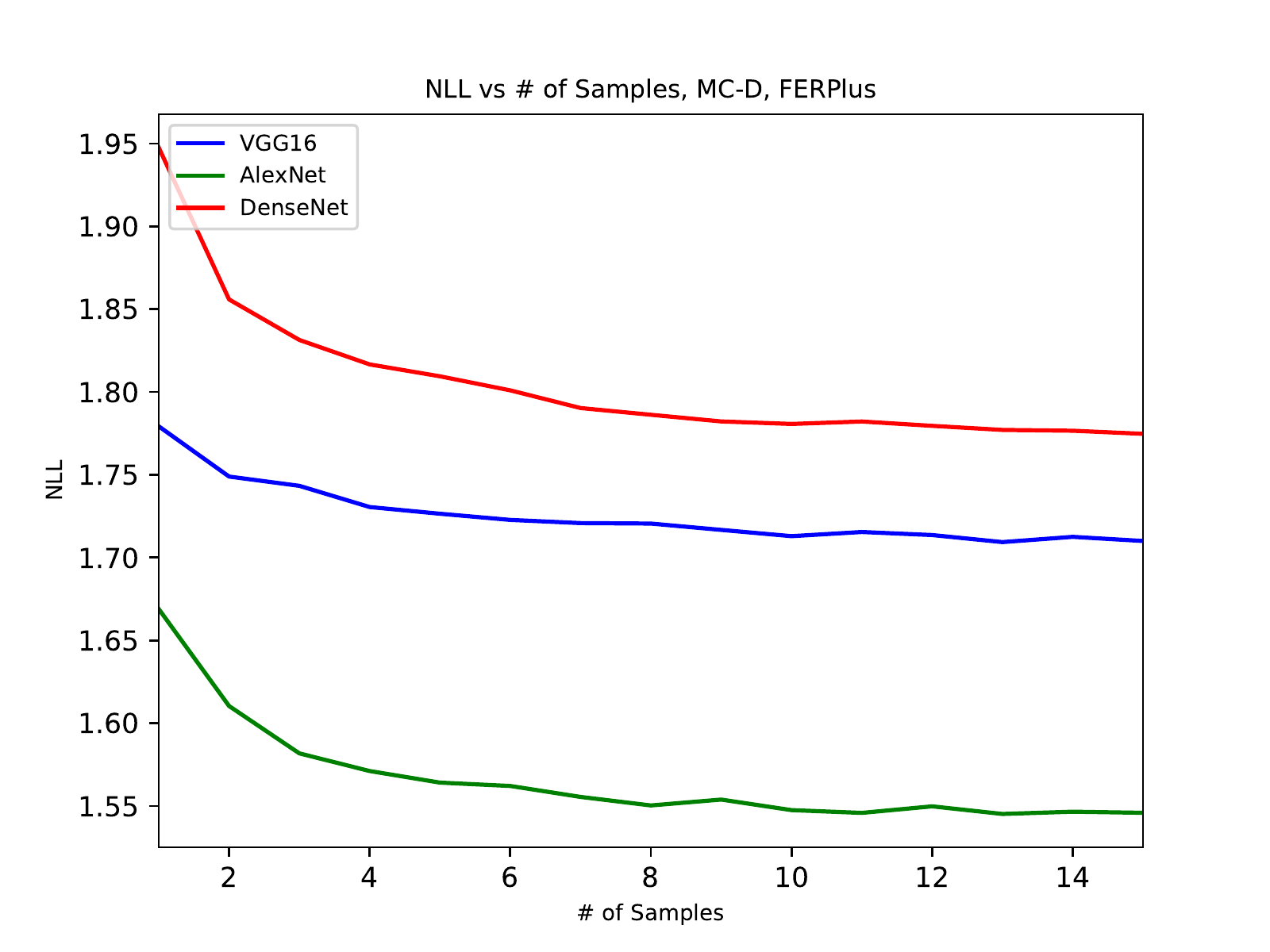}
        \caption{MC-D}
    \end{subfigure}
    \begin{subfigure}[b]{0.28\linewidth}
        \includegraphics[width=\linewidth]{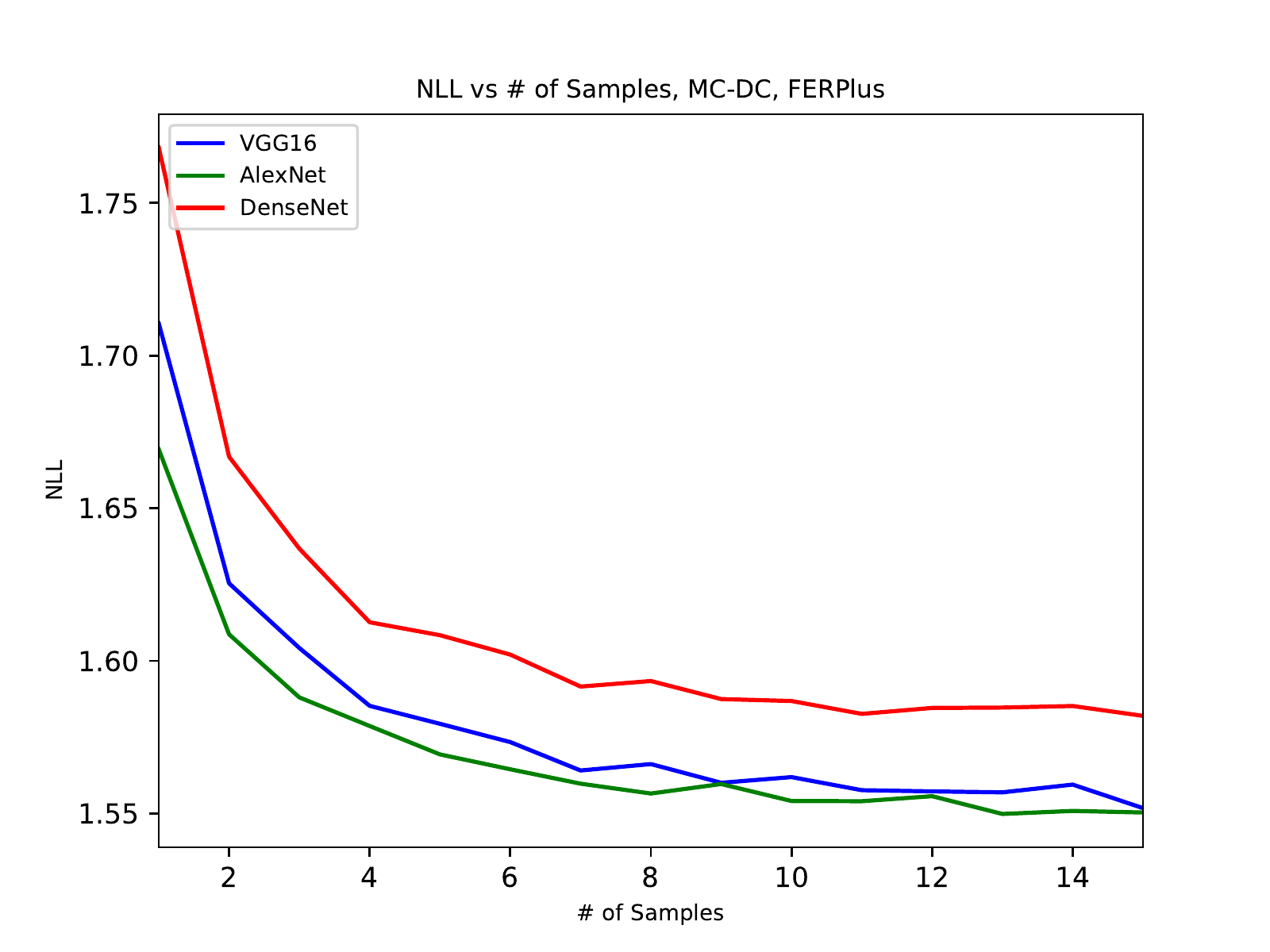}
        \caption{MC-DC}
    \end{subfigure}
    \begin{subfigure}[b]{0.28\linewidth}
        \includegraphics[width=\linewidth]{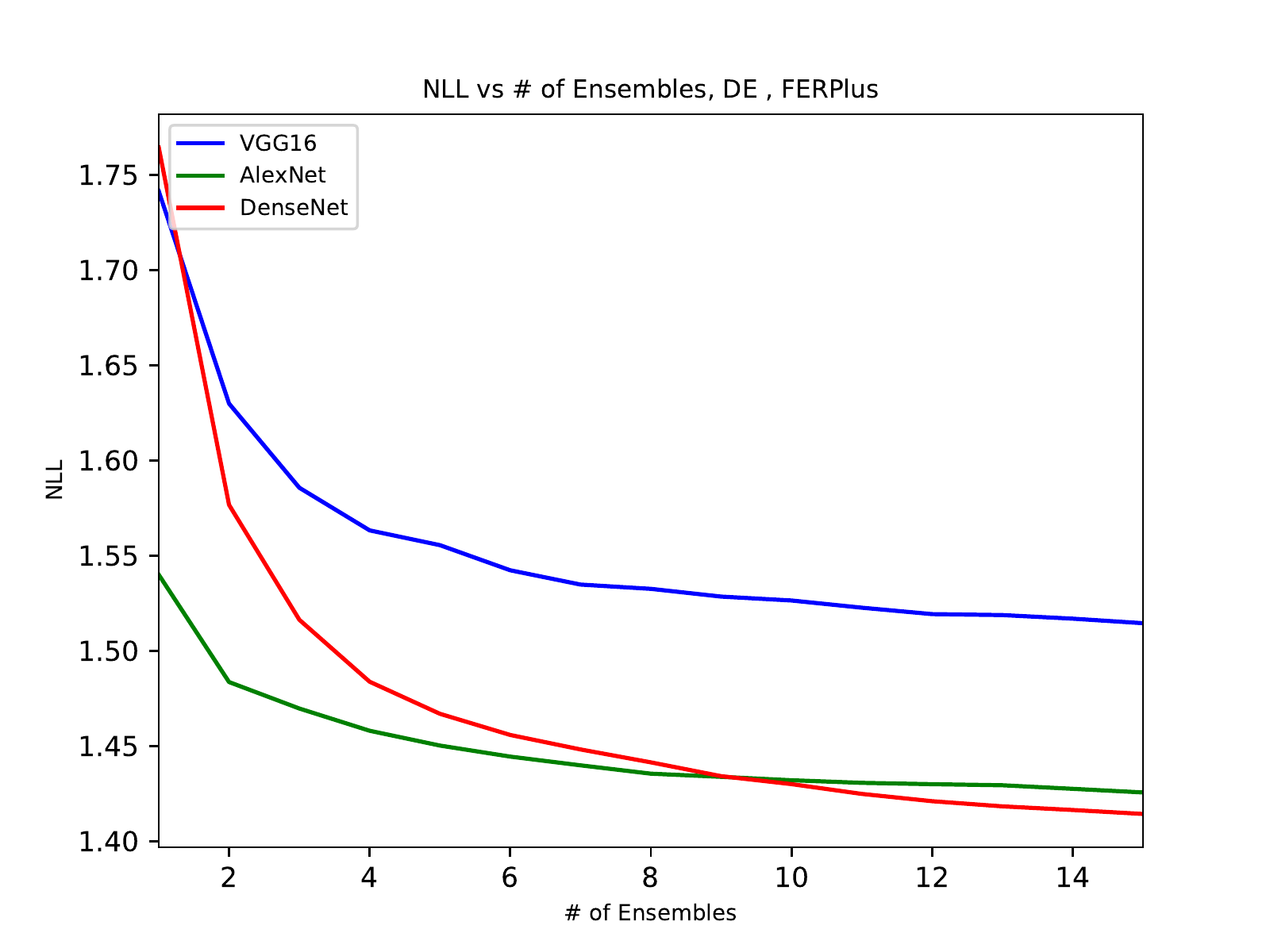}
        \caption{DE}
    \end{subfigure}
    \caption{NLL as a function of \# of samples/ensembles for all methods in different models on FERPlus dataset}
    \label{fig:ferplus_nll}
\end{figure}

\begin{figure}[!t]
    \centering
    \begin{subfigure}[b]{0.28\linewidth}
        \includegraphics[width=\linewidth]{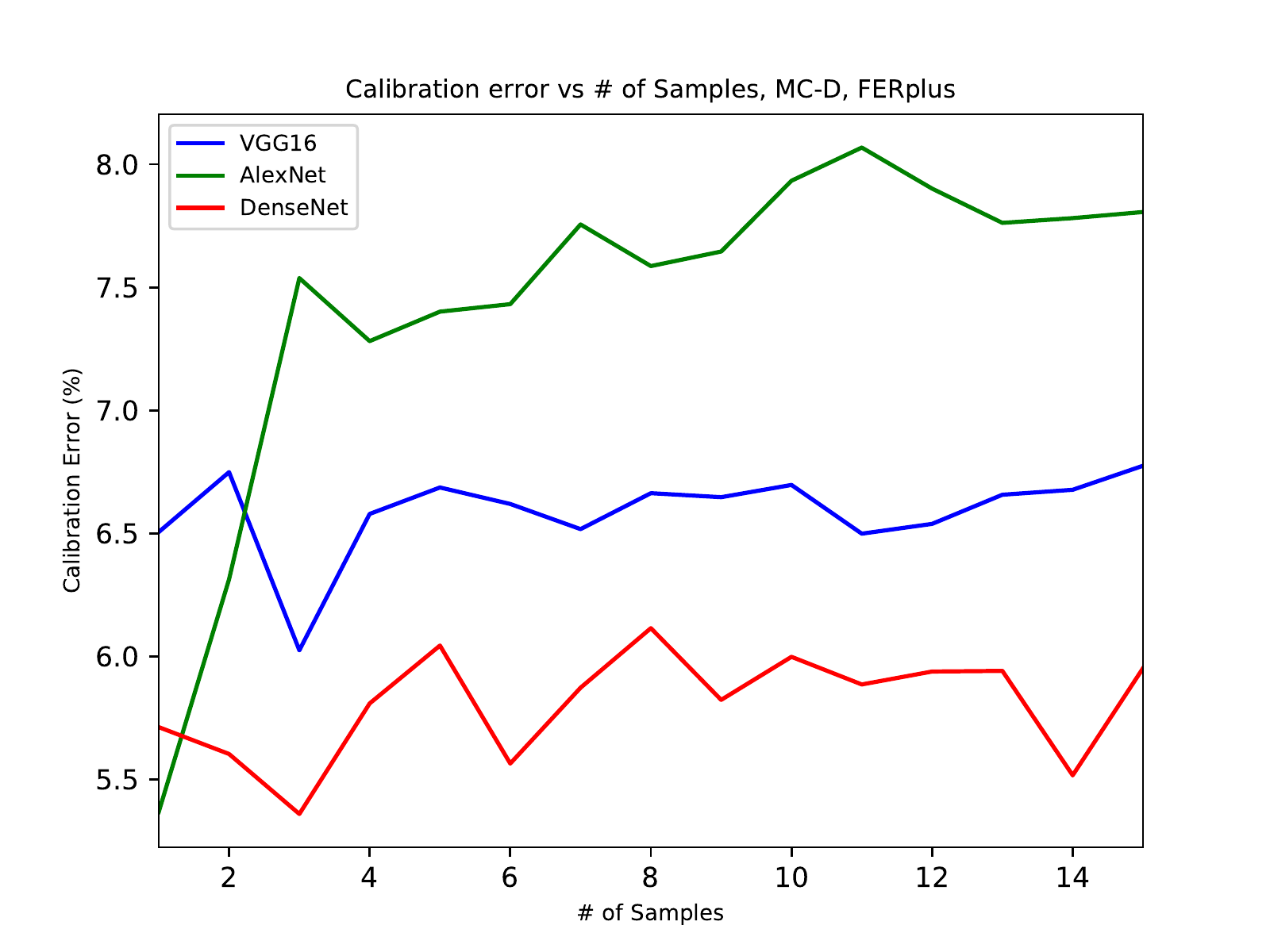}
        \caption{MC-D}
    \end{subfigure}
    \begin{subfigure}[b]{0.28\linewidth}
        \includegraphics[width=\linewidth]{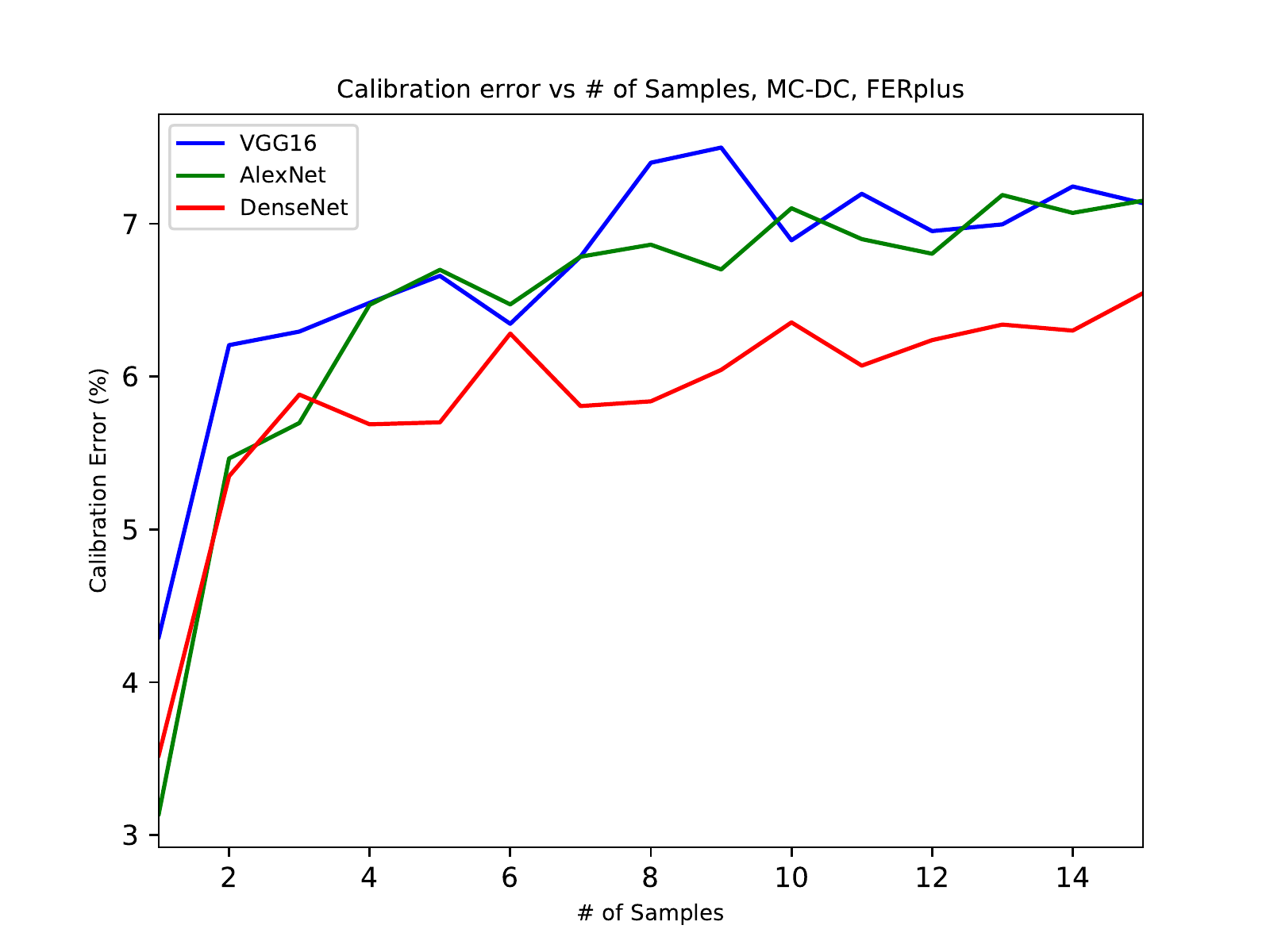}
        \caption{MC-DC}
    \end{subfigure}
    \begin{subfigure}[b]{0.28\linewidth}
        \includegraphics[width=\linewidth]{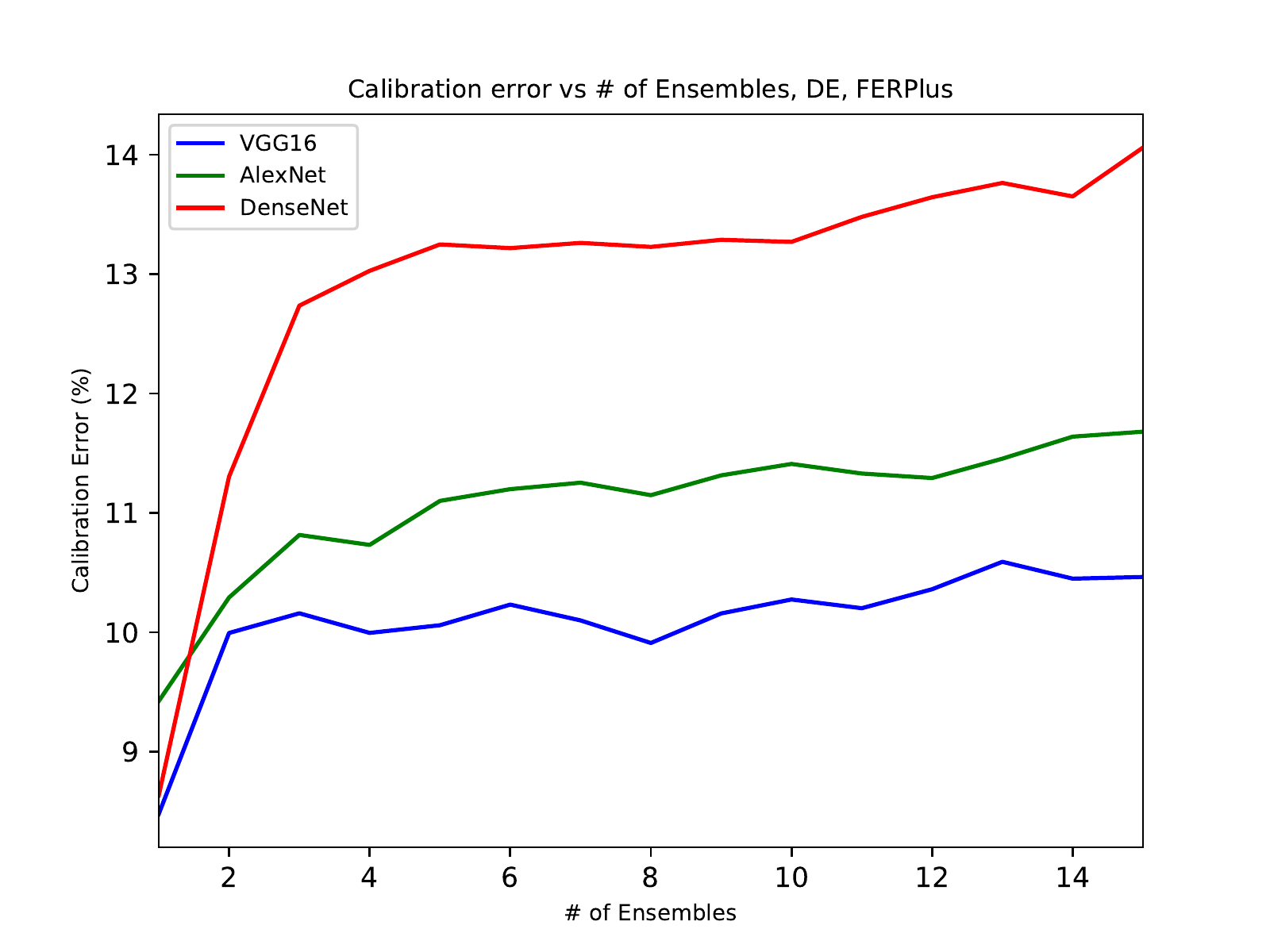}
        \caption{DE}
    \end{subfigure}
    \caption{Calibration error as a function of \# of samples/ensembles for all methods in different models on FERPlus dataset}
    \label{fig:ferplus_calibratione-sample}
    \vspace*{-2em}
\end{figure}

Our main results are presented in Figure \ref{fig:ferplus_classification} for classification error, Figure \ref{fig:ferplus_nll} for the negative log-likelihood, and Figure \ref{fig:ferplus_calibratione-sample} for expected calibration error \cite{guo2017calibration}.

The effect on task performance (accuracy/error) is as expected, with overall decreasing error for all uncertainty methods, but this is more pronounced with an ensemble of DenseNets, which is also confirmed with the plots of negative log-likelihood. There are large variations in performance across different models, and MC-Dropout and MC-DropConnect seem to be less stable than ensembles.

\begin{figure}[h!]
    \centering
    \begin{subfigure}[b]{0.09\linewidth}
        \caption*{\textbf{Image}}
    \end{subfigure}
    \begin{subfigure}[b]{0.16\linewidth}
        \caption*{\textbf{Labels}}
    \end{subfigure}
    \begin{subfigure}[b]{0.16\linewidth}	
        \caption*{\textbf{1 Ens}}
    \end{subfigure}
    \begin{subfigure}[b]{0.16\linewidth}
        \caption*{\textbf{5 Ens}}
    \end{subfigure}
    \begin{subfigure}[b]{0.16\linewidth}
        \caption*{\textbf{10 Ens}}
    \end{subfigure}
    \begin{subfigure}[b]{0.16\linewidth}
        \caption*{\textbf{15 Ens}}
    \end{subfigure}
    \begin{subfigure}[b]{0.09\linewidth}
        \includegraphics[width=\linewidth]{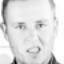}
        \caption*{Contempt}
    \end{subfigure}
    \begin{subfigure}[b]{0.17\linewidth}
        \includegraphics[width=\linewidth]{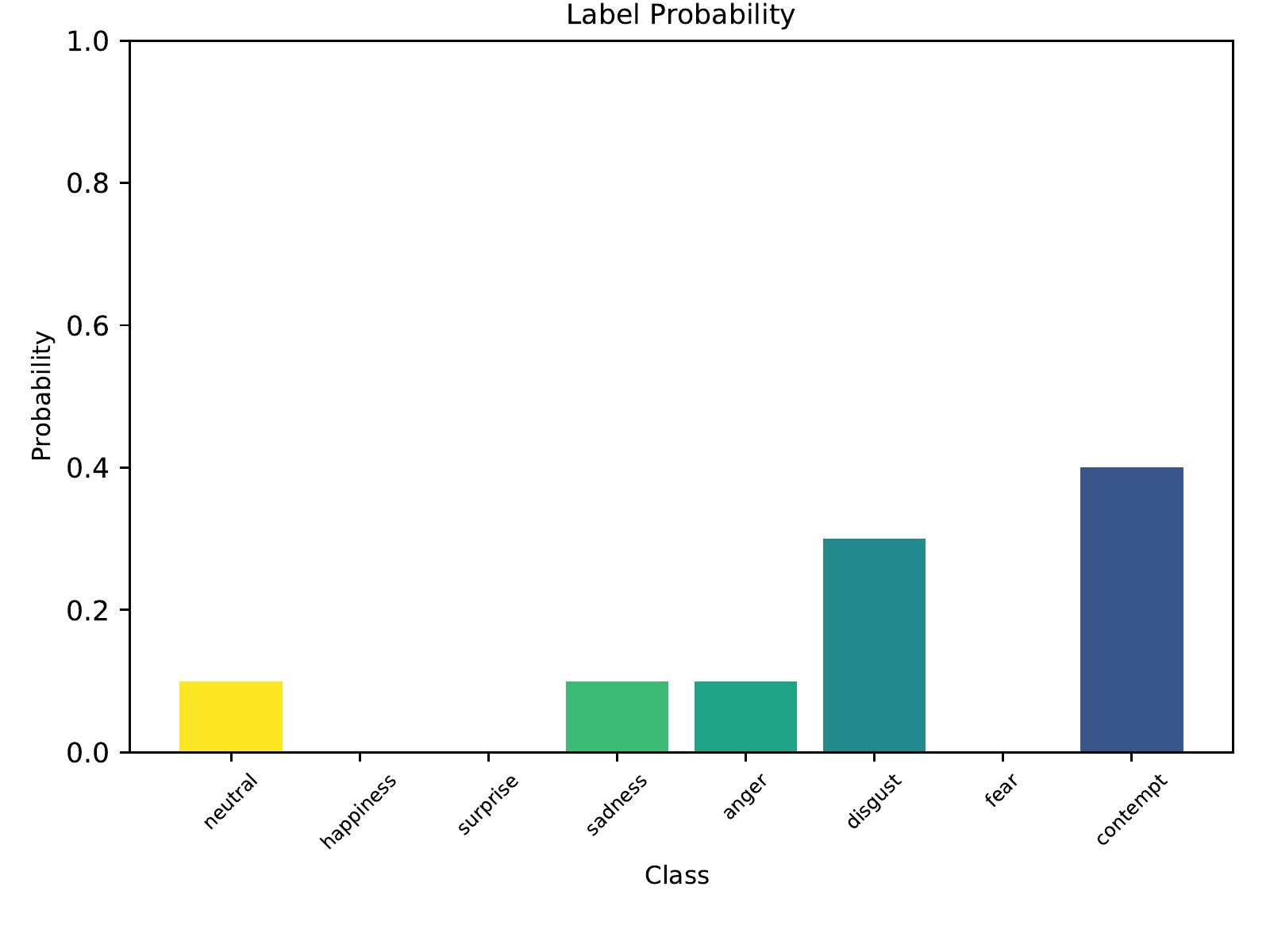}
        \caption*{True Label}
    \end{subfigure}
    \begin{subfigure}[b]{0.17\linewidth}
        \includegraphics[width=\linewidth]{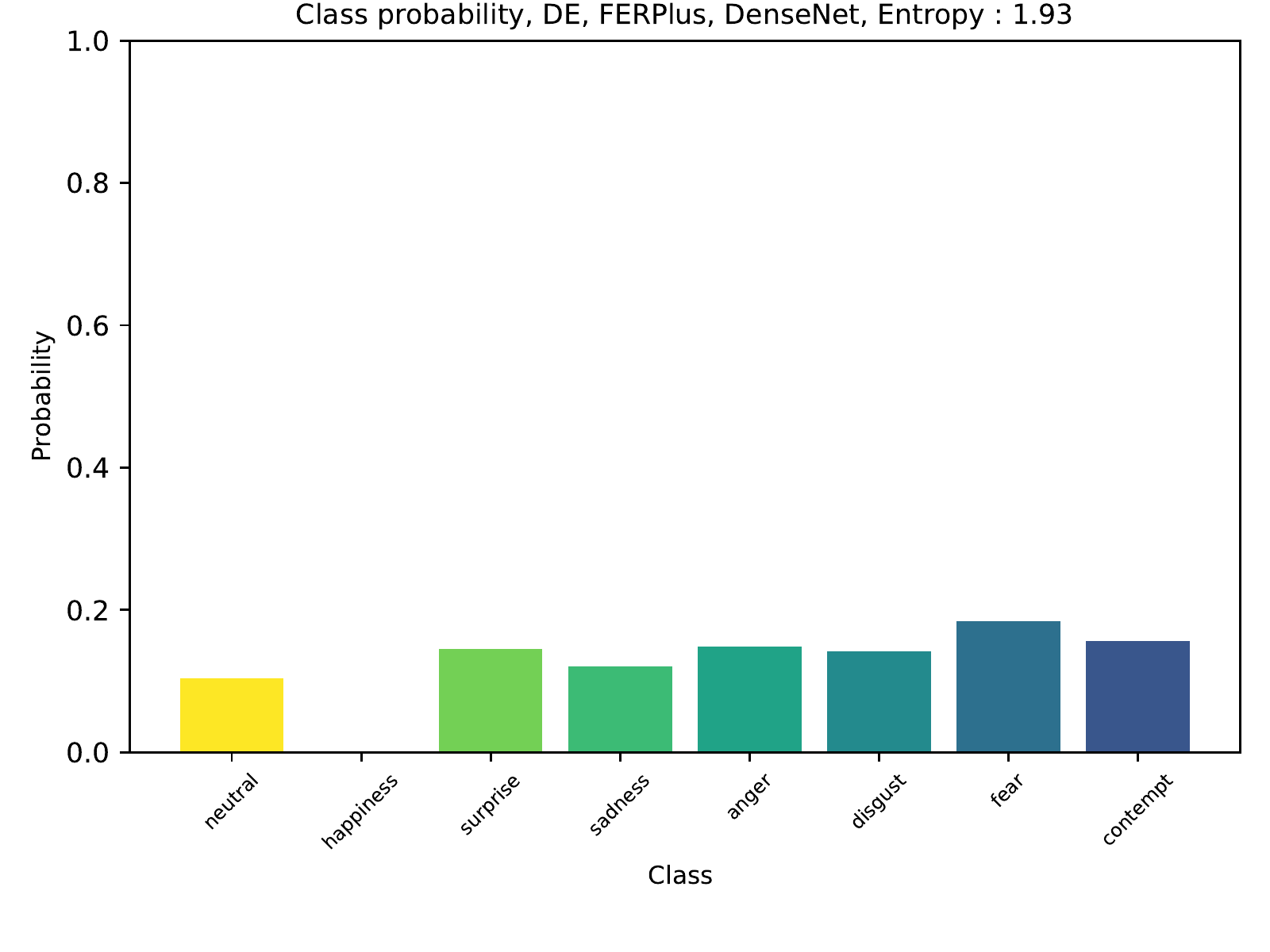}
        \caption*{Fear}
    \end{subfigure}
    \begin{subfigure}[b]{0.17\linewidth}
        \includegraphics[width=\linewidth]{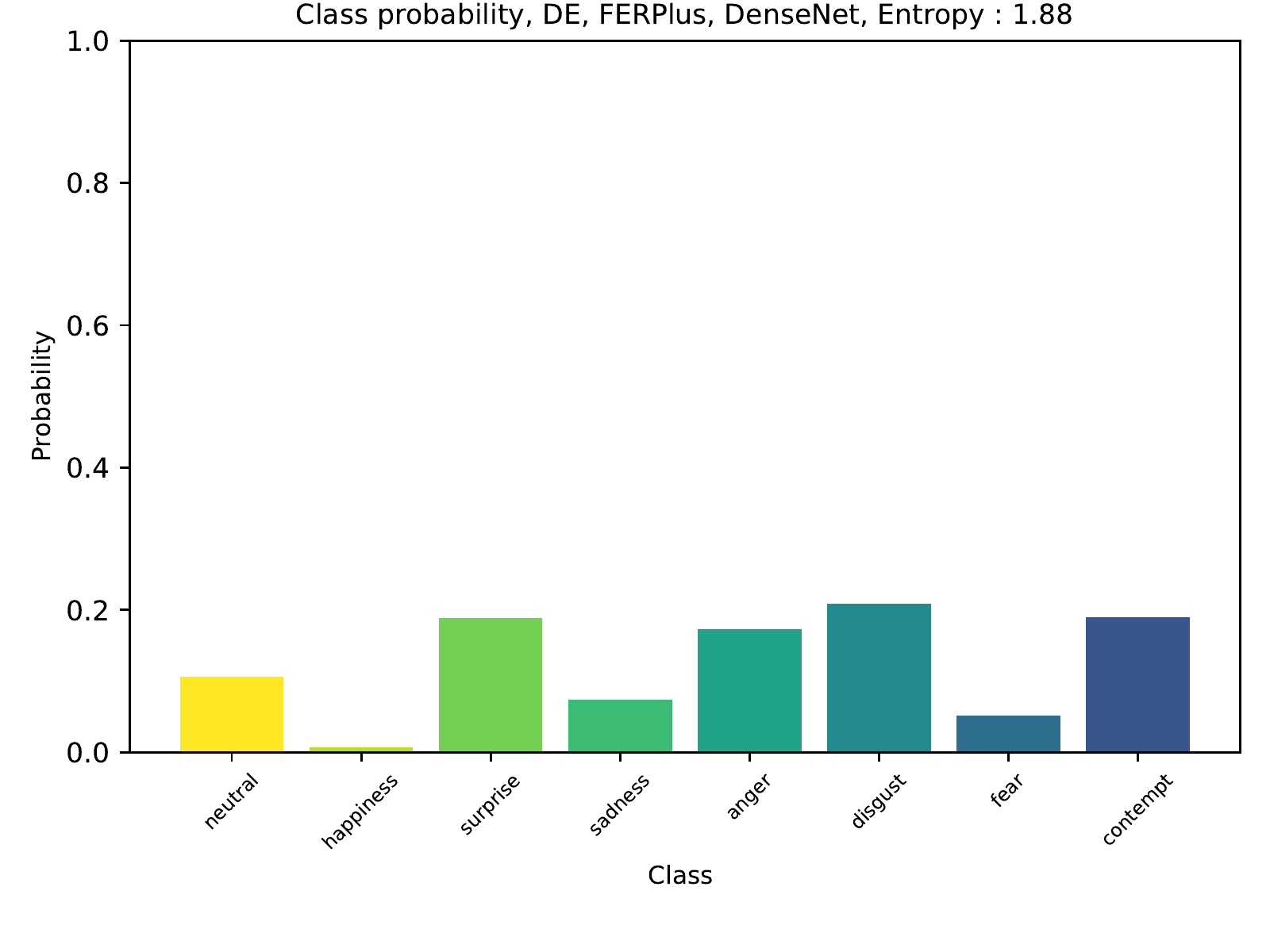}
        \caption*{Disgust}
    \end{subfigure}
    \begin{subfigure}[b]{0.17\linewidth}
        \includegraphics[width=\linewidth]{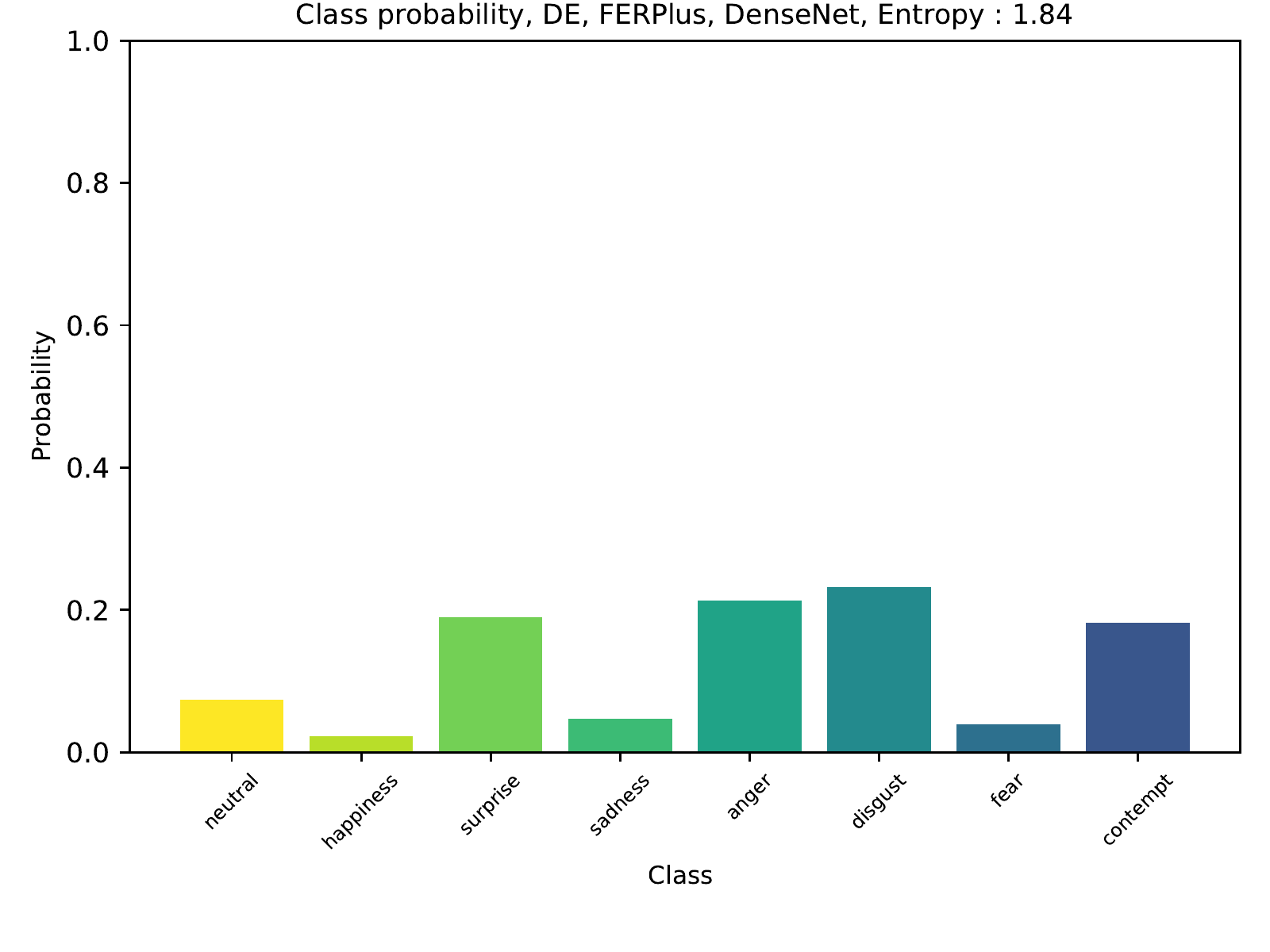}
        \caption*{Disgust}
    \end{subfigure}
    \begin{subfigure}[b]{0.17\linewidth}
        \includegraphics[width=\linewidth]{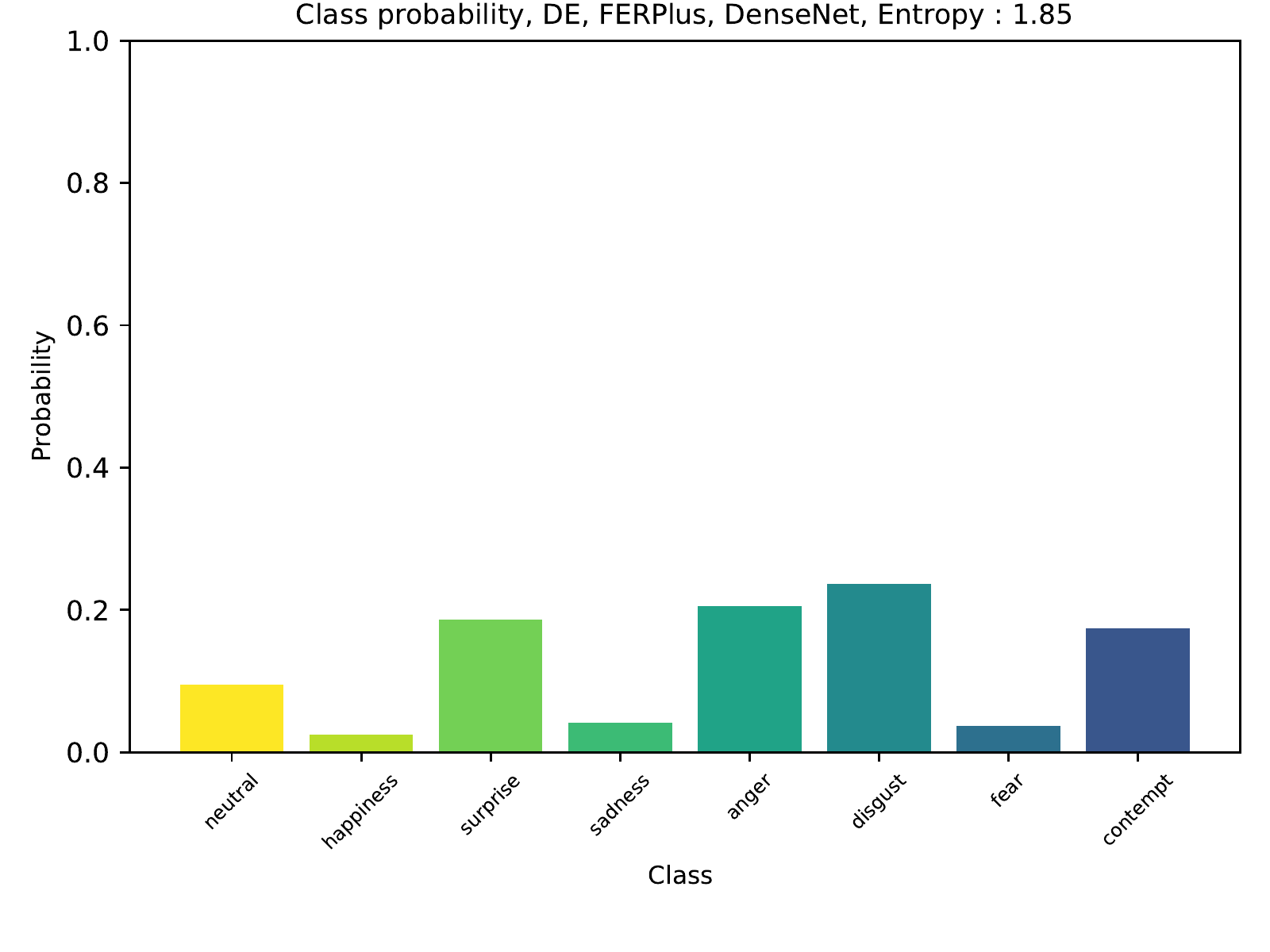}
        \caption*{Disgust}
    \end{subfigure}
    \begin{subfigure}[b]{0.09\linewidth}
        \includegraphics[width=\linewidth]{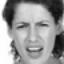}
        \vspace*{0.1em}
        \caption*{Anger}
    \end{subfigure}
    \begin{subfigure}[b]{0.17\linewidth}
        \includegraphics[width=\linewidth]{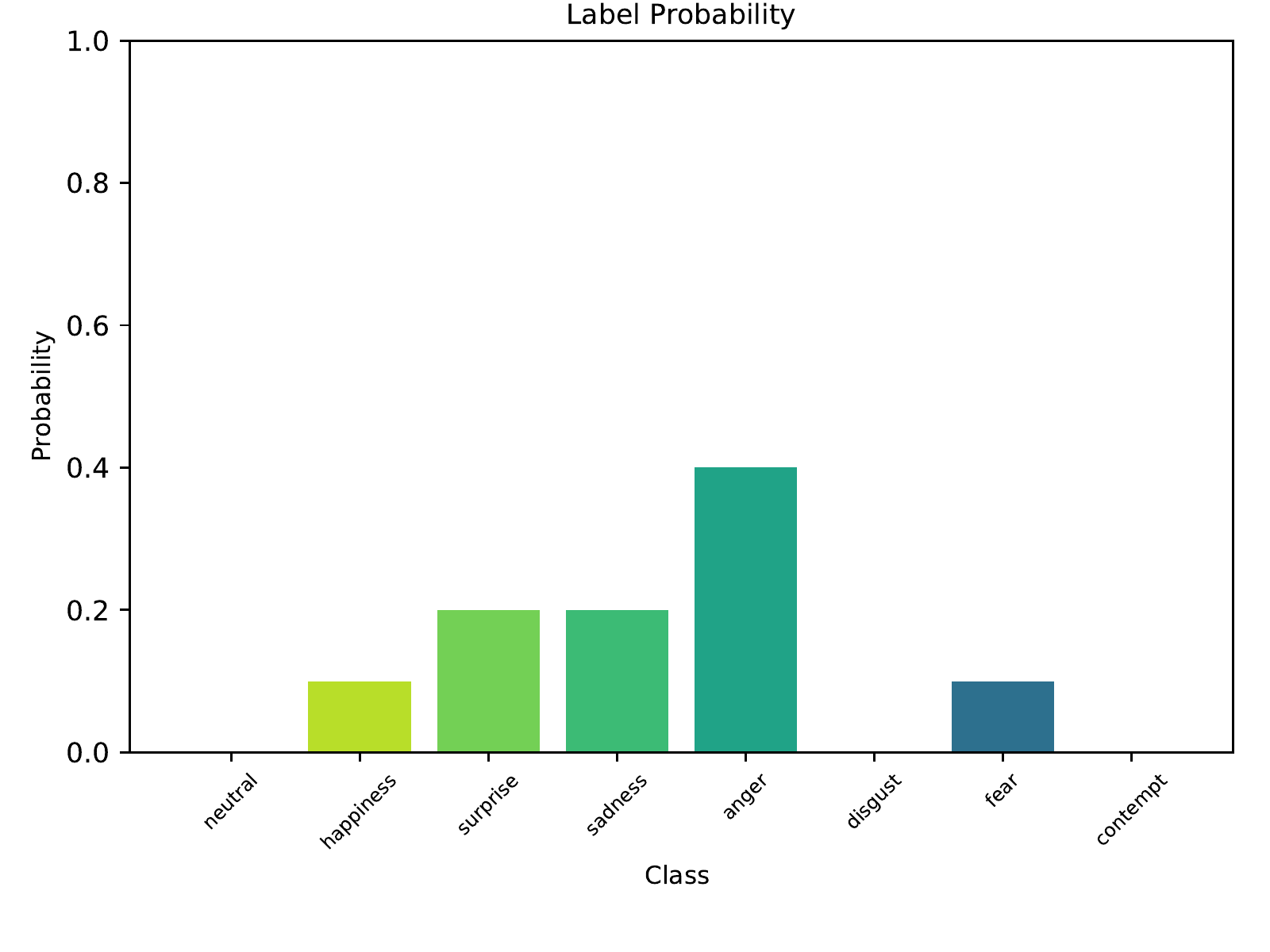}
        \caption*{True Label}
    \end{subfigure}
    \begin{subfigure}[b]{0.17\linewidth}
        \includegraphics[width=\linewidth]{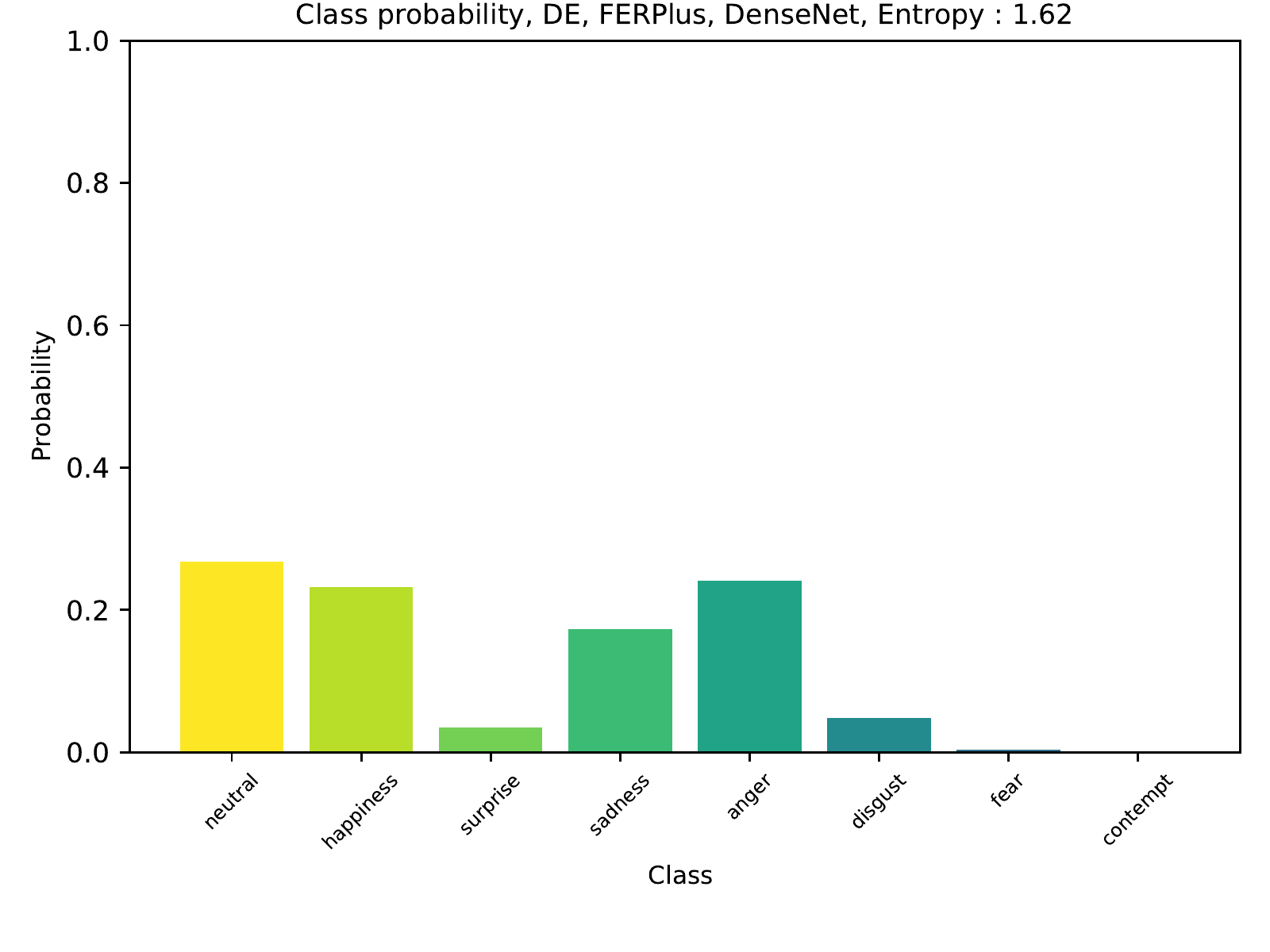}
        \caption*{Neutral}
    \end{subfigure}
    \begin{subfigure}[b]{0.17\linewidth}
        \includegraphics[width=\linewidth]{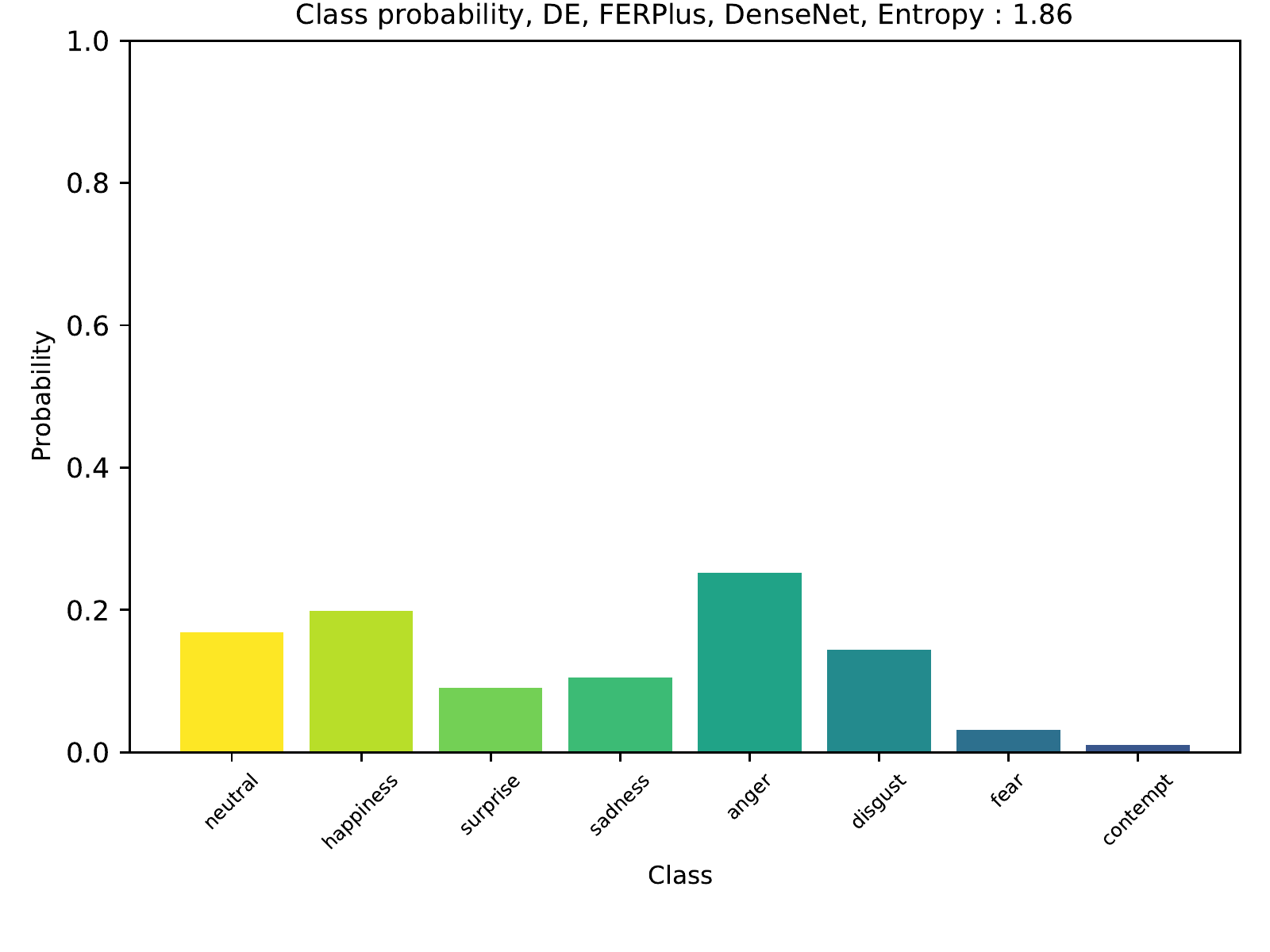}
        \caption*{Anger}
    \end{subfigure}
    \begin{subfigure}[b]{0.17\linewidth}
        \includegraphics[width=\linewidth]{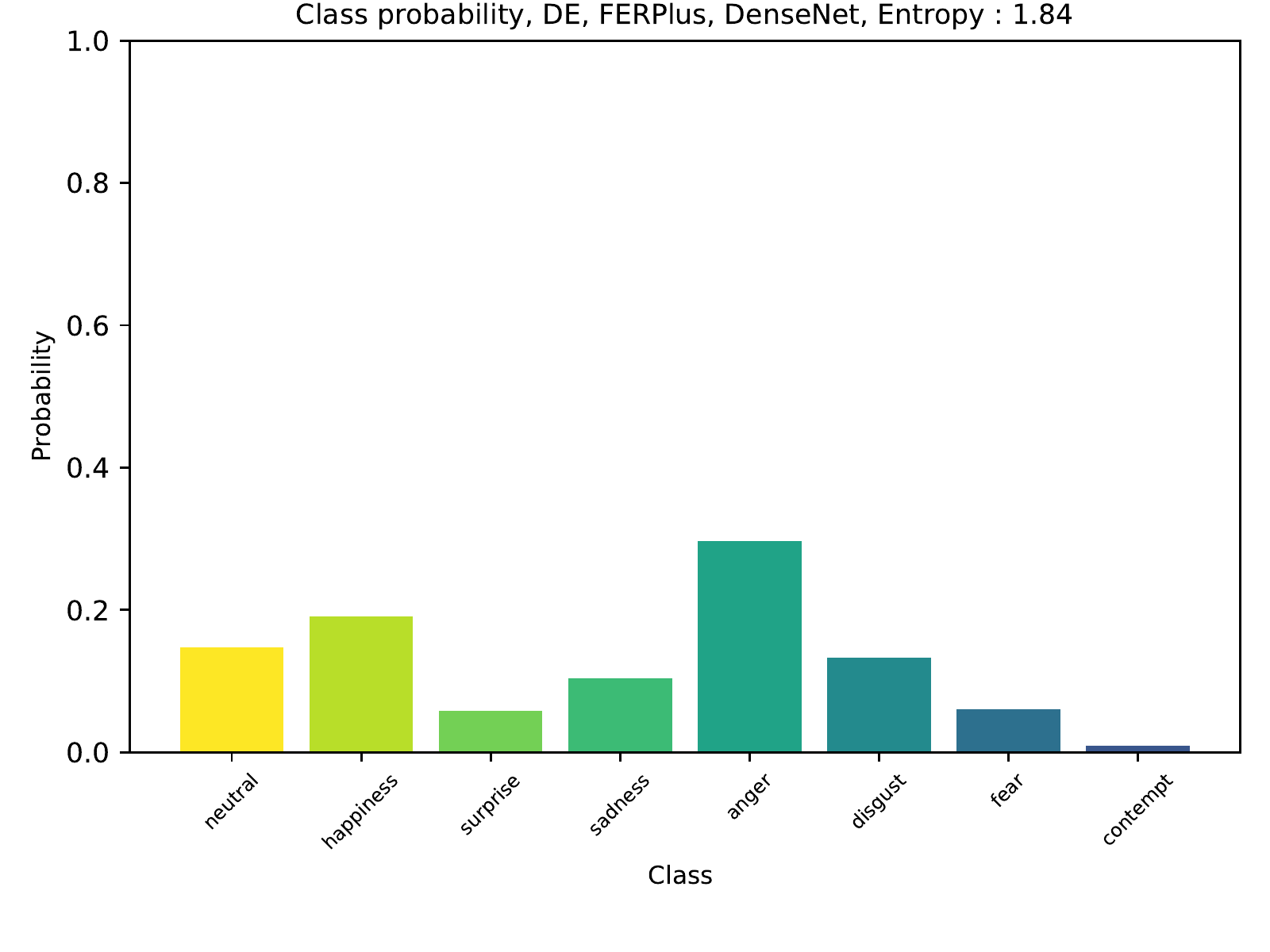}
        \caption*{Anger}
    \end{subfigure}
    \begin{subfigure}[b]{0.17\linewidth}
        \includegraphics[width=\linewidth]{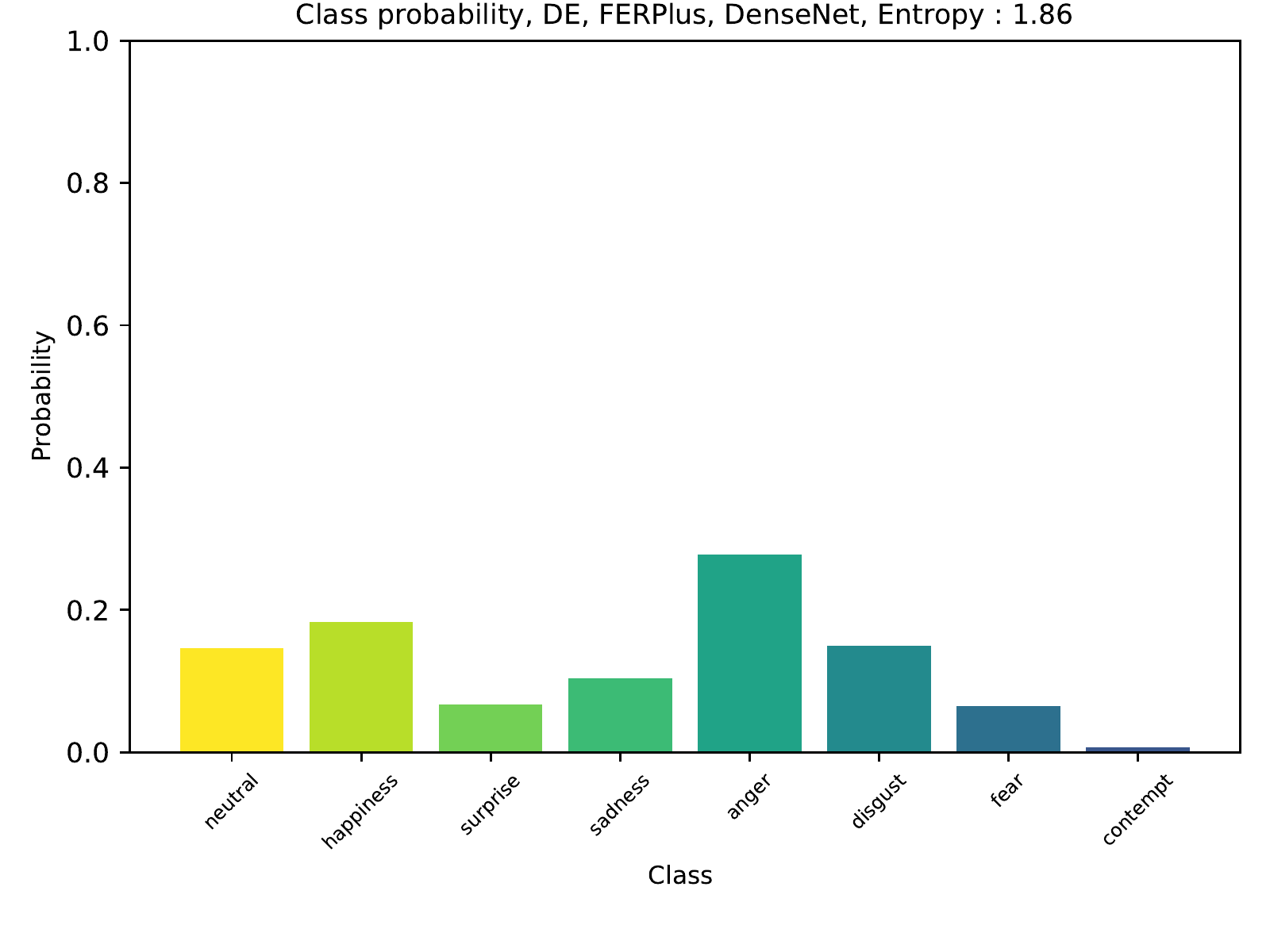}
        \caption*{Anger}
    \end{subfigure}
    
    \begin{subfigure}[b]{0.09\linewidth}
        \includegraphics[width=\linewidth]{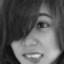}
        \caption*{Happiness}
    \end{subfigure}
    \begin{subfigure}[b]{0.17\linewidth}
        \includegraphics[width=\linewidth]{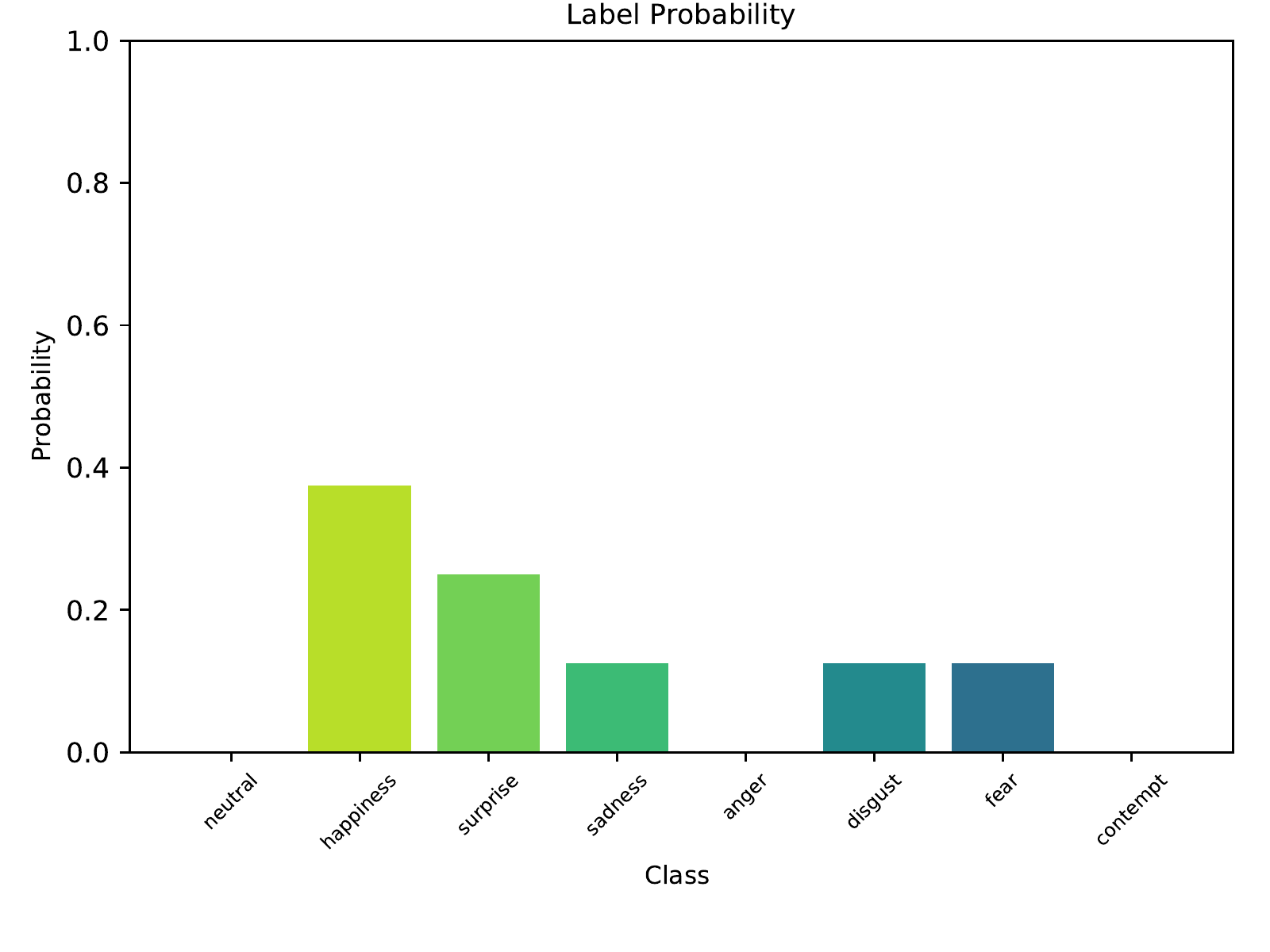}
        \caption*{True Label}
    \end{subfigure}
    \begin{subfigure}[b]{0.17\linewidth}
        \includegraphics[width=\linewidth]{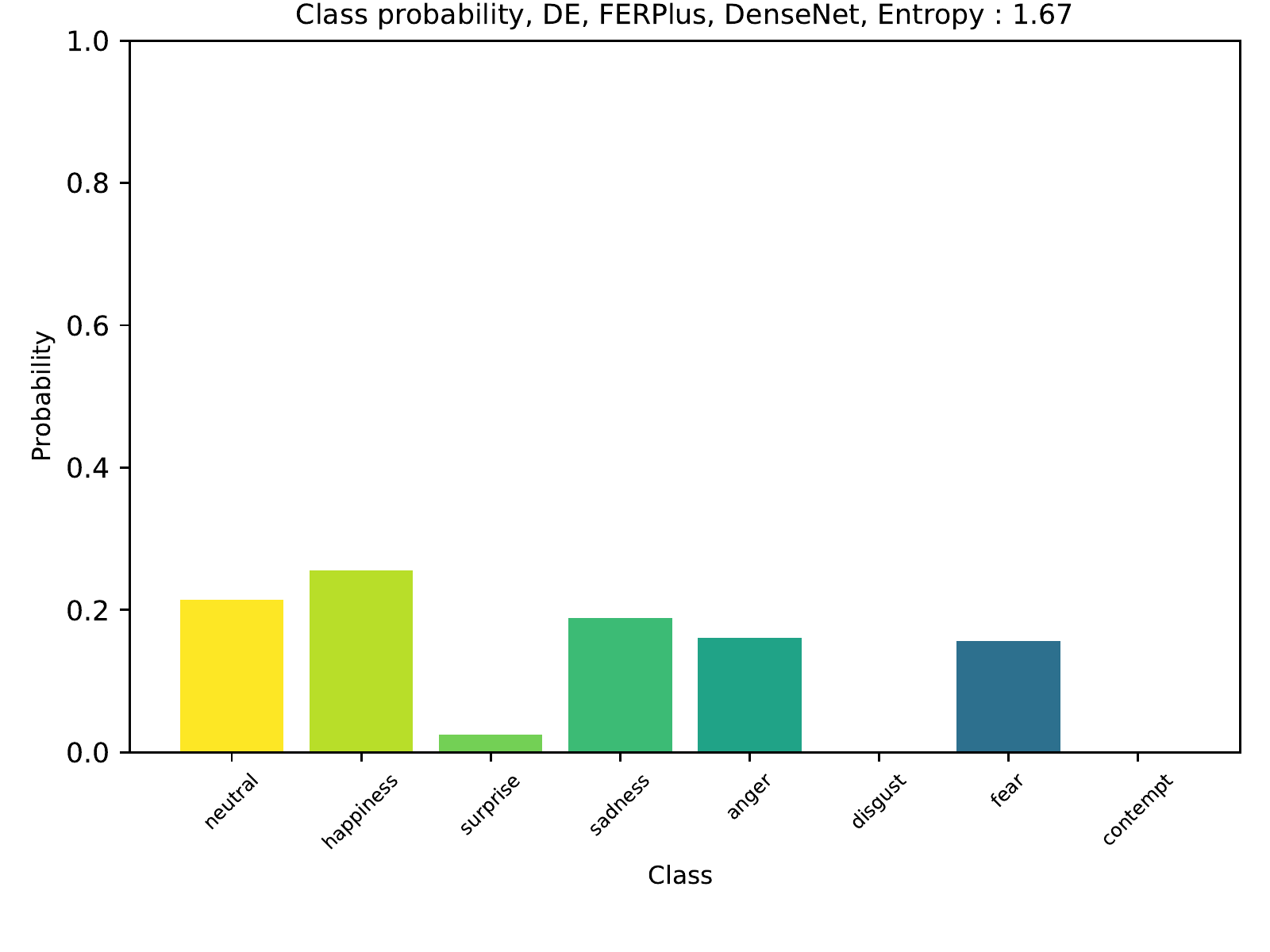}
        \caption*{Happiness}
    \end{subfigure}
    \begin{subfigure}[b]{0.17\linewidth}
        \includegraphics[width=\linewidth]{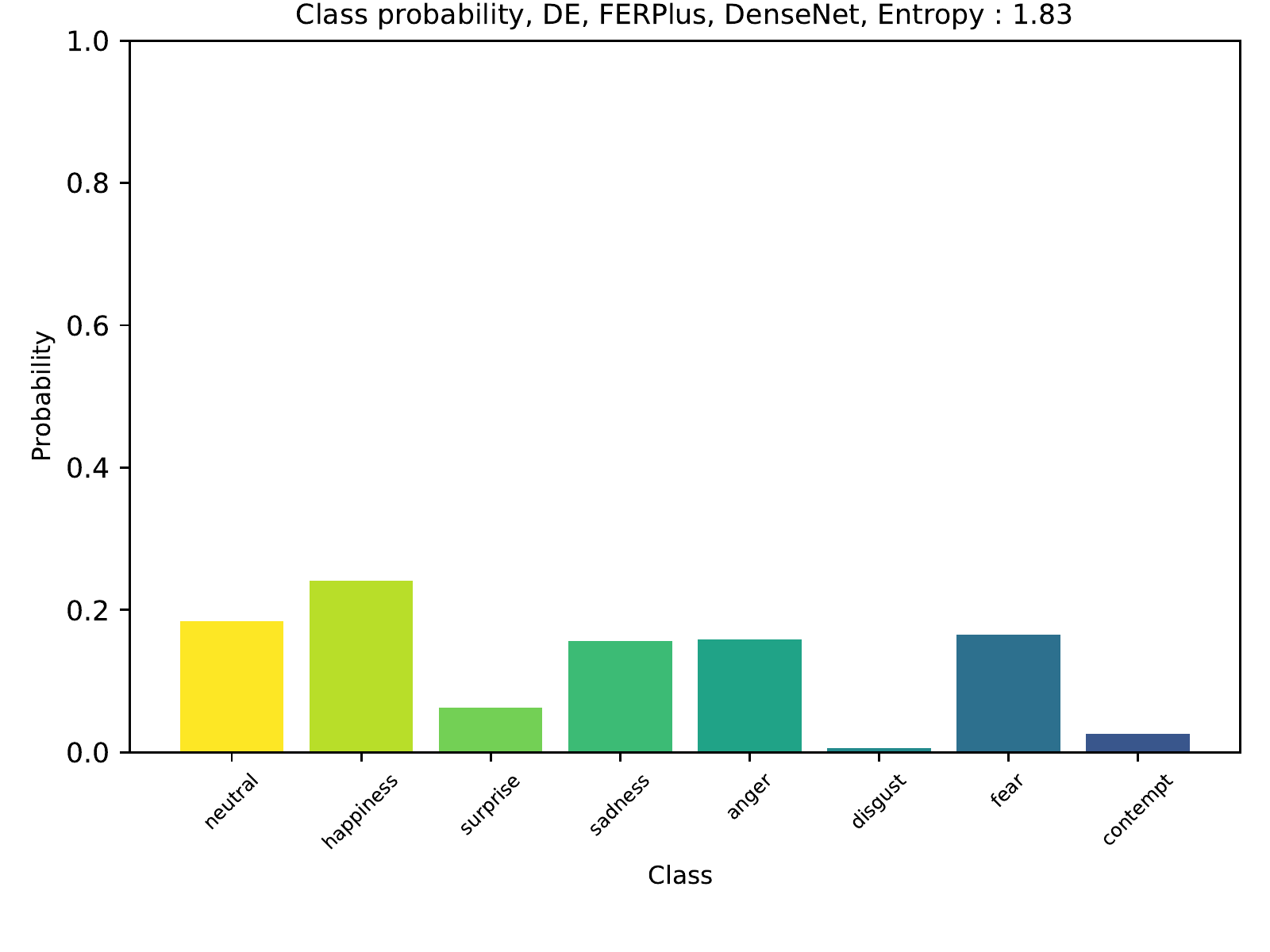}
        \caption*{Happiness}
    \end{subfigure}
    \begin{subfigure}[b]{0.17\linewidth}
        \includegraphics[width=\linewidth]{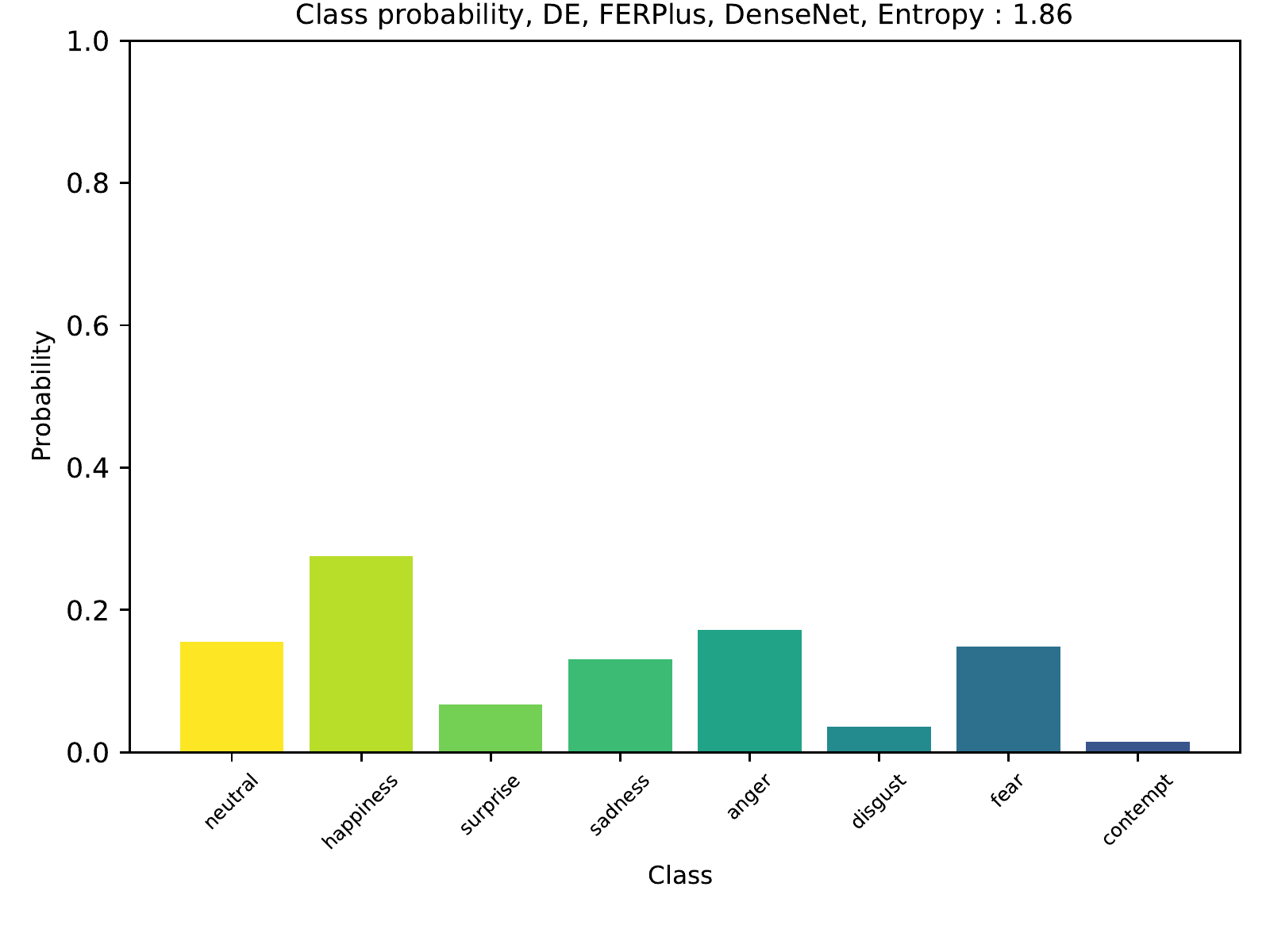}
        \caption*{Happiness}
    \end{subfigure}
    \begin{subfigure}[b]{0.17\linewidth}
        \includegraphics[width=\linewidth]{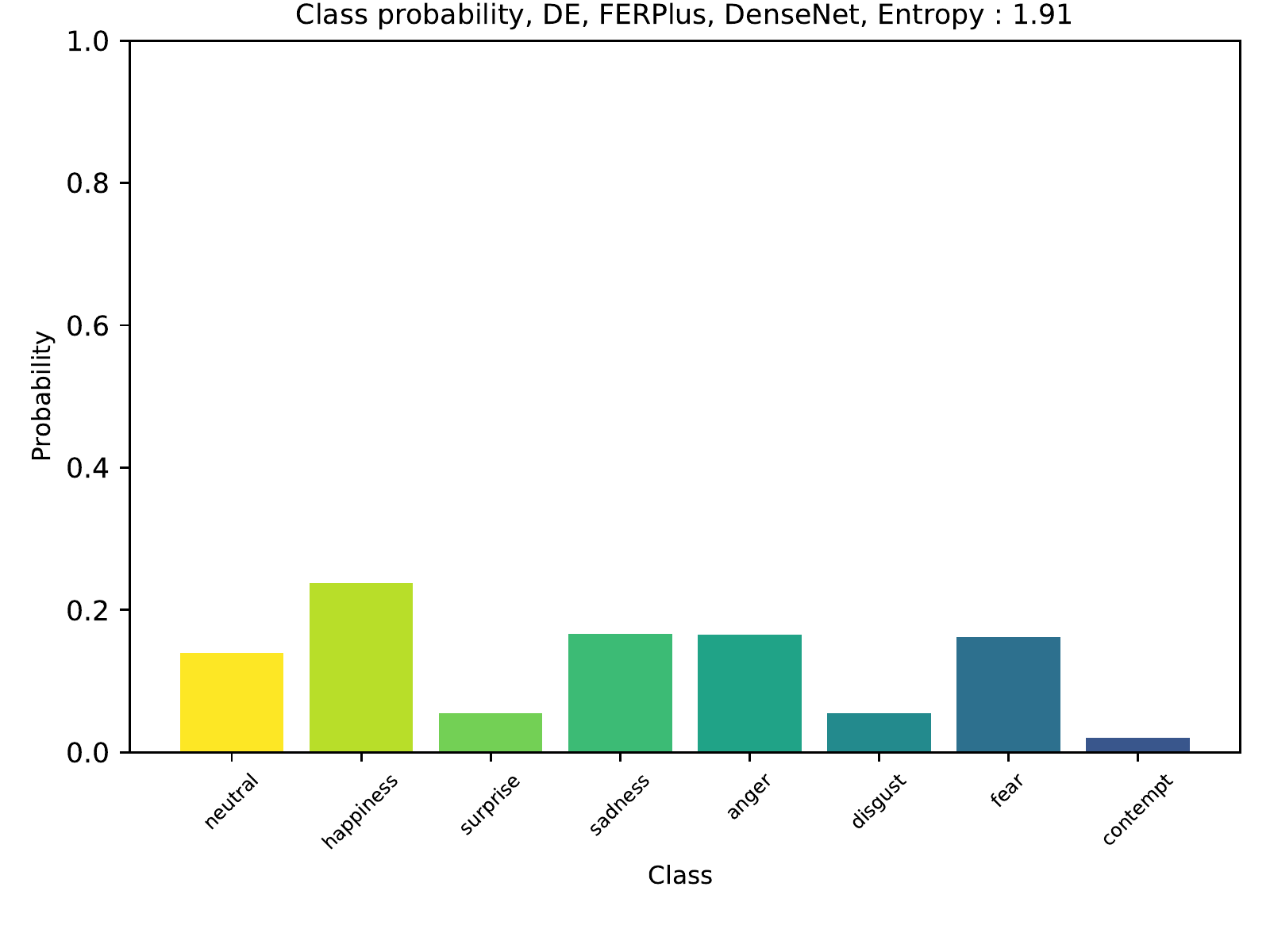}
        \caption*{Happiness}
    \end{subfigure}
    
    \begin{subfigure}[b]{0.09\linewidth}
        \includegraphics[width=\linewidth]{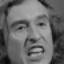}
        \vspace*{0.1em}
        \caption*{Anger}
    \end{subfigure}
    \begin{subfigure}[b]{0.17\linewidth}
        \includegraphics[width=\linewidth]{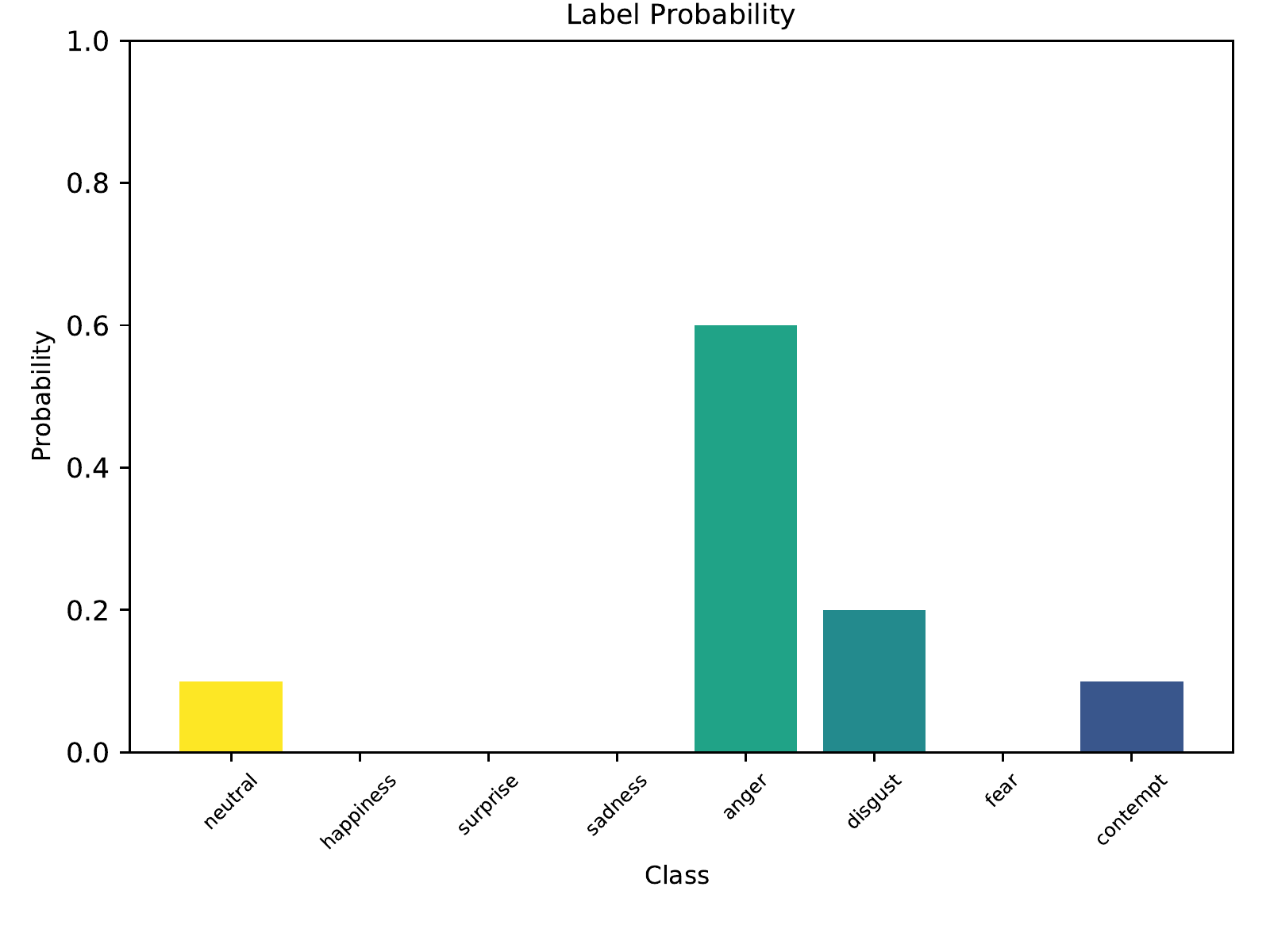}
        \caption*{True Label}
    \end{subfigure}
    \begin{subfigure}[b]{0.17\linewidth}
        \includegraphics[width=\linewidth]{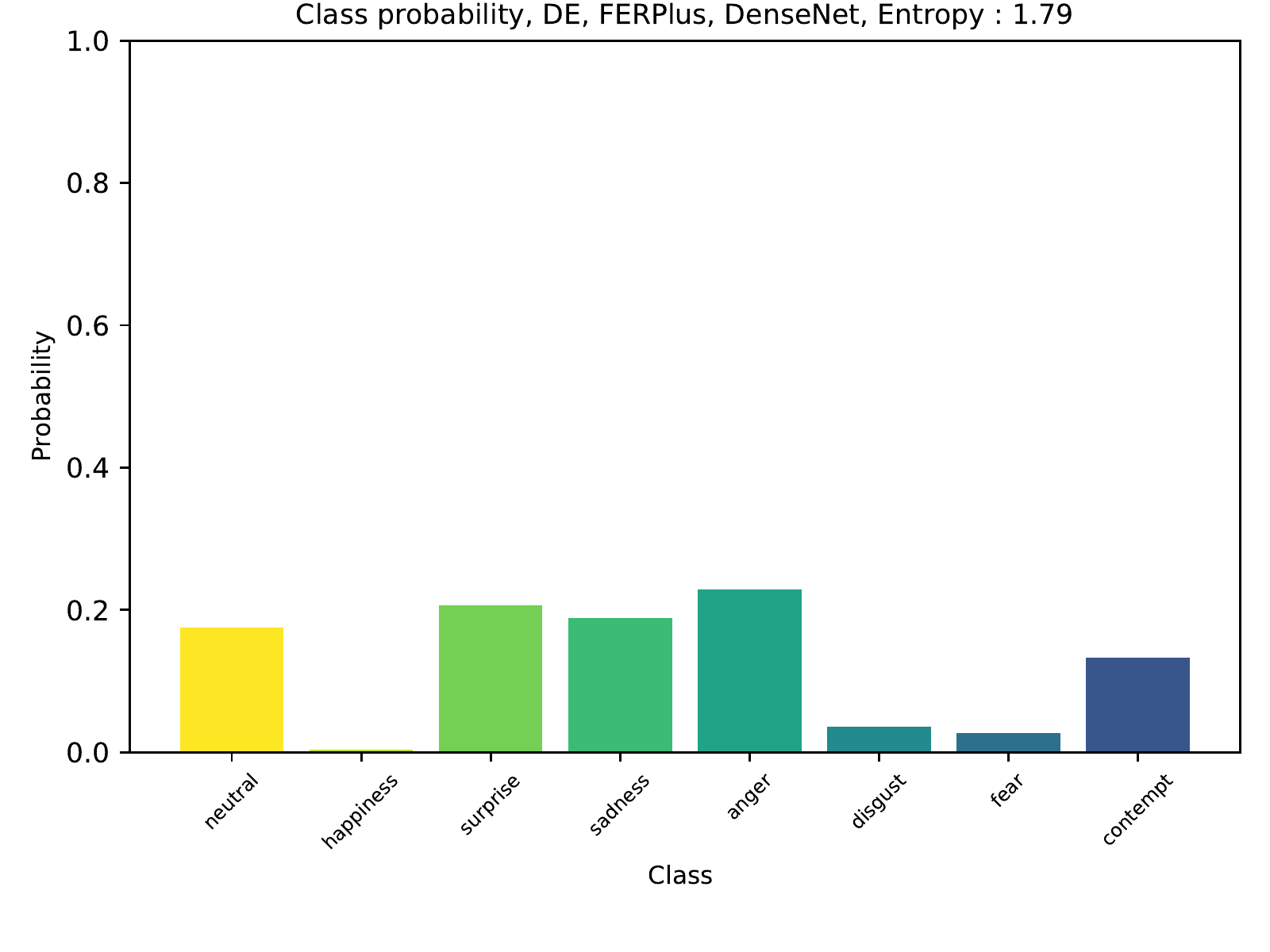}
        \caption*{Anger}
    \end{subfigure}
    \begin{subfigure}[b]{0.17\linewidth}
        \includegraphics[width=\linewidth]{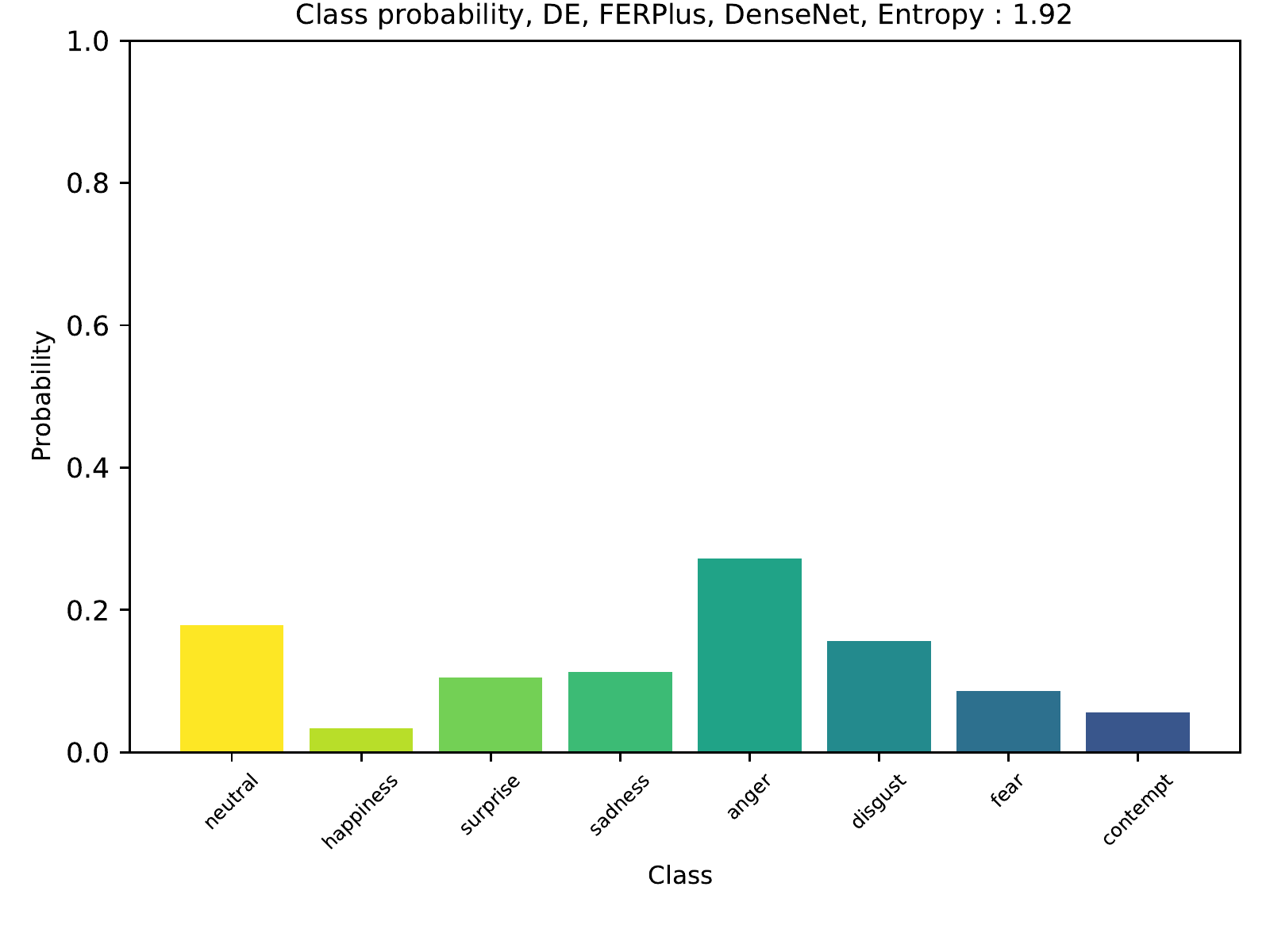}
        \caption*{Anger}
    \end{subfigure}
    \begin{subfigure}[b]{0.17\linewidth}
        \includegraphics[width=\linewidth]{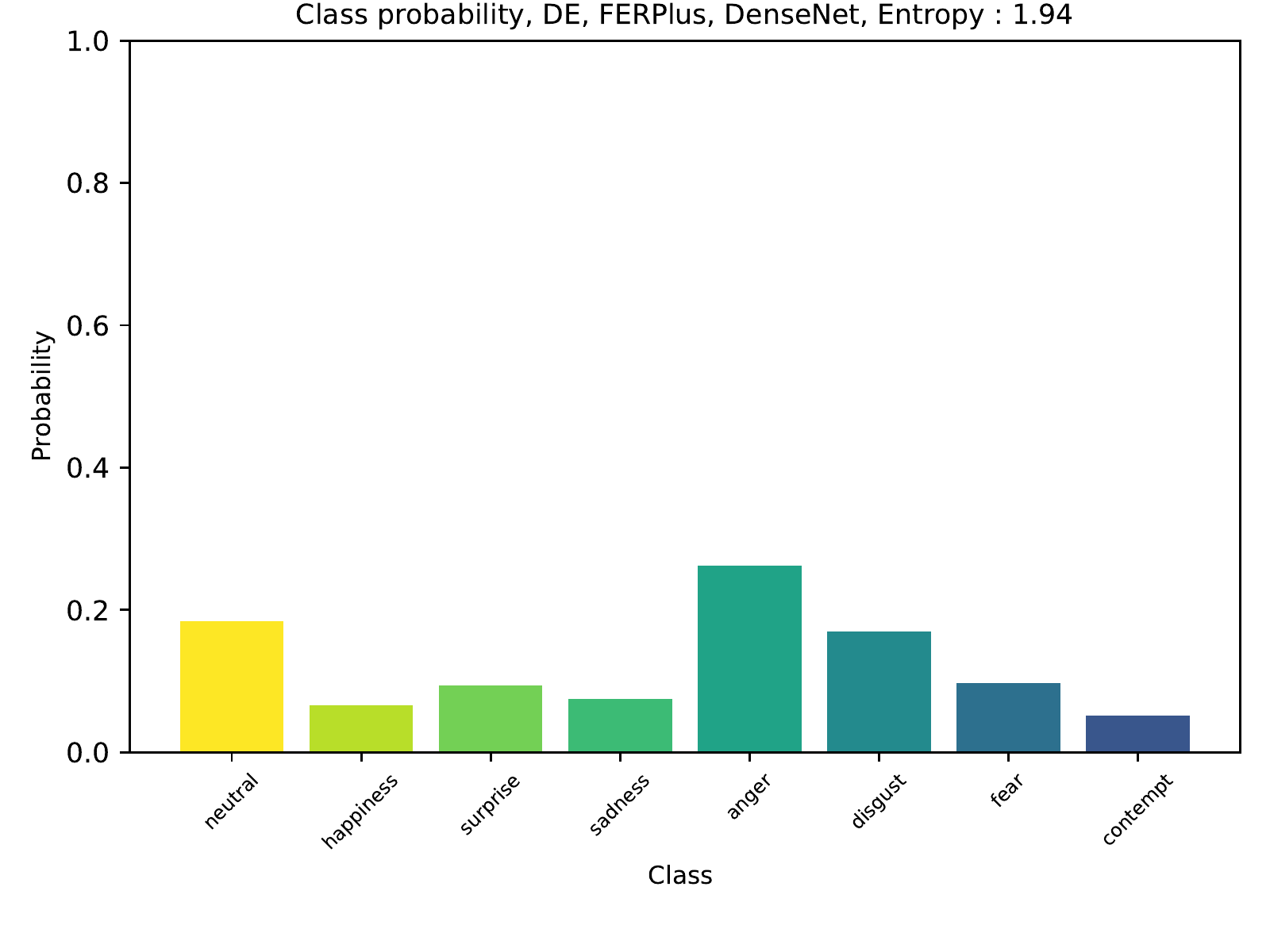}
        \caption*{Anger}
    \end{subfigure}
    \begin{subfigure}[b]{0.17\linewidth}
        \includegraphics[width=\linewidth]{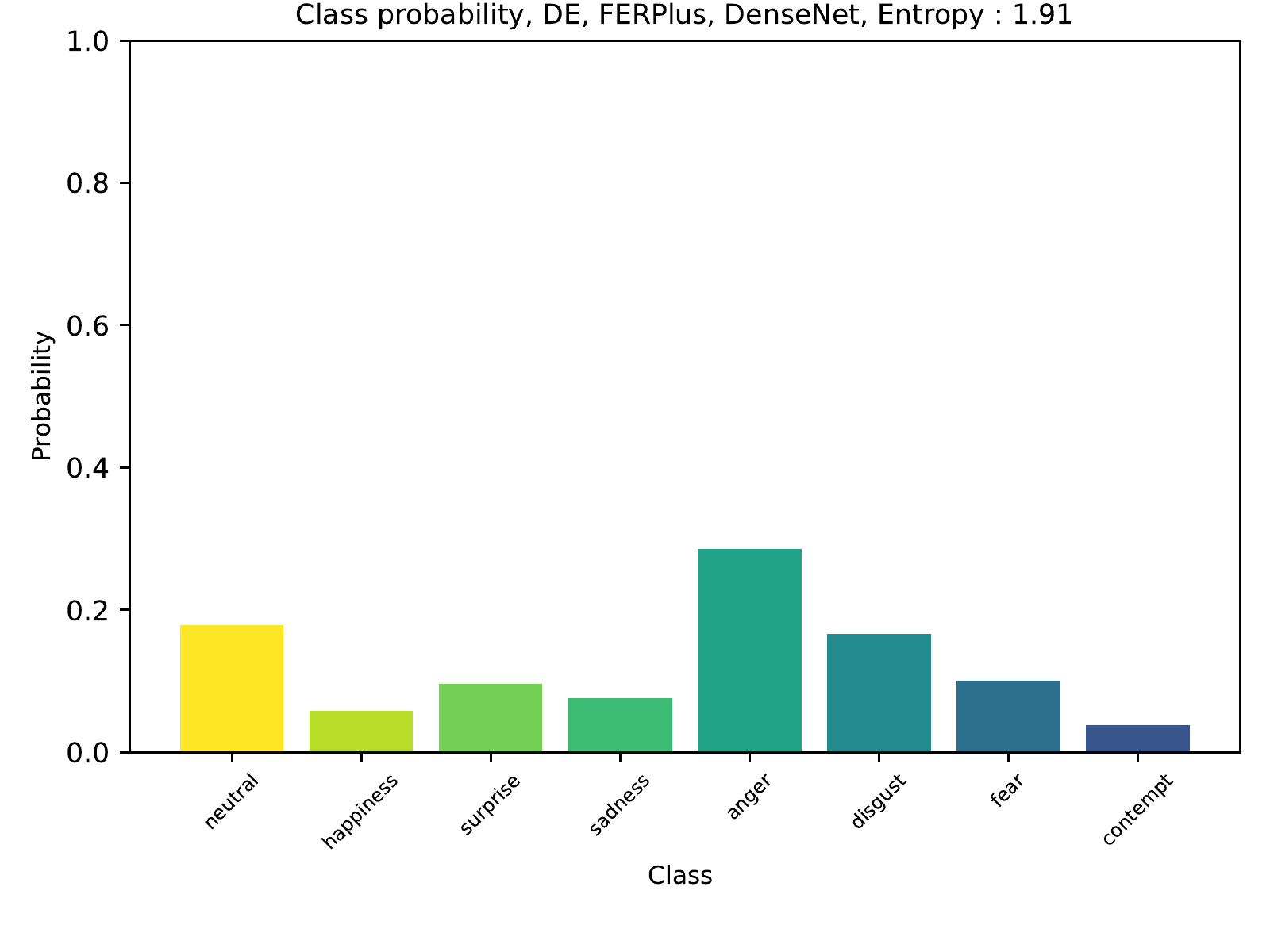}
        \caption*{Anger}
    \end{subfigure}
    
    \begin{subfigure}[b]{0.09\linewidth}
        \includegraphics[width=\linewidth]{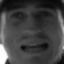}
        \caption*{Happiness}
    \end{subfigure}
    \begin{subfigure}[b]{0.17\linewidth}
        \includegraphics[width=\linewidth]{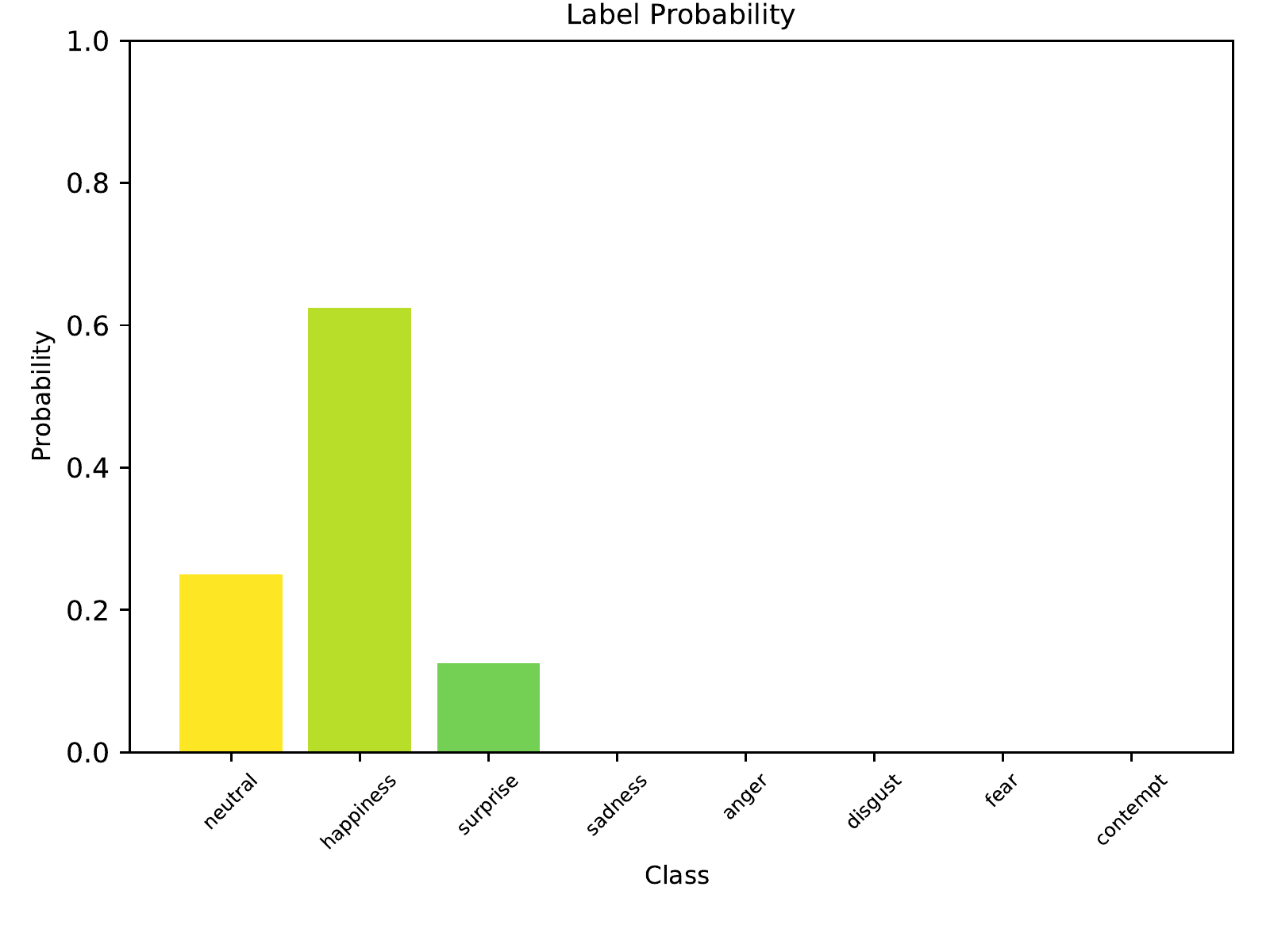}
        \caption*{True Label}
    \end{subfigure}
    \begin{subfigure}[b]{0.17\linewidth}
        \includegraphics[width=\linewidth]{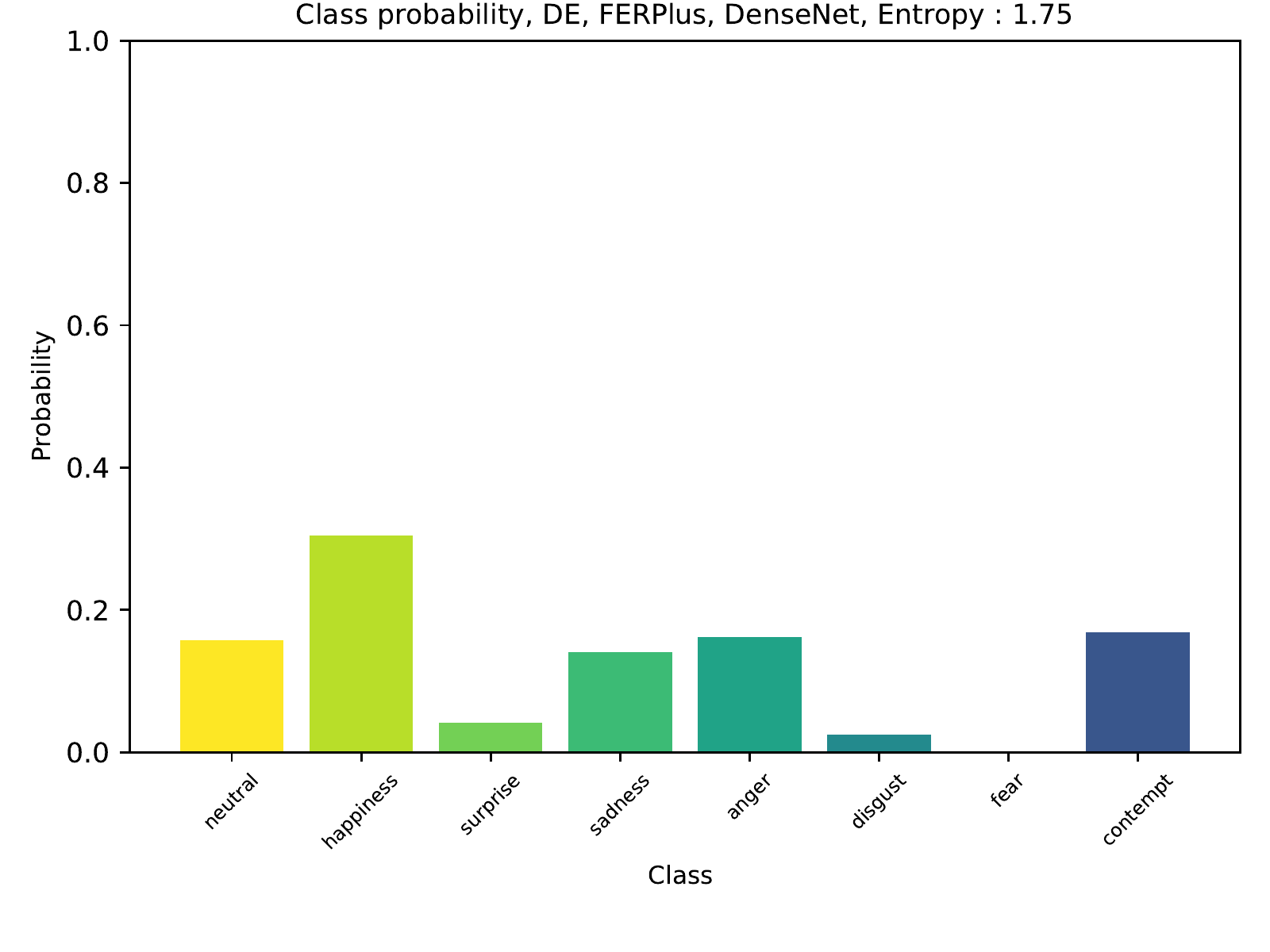}
        \caption*{Happiness}
    \end{subfigure}
    \begin{subfigure}[b]{0.17\linewidth}
        \includegraphics[width=\linewidth]{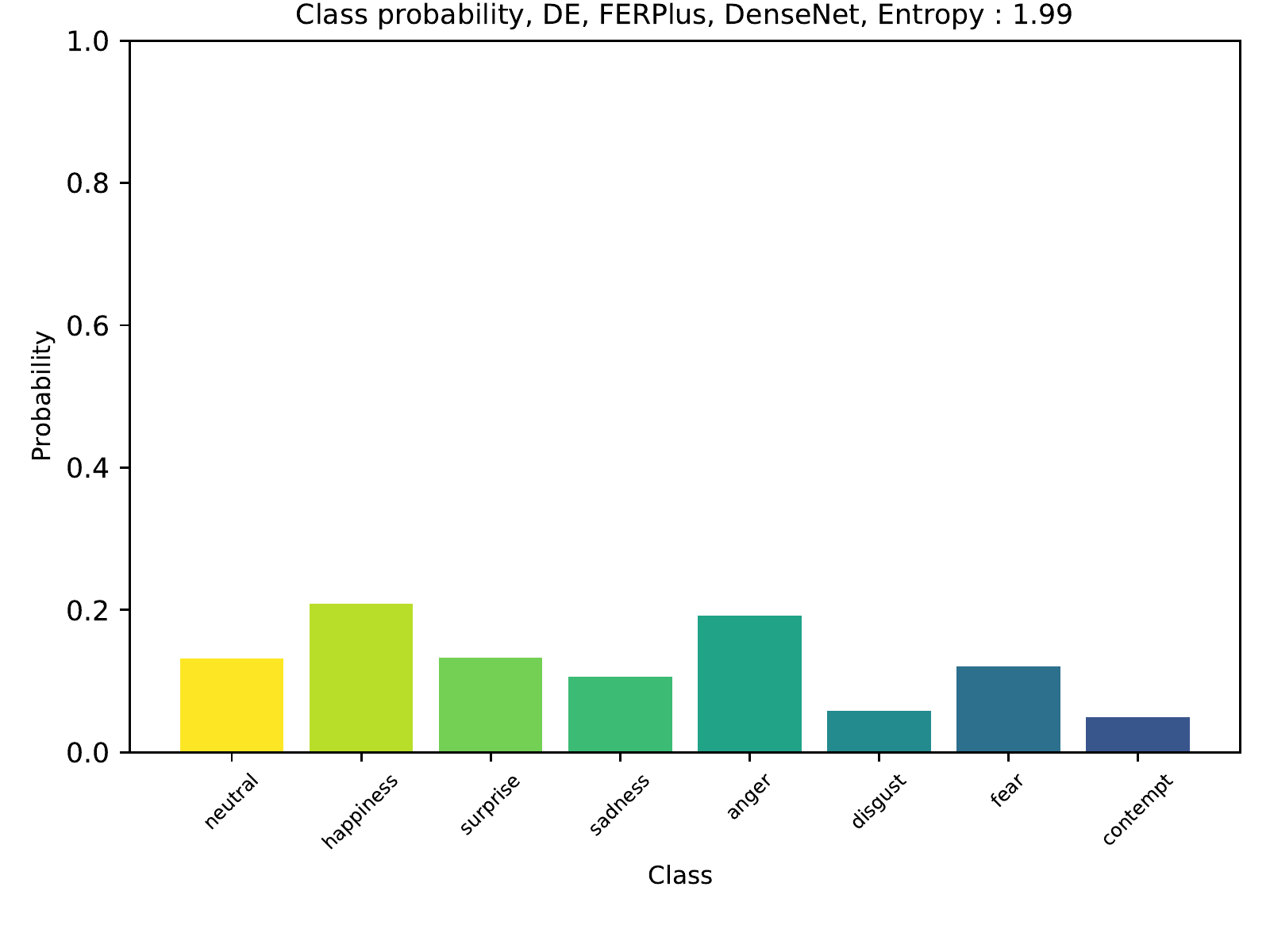}
        \caption*{Happiness}
    \end{subfigure}
    \begin{subfigure}[b]{0.17\linewidth}
        \includegraphics[width=\linewidth]{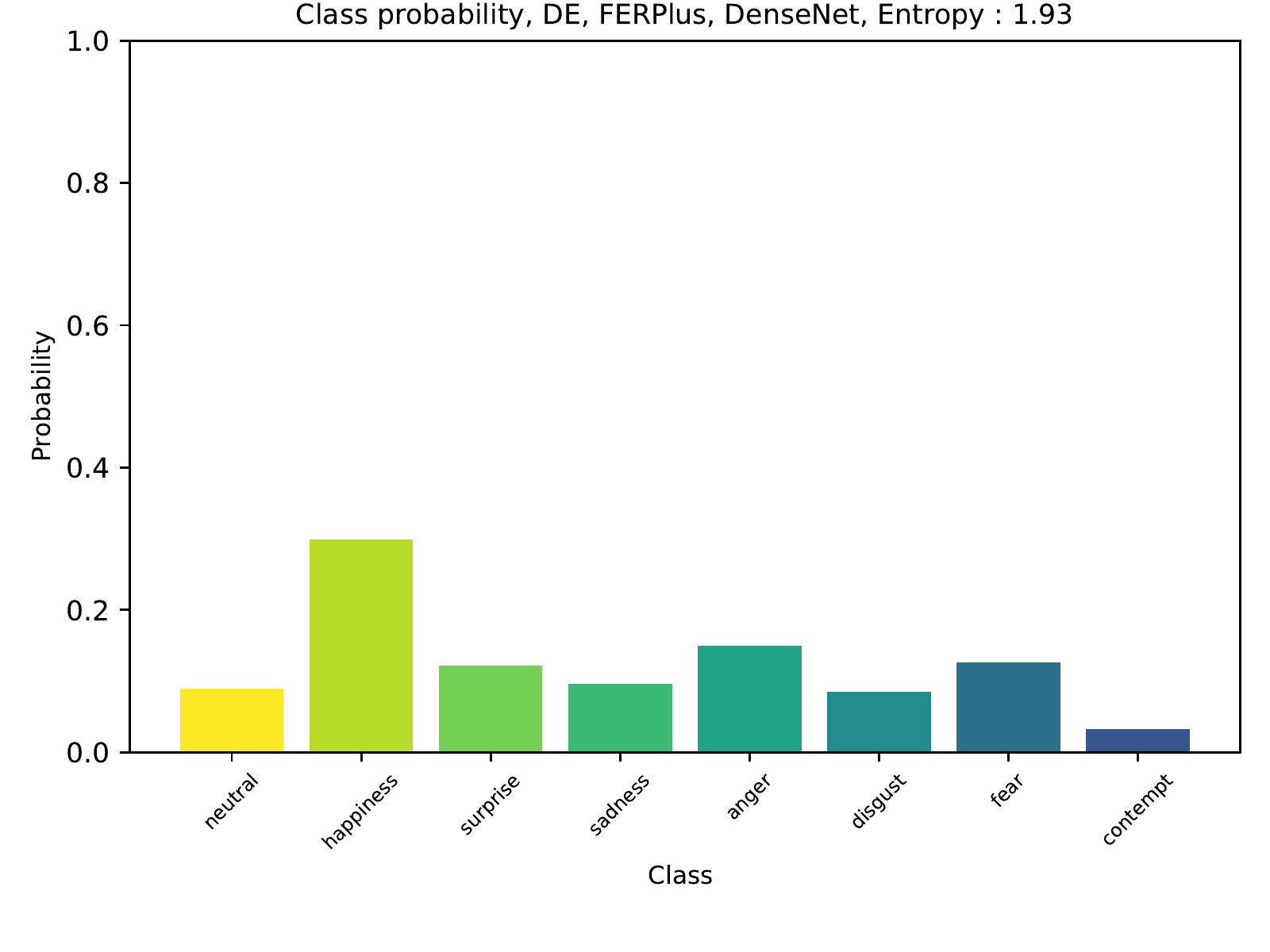}
        \caption*{Happiness}
    \end{subfigure}
    \begin{subfigure}[b]{0.17\linewidth}
        \includegraphics[width=\linewidth]{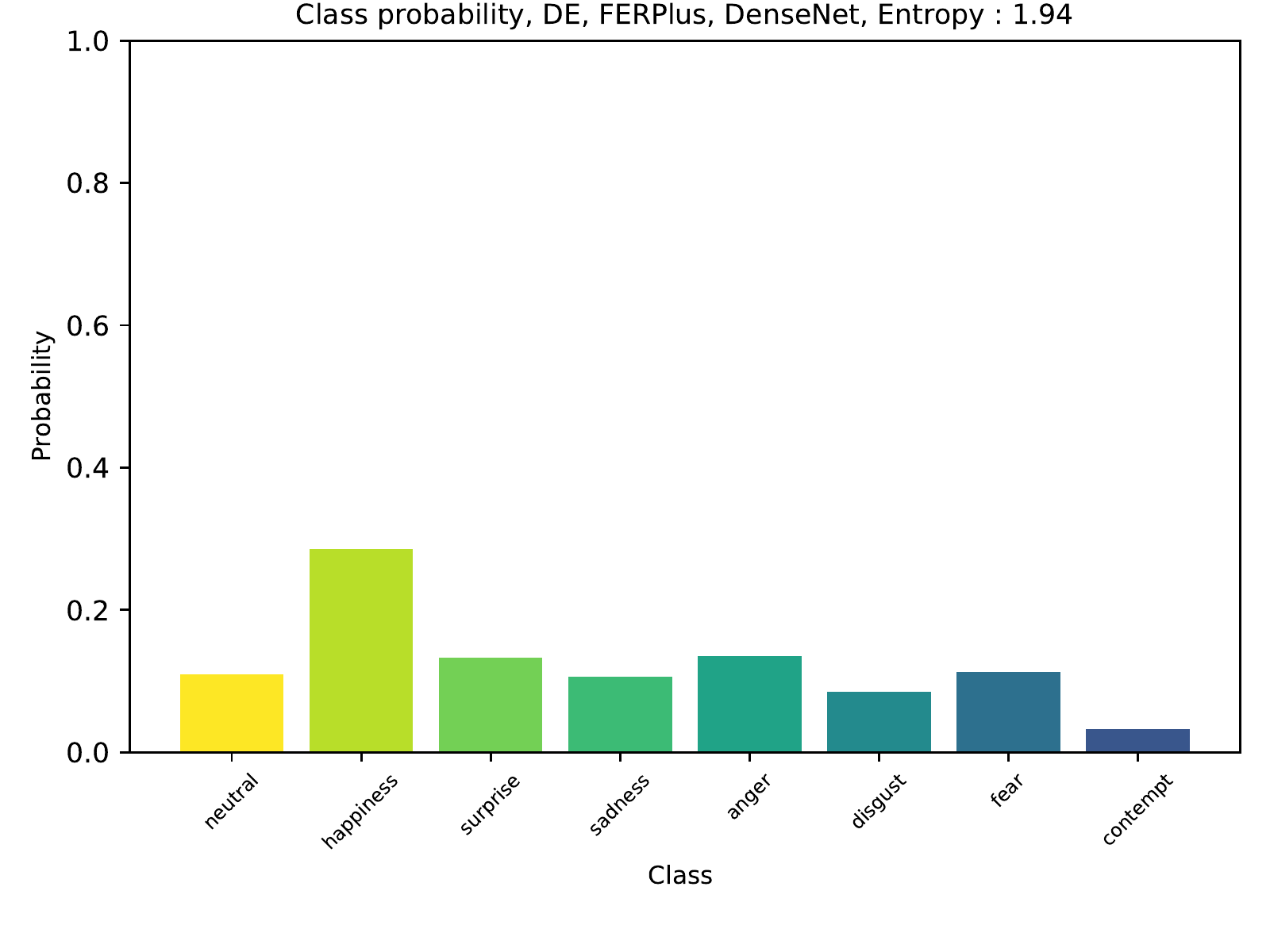}
        \caption*{Happiness}
    \end{subfigure}        
    \caption{Five most uncertain images based on DenseNet model and Deep Ensembles with \# of ensembles and a plot of predictive probabilities using 1, 5, 10 and 15 ensembles. The first column represents the image, and the second its ground truth label distribution. Under each probability plot, the predicted class is presented.}
    \label{fig:ferplus_DE_Densenet_probs}
    \vspace*{-2em}
\end{figure}

Calibration error shown in Fig. \ref{fig:ferplus_calibratione-sample} shows an unusual pattern for all models and uncertainty methods, as the calibration error increases with more samples or ensemble members, instead of decreasing as it does with other datasets (like CIFAR10 and SVHN, as shown by Valdenegro-Toro \cite{valdenegro2019deepsubensemblesBDL}). We interpret these results as that the model's probabilities are closer to represent the true label distribution than the classical network (which can also be seen in Fig. \ref{fig:ferplus_DE_Densenet_probs}).

We believe that our Bayesian neural network models are overall underconfident, which might be undesirable, but this is due to the large aleatoric uncertainty in the labels and input images (which can be validated by looking at the label entropy), not because the models are producing incorrect predictions. The calibration error considers both the correct class and the prediction confidence of that class, but this considers only one correct class per sample, there are no calibration metrics that consider cases of high aleatoric uncertainty, where some classes are visually similar and should be allowed for the model to be confused.

We visualize the top five most uncertain images as computed using entropy of the output probabilities for the DenseNet model using a Deep Ensemble. This is shown in Figure \ref{fig:ferplus_DE_Densenet_probs}. These results show the uncertainty and visual ambiguity between the classes. An ensemble with a single model is equivalent to a classical neural network, and overall it produces a correct but overconfident result. A Deep Ensemble produces probabilities that are more spread across classes, which make more sense for face images with visual ambiguity in terms of which emotion is actually conveyed.

\section{Conclusions and Future Work}

Overall we see multiple benefits from using BNNs over a classical neural network for facial emotion recognition in the FER+ dataset. A BNN is able to model aleatoric and epistemic uncertainty, while providing small improvements in accuracy (in the order of 2-4\%) compared to a classical network, and providing more realistic probability estimates, specially when considering overconfident point predictions made by classical networks \cite{guo2017calibration}.

We also find that usual calibration metrics behave strangely in the presence of high aleatoric uncertainty, with calibration error increasing along with number of samples or ensembles, while in other datasets it generally decreases producing a more calibrated model.

For future work, we wish to evaluate the potential of out of distribution detection based on probability entropy, as a way to detect biases in the model, and prevent wrong predictions to be made in out of distribution settings, which is certainly a concern for skin shades that are far away from the training set \cite{buolamwini2018gender}.

Finally, we wish to explore ways to disentangle aleatoric and epistemic uncertainty, and to train BNNs using other losses that are able to fully utilize the soft labels \cite{peterson2019human} in the FER+ dataset.

\clearpage
%
%
\bibliographystyle{splncs04}
\bibliography{references}

\appendix
\clearpage
\section{Additional Results on Calibration}

This section presents calibration plots for each model and uncertainty method combination, as they did not fit in the main paper.

\begin{figure}[!hb]
    \centering
    \begin{subfigure}[t]{0.43\linewidth}
        \includegraphics[width=\linewidth]{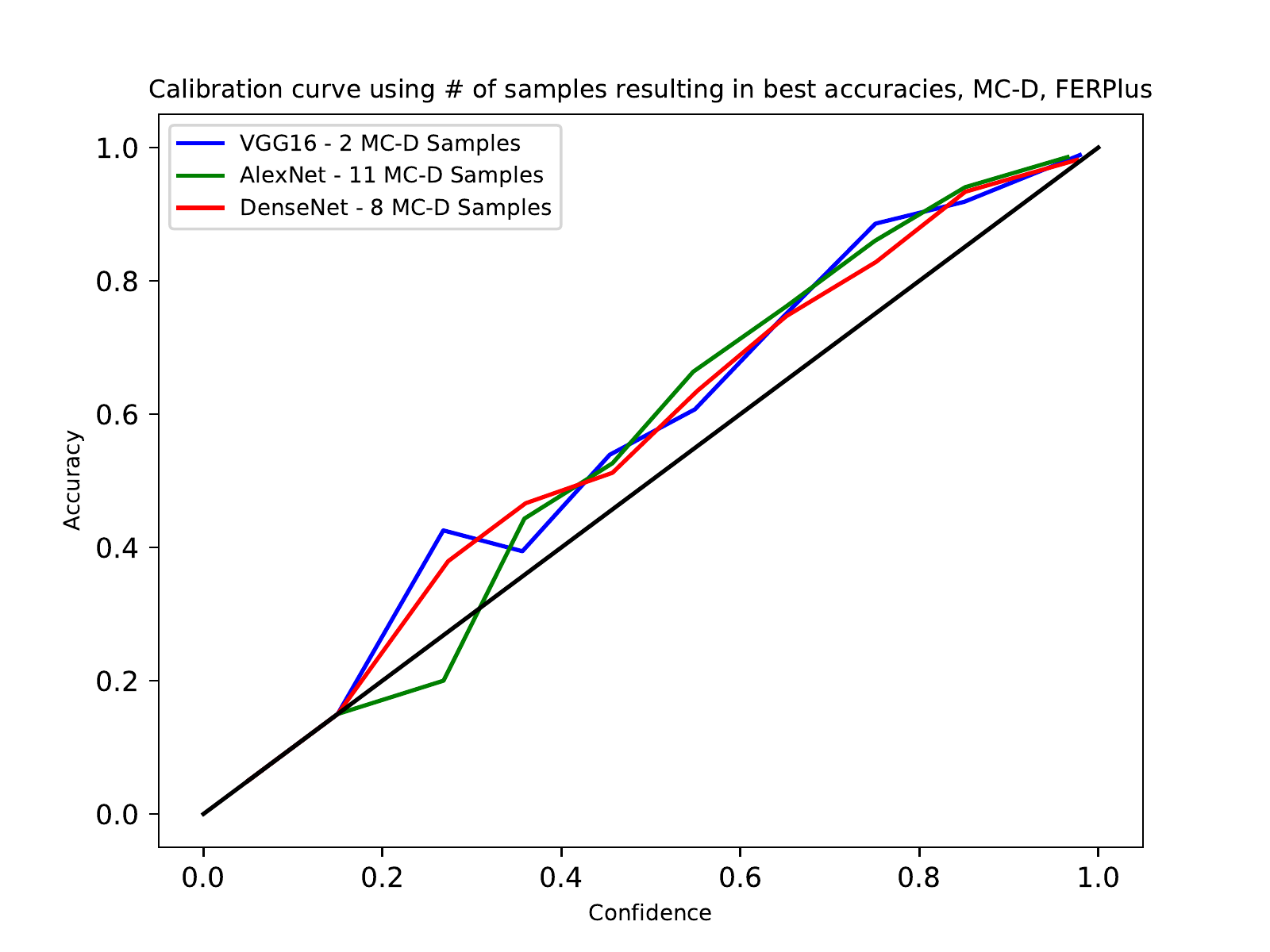}
        \caption{MC-D, \# of samples based on best classification accuracy}
    \end{subfigure}
    \begin{subfigure}[t]{0.43\linewidth}
        \includegraphics[width=\linewidth]{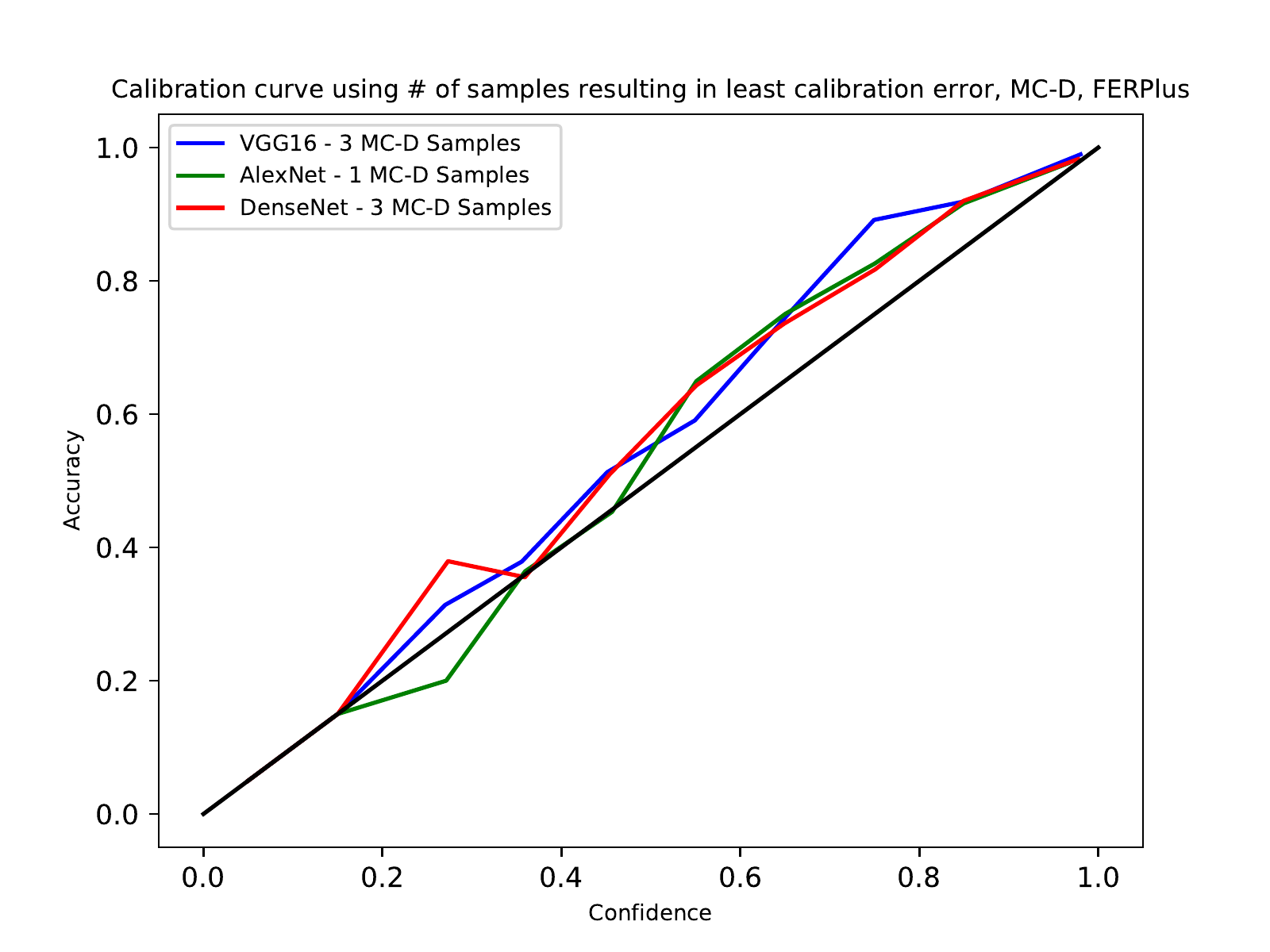}
        \caption{MC-D, \# of samples based on best calibration error}
    \end{subfigure}
    
    \begin{subfigure}[t]{0.43\linewidth}
        \includegraphics[width=\linewidth]{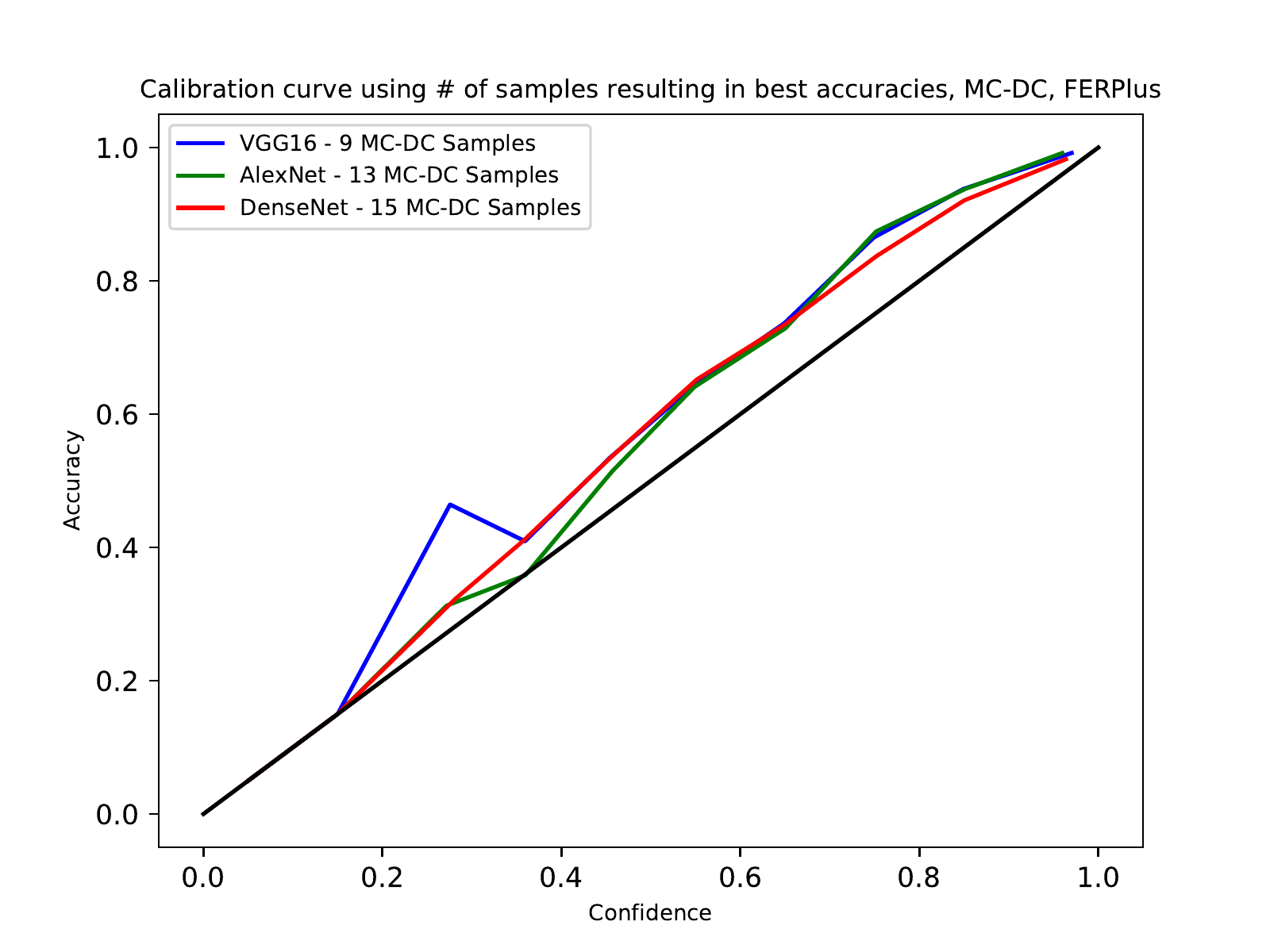}
        \caption{MC-DC, \# of samples based on best classification accuracy}
    \end{subfigure}
    \begin{subfigure}[t]{0.43\linewidth}
        \includegraphics[width=\linewidth]{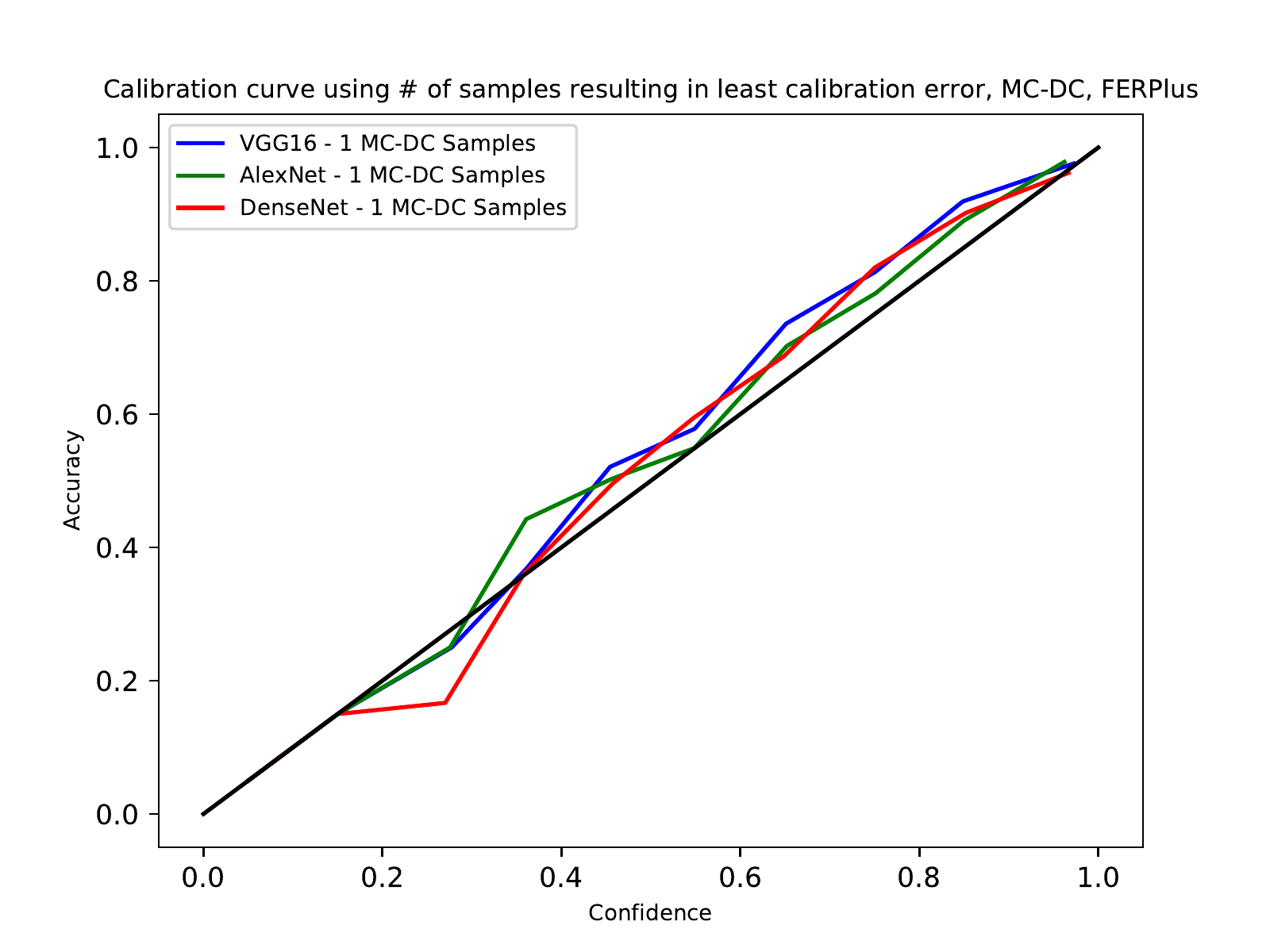}
        \caption{MC-DC, \# of samples based on best calibration error}
        \label{fig:ferplus_calibration_d}
    \end{subfigure}
    
    \begin{subfigure}[t]{0.43\linewidth}
        \includegraphics[width=\linewidth]{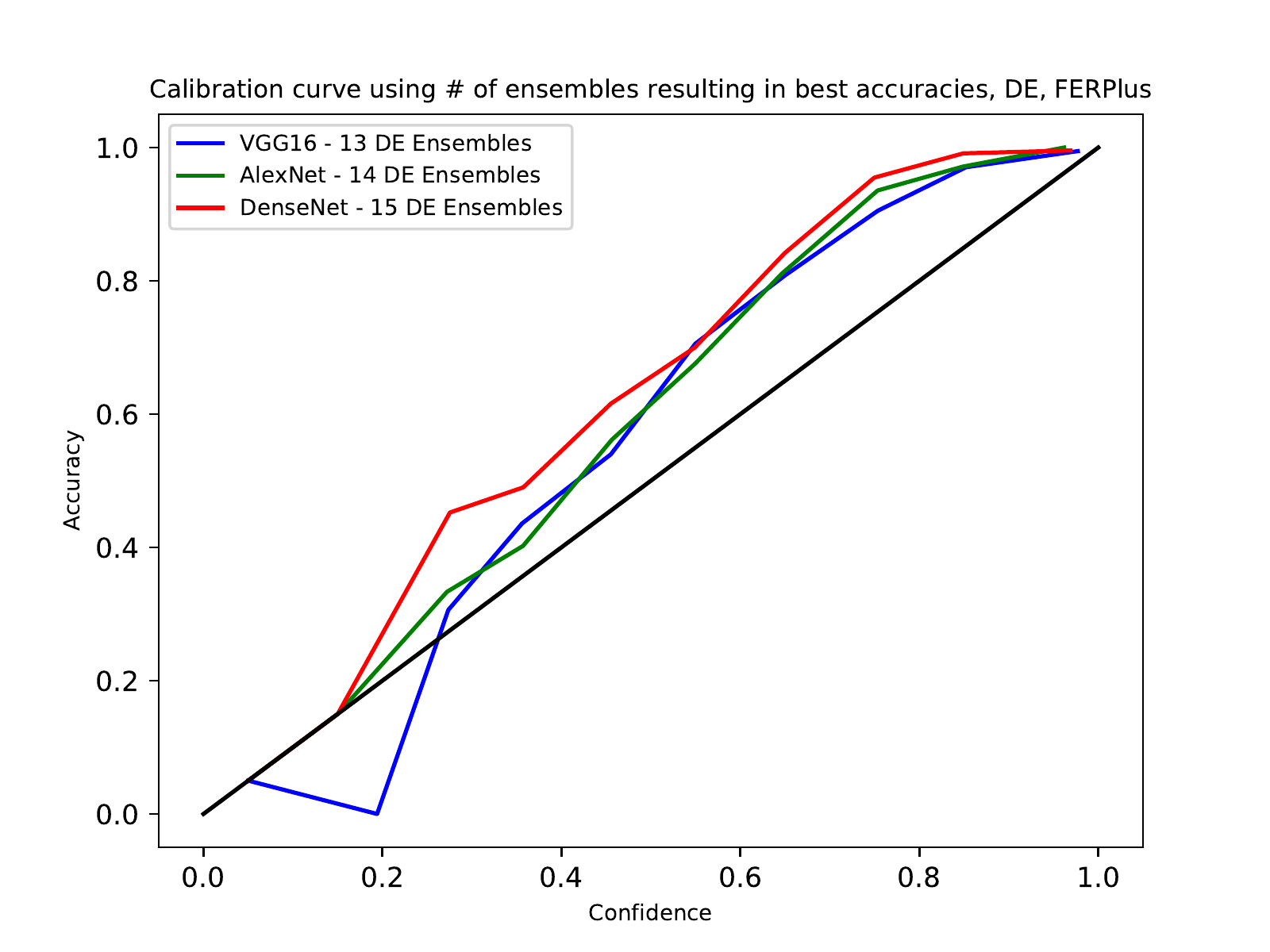}
        \caption{DE, \# of ensembles based on best classification accuracy}
    \end{subfigure}
    \begin{subfigure}[t]{0.43\linewidth}
        \includegraphics[width=\linewidth]{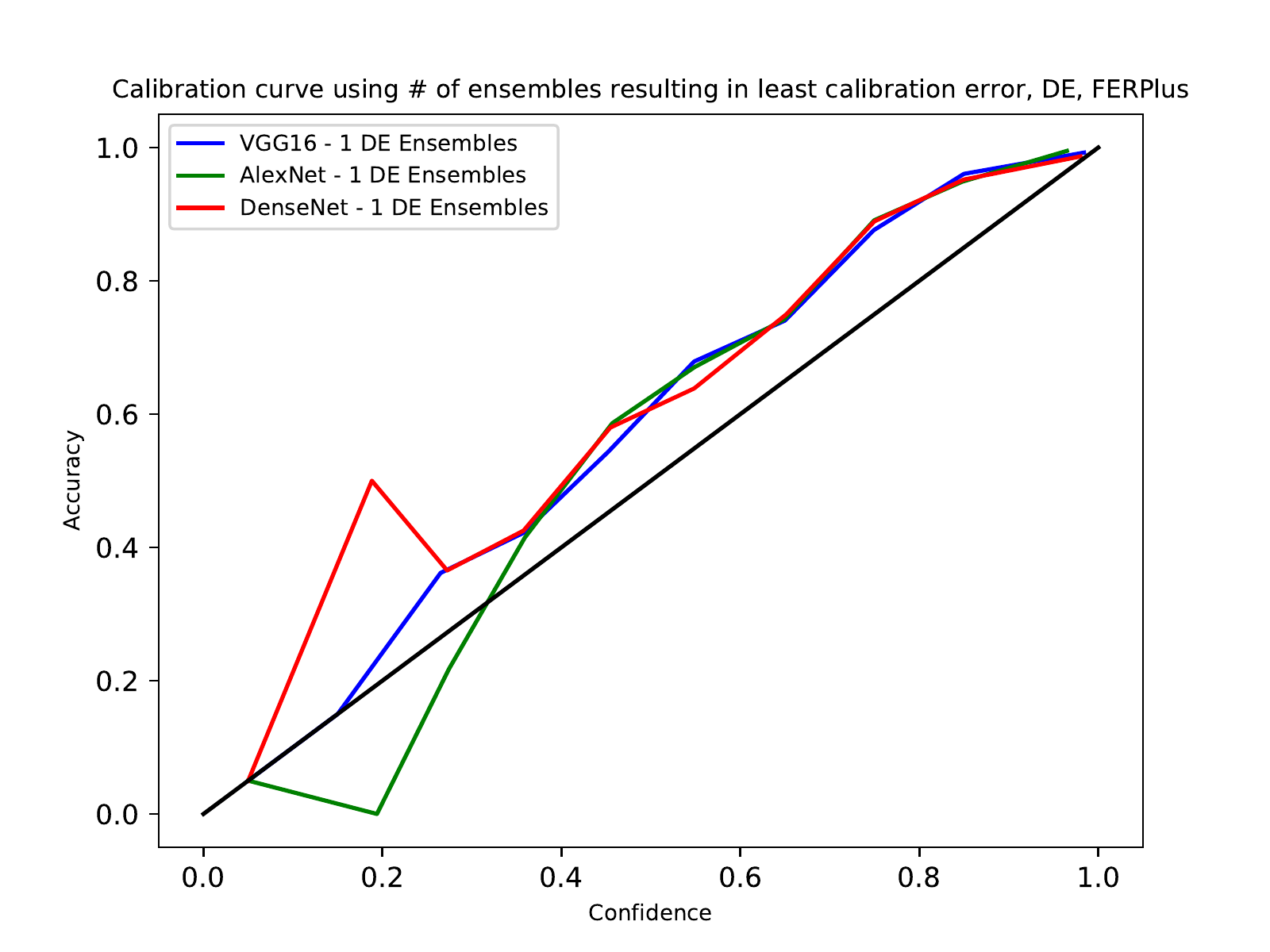}
        \caption{DE, \# of ensembles based on best calibration error}
        \label{fig:ferplus_calibration_f}
    \end{subfigure}
    \caption{Calibration curve for all methods in different models on FERPlus dataset using \# of samples resulting in best classification accuracy (left) or least calibration error (right).}
    
    \label{fig:ferplus_calibration}
\end{figure}

\clearpage
\section{Additional Results on Most Uncertain Images}

This section presents additional probability plots for Deep Ensembles across different models. One important conclusion that can be drawn from these plots is that the number of ensembles has a big influence on the output probabilities, and for task with high uncertainty such as facial emotion recognition, this can lead to very different class predictions as computed by taking the maximum probability.

\begin{figure}[h!]
    \centering
    \begin{subfigure}[b]{0.09\linewidth}
        \caption*{\textbf{Image}}
    \end{subfigure}
    \begin{subfigure}[b]{0.16\linewidth}
        \caption*{\textbf{Labels}}
    \end{subfigure}
    \begin{subfigure}[b]{0.16\linewidth}	
        \caption*{\textbf{1 Ens}}
    \end{subfigure}
    \begin{subfigure}[b]{0.16\linewidth}
        \caption*{\textbf{5 Ens}}
    \end{subfigure}
    \begin{subfigure}[b]{0.16\linewidth}
        \caption*{\textbf{10 Ens}}
    \end{subfigure}
    \begin{subfigure}[b]{0.16\linewidth}
        \caption*{\textbf{15 Ens}}
    \end{subfigure}
    \begin{subfigure}[b]{0.09\linewidth}
        \includegraphics[width=\linewidth]{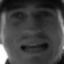}
        \caption*{Happiness}
    \end{subfigure}
    \begin{subfigure}[b]{0.17\linewidth}
        \includegraphics[width=\linewidth]{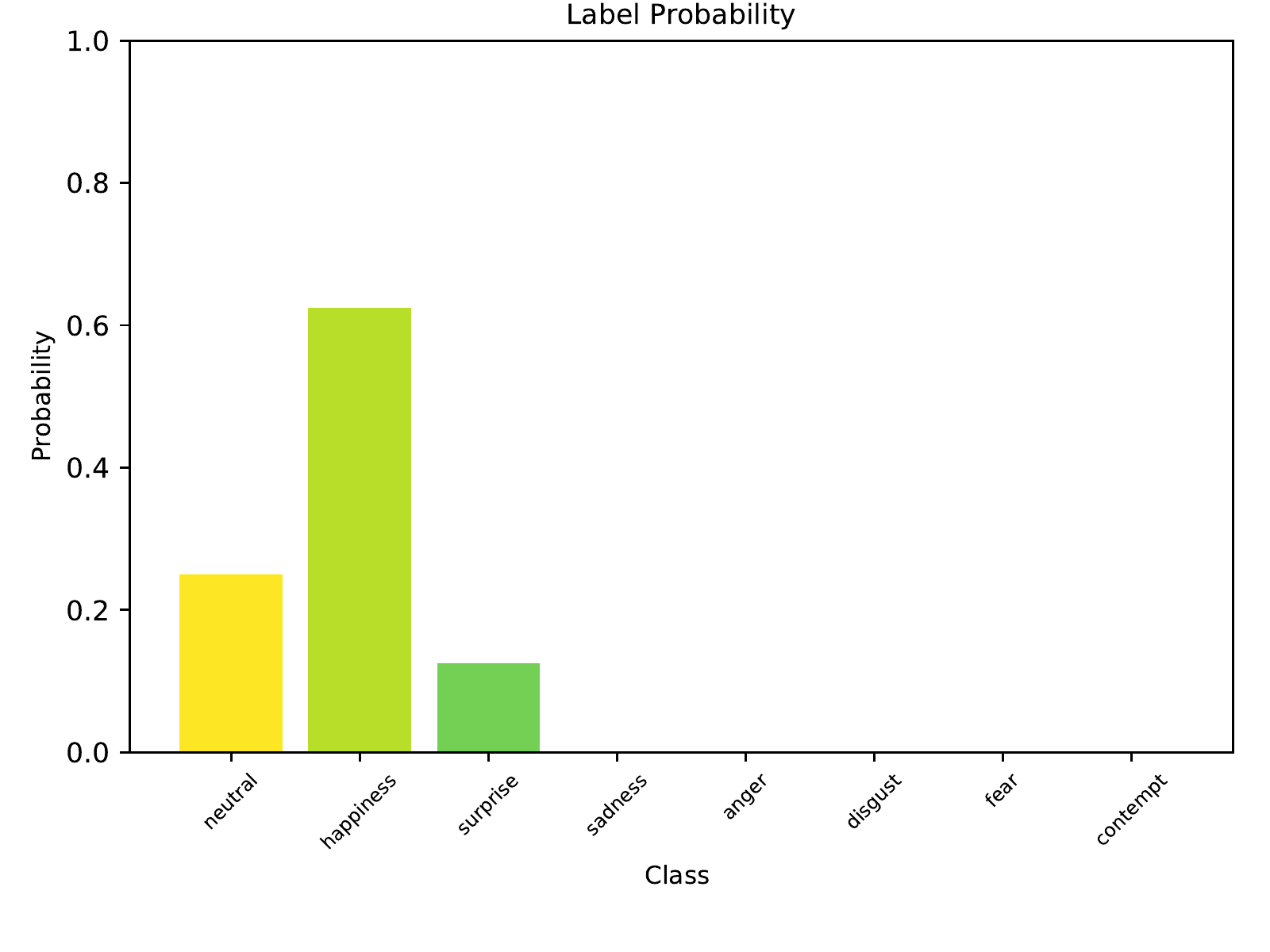}
        \caption*{True Label}
    \end{subfigure}
    \begin{subfigure}[b]{0.17\linewidth}
        \includegraphics[width=\linewidth]{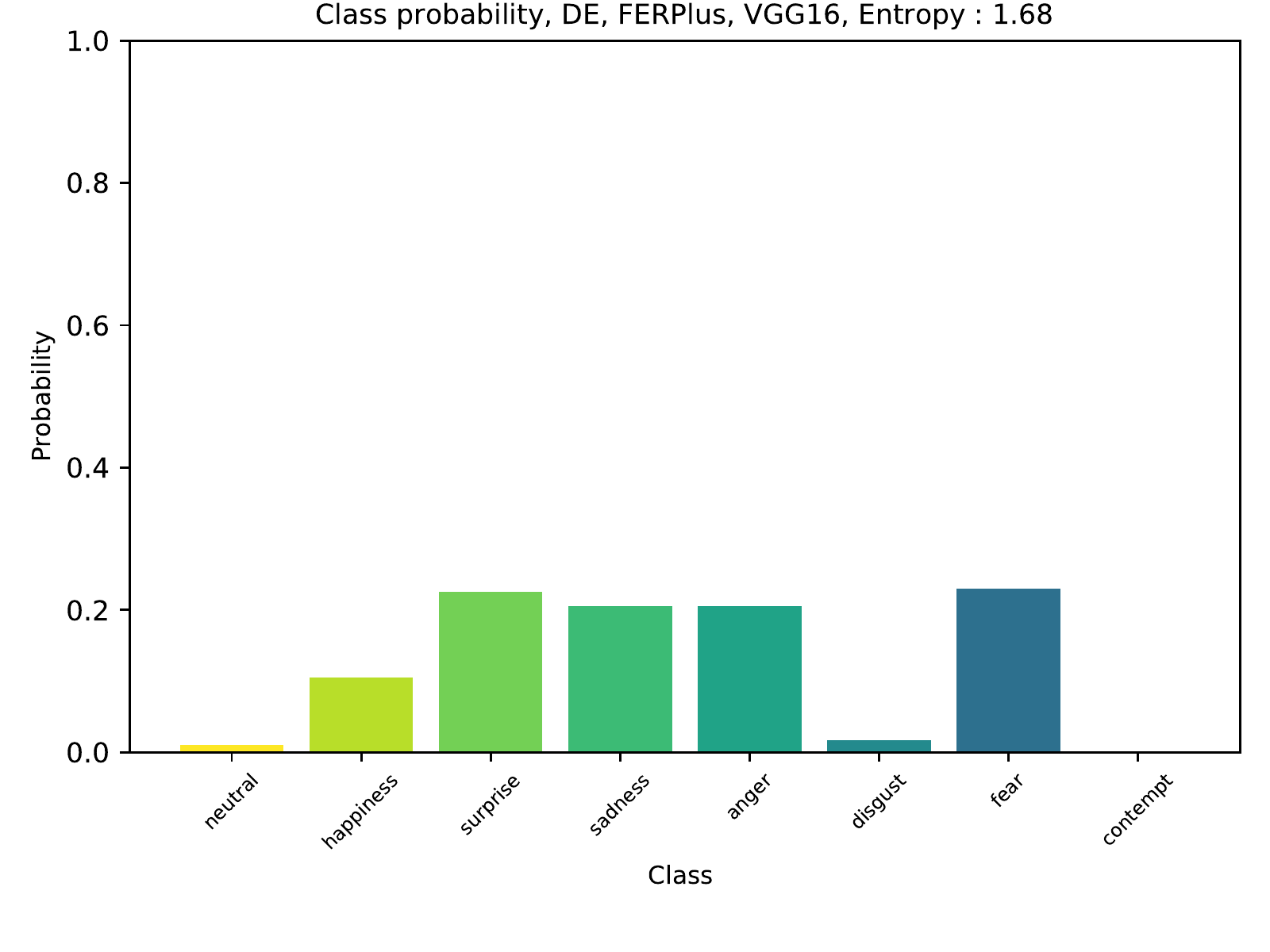}
        \caption*{Fear}
    \end{subfigure}
    \begin{subfigure}[b]{0.17\linewidth}
        \includegraphics[width=\linewidth]{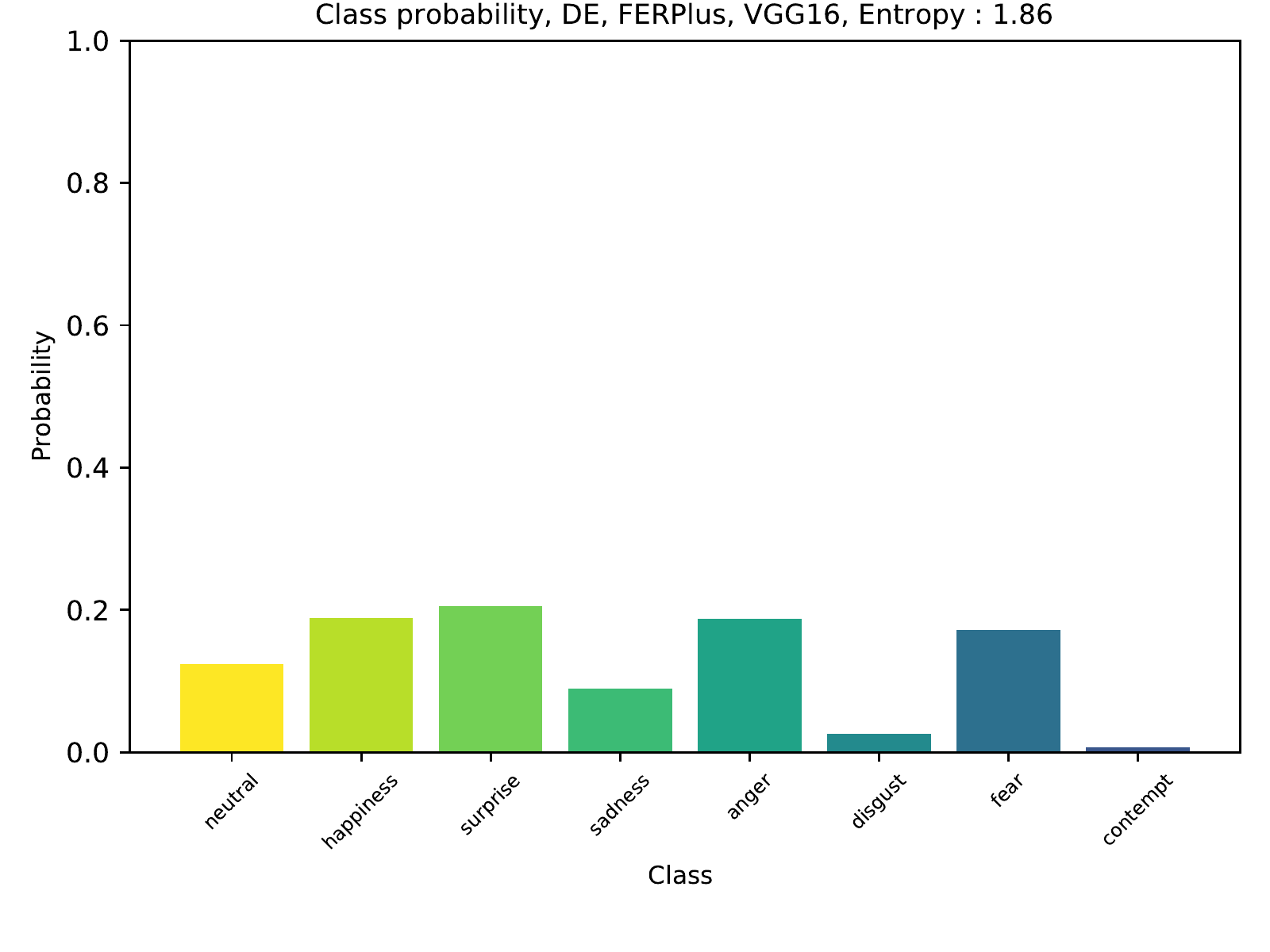}
        \caption*{Surprise}
    \end{subfigure}
    \begin{subfigure}[b]{0.17\linewidth}
        \includegraphics[width=\linewidth]{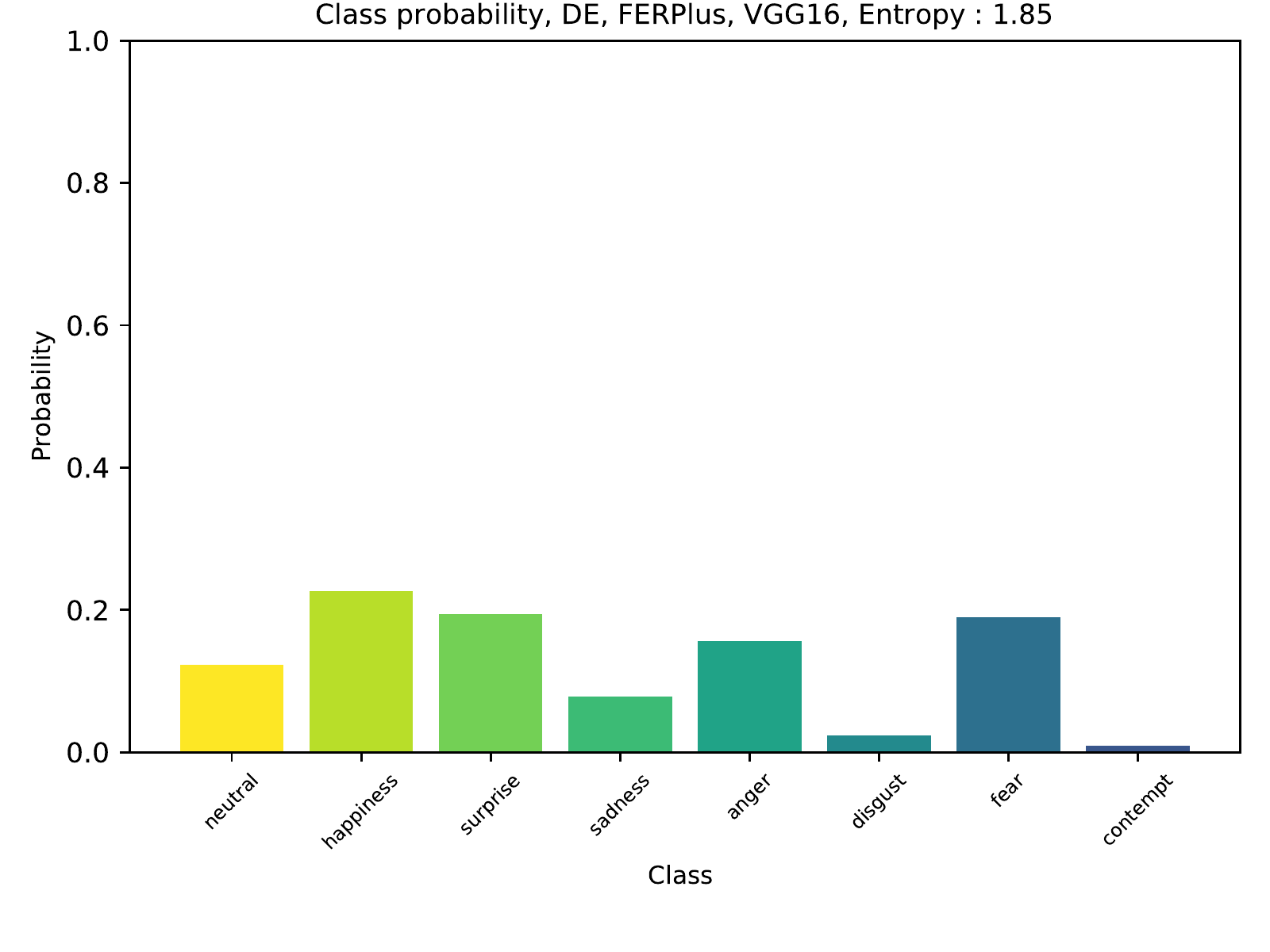}
        \caption*{Happiness}
    \end{subfigure}
    \begin{subfigure}[b]{0.17\linewidth}
        \includegraphics[width=\linewidth]{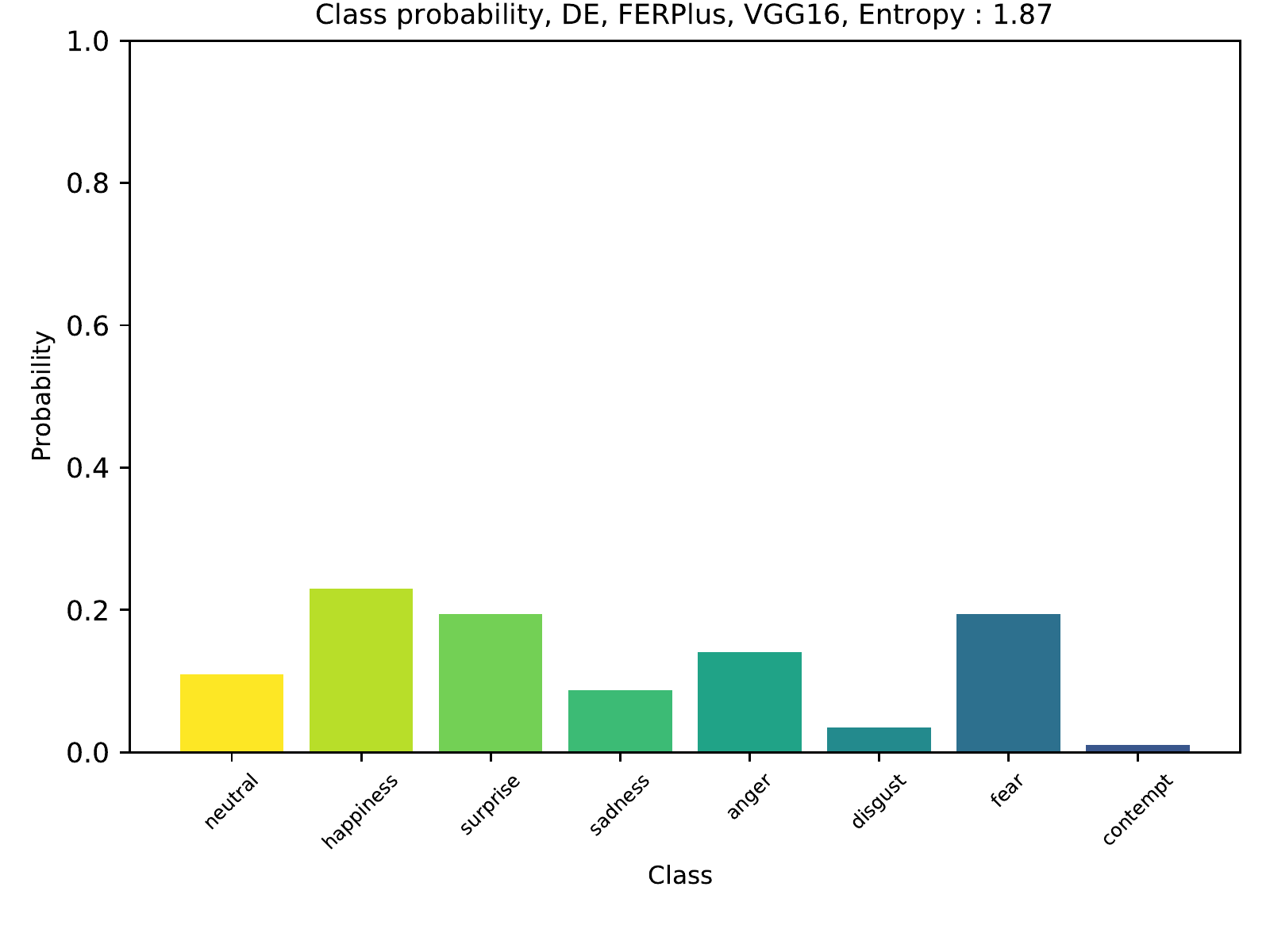}
        \caption*{Happiness}
    \end{subfigure}
    \begin{subfigure}[b]{0.09\linewidth}
        \includegraphics[width=\linewidth]{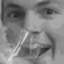}
        \caption*{Happiness}
    \end{subfigure}
    \begin{subfigure}[b]{0.17\linewidth}
        \includegraphics[width=\linewidth]{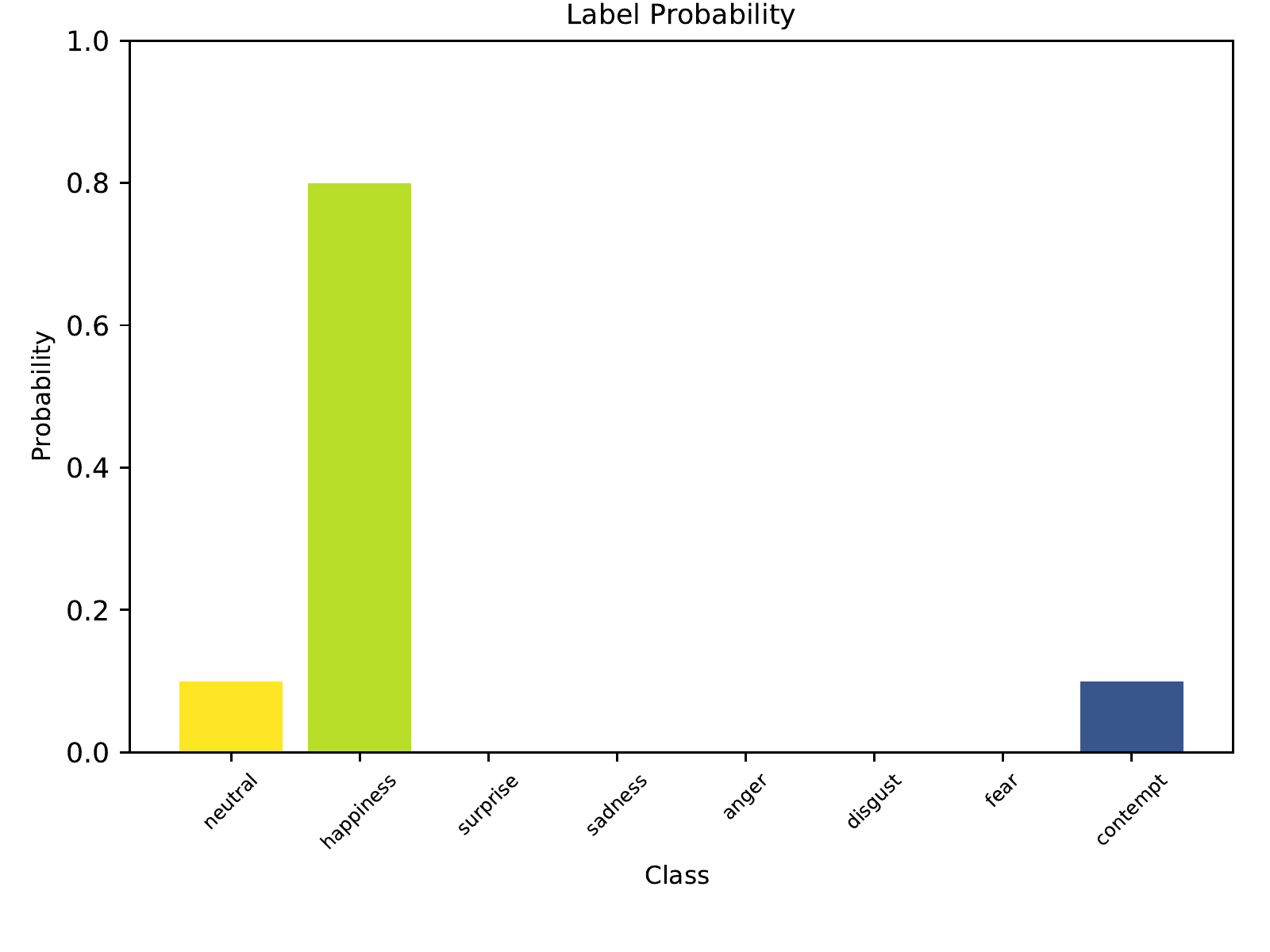}
        \caption*{True Label}
    \end{subfigure}
    \begin{subfigure}[b]{0.17\linewidth}
        \includegraphics[width=\linewidth]{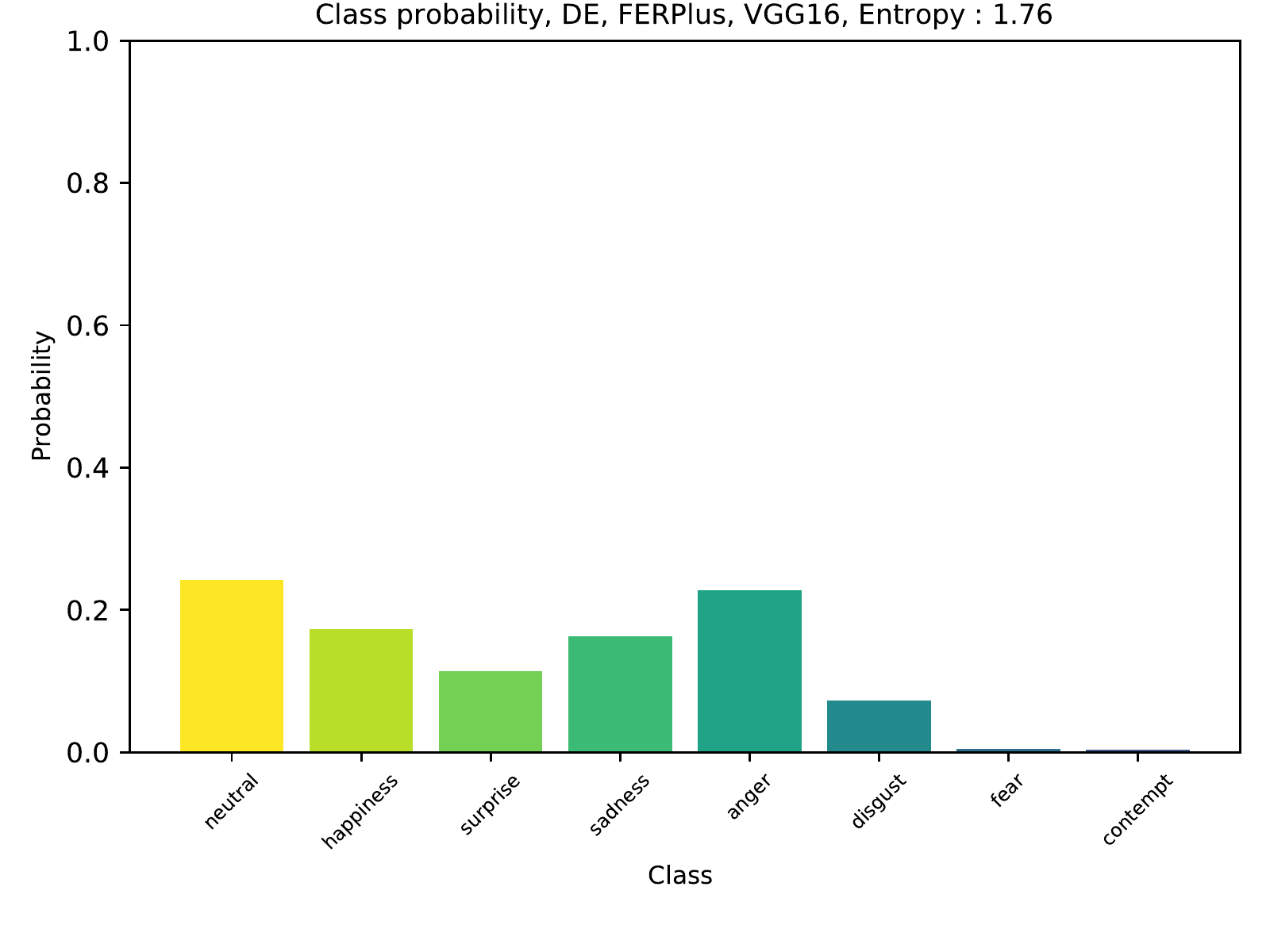}
        \caption*{Neutral}
    \end{subfigure}
    \begin{subfigure}[b]{0.17\linewidth}
        \includegraphics[width=\linewidth]{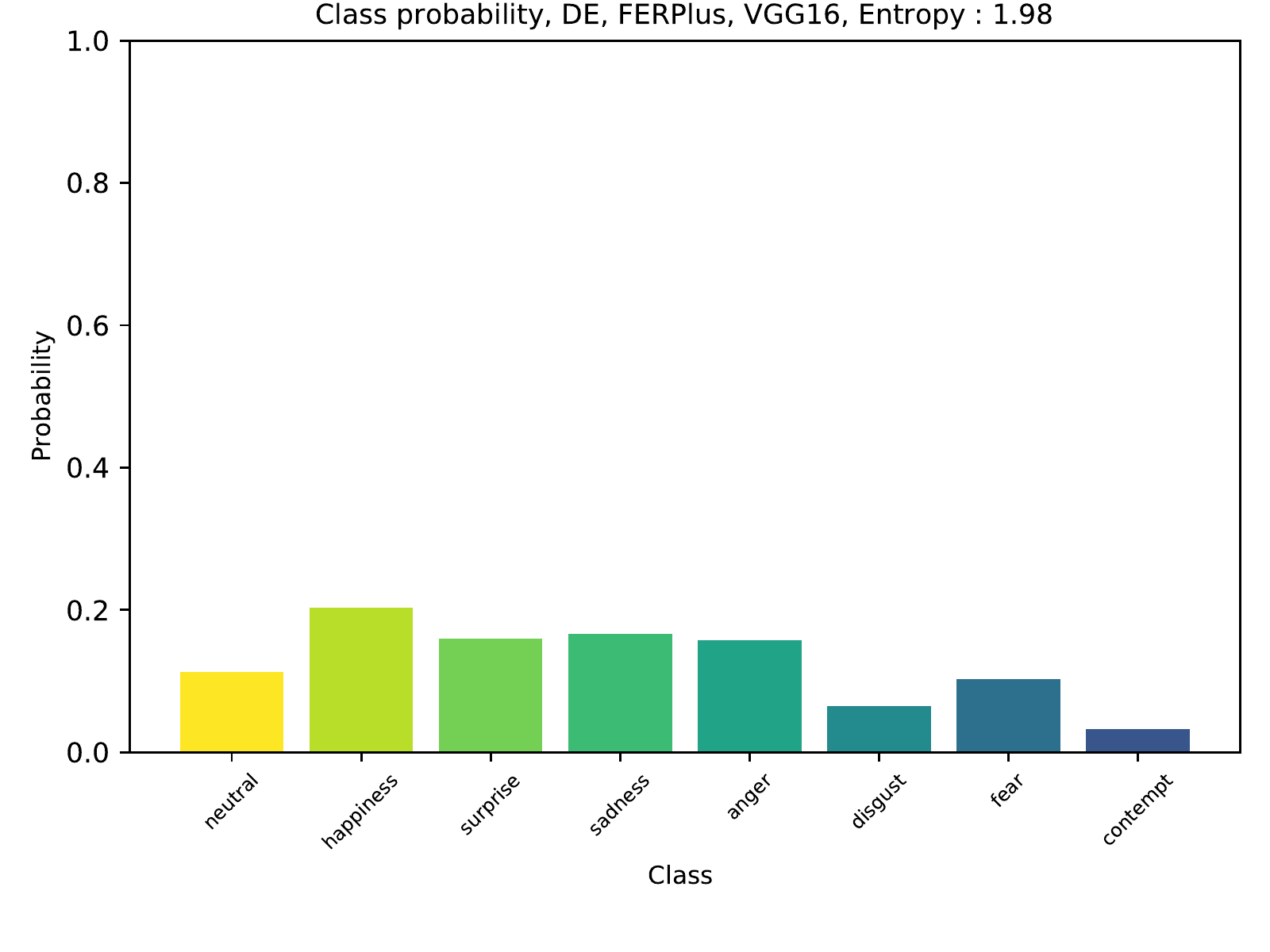}
        \caption*{Happiness}
    \end{subfigure}
    \begin{subfigure}[b]{0.17\linewidth}
        \includegraphics[width=\linewidth]{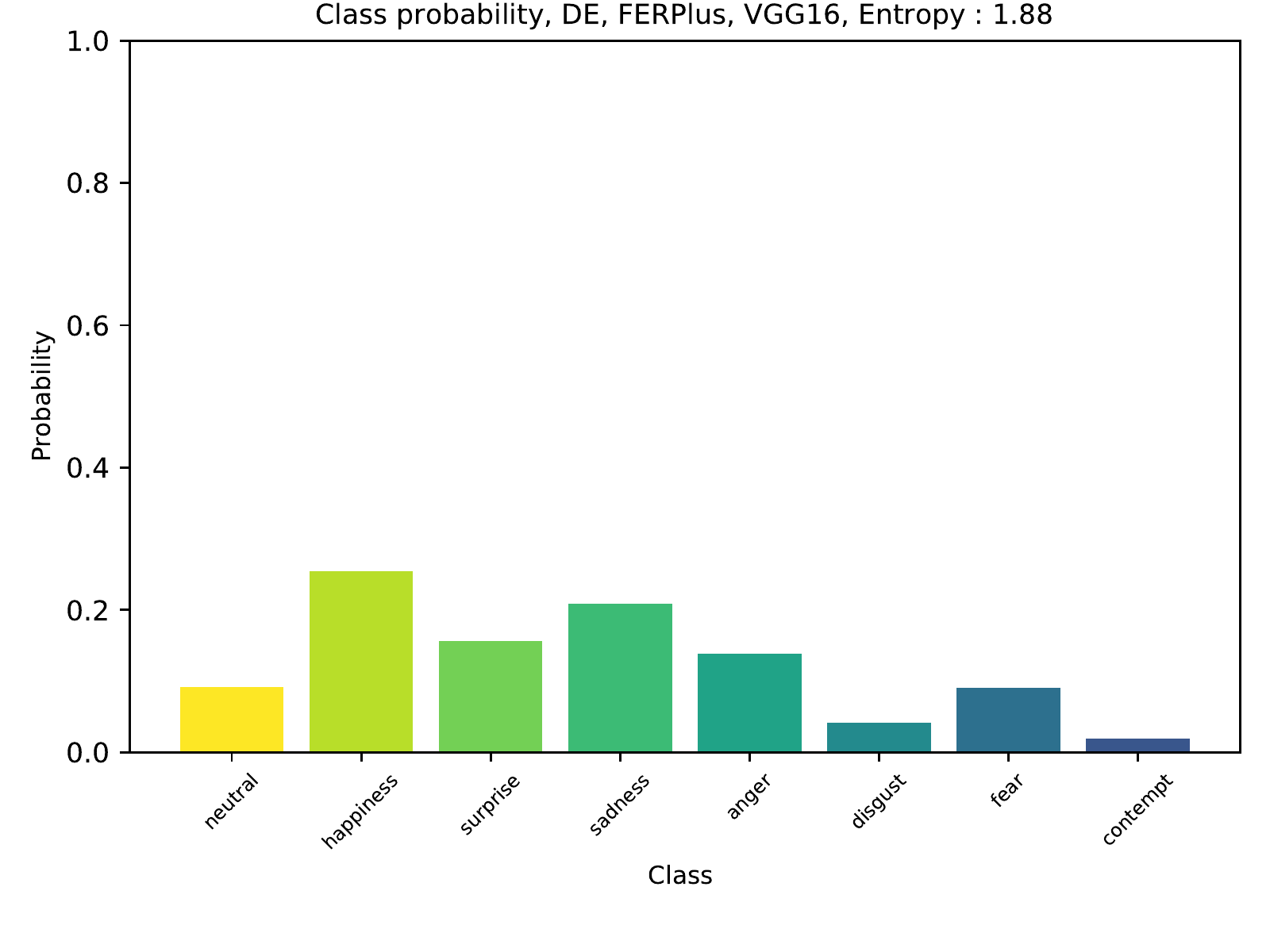}
        \caption*{Happiness}
    \end{subfigure}
    \begin{subfigure}[b]{0.17\linewidth}
        \includegraphics[width=\linewidth]{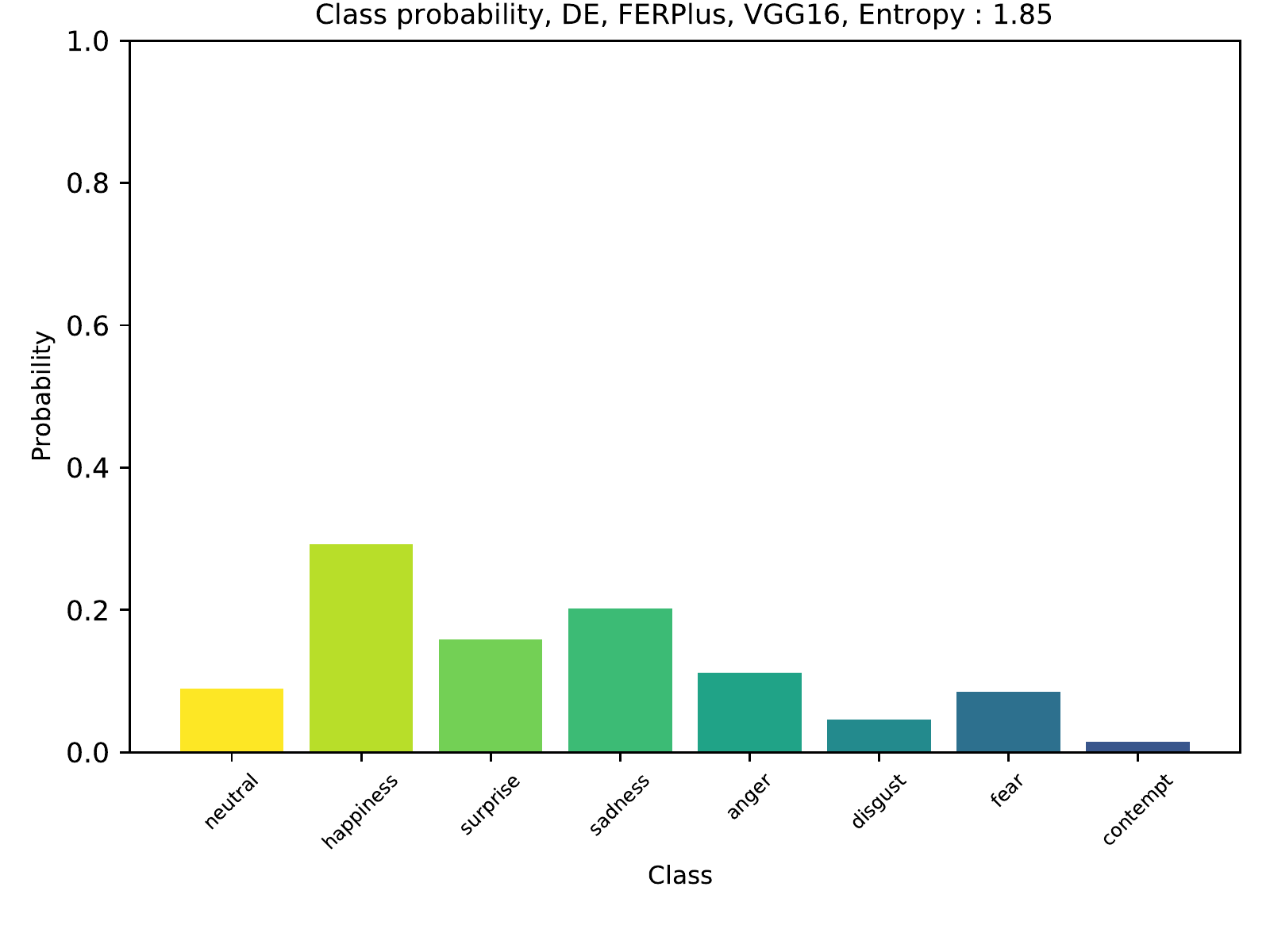}
        \caption*{Happiness}
    \end{subfigure}
    
    \begin{subfigure}[b]{0.09\linewidth}
        \includegraphics[width=\linewidth]{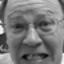}
        \vspace*{0.1em}
        \caption*{Anger}
    \end{subfigure}
    \begin{subfigure}[b]{0.17\linewidth}
        \includegraphics[width=\linewidth]{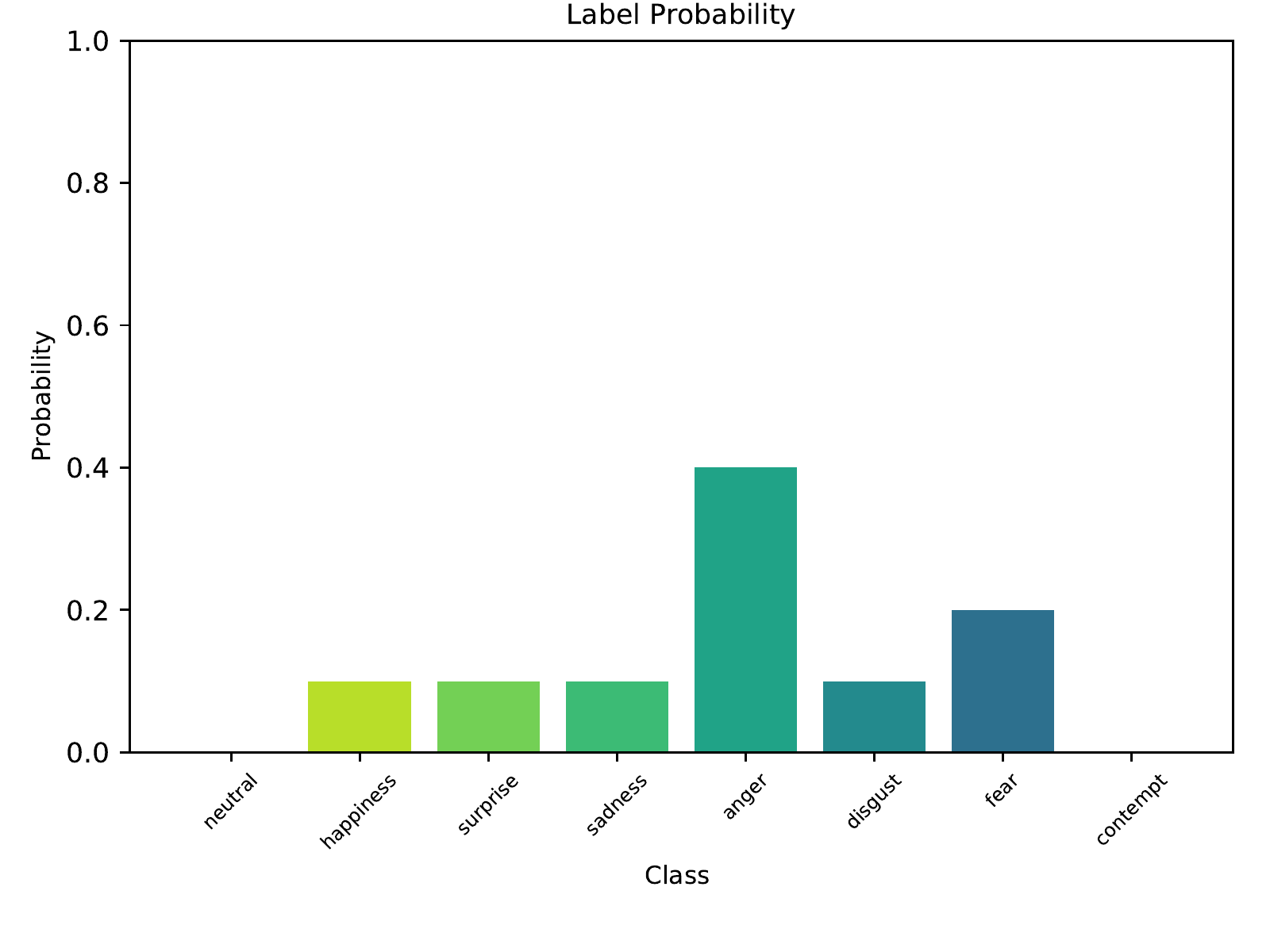}
        \caption*{True Label}
    \end{subfigure}
    \begin{subfigure}[b]{0.17\linewidth}
        \includegraphics[width=\linewidth]{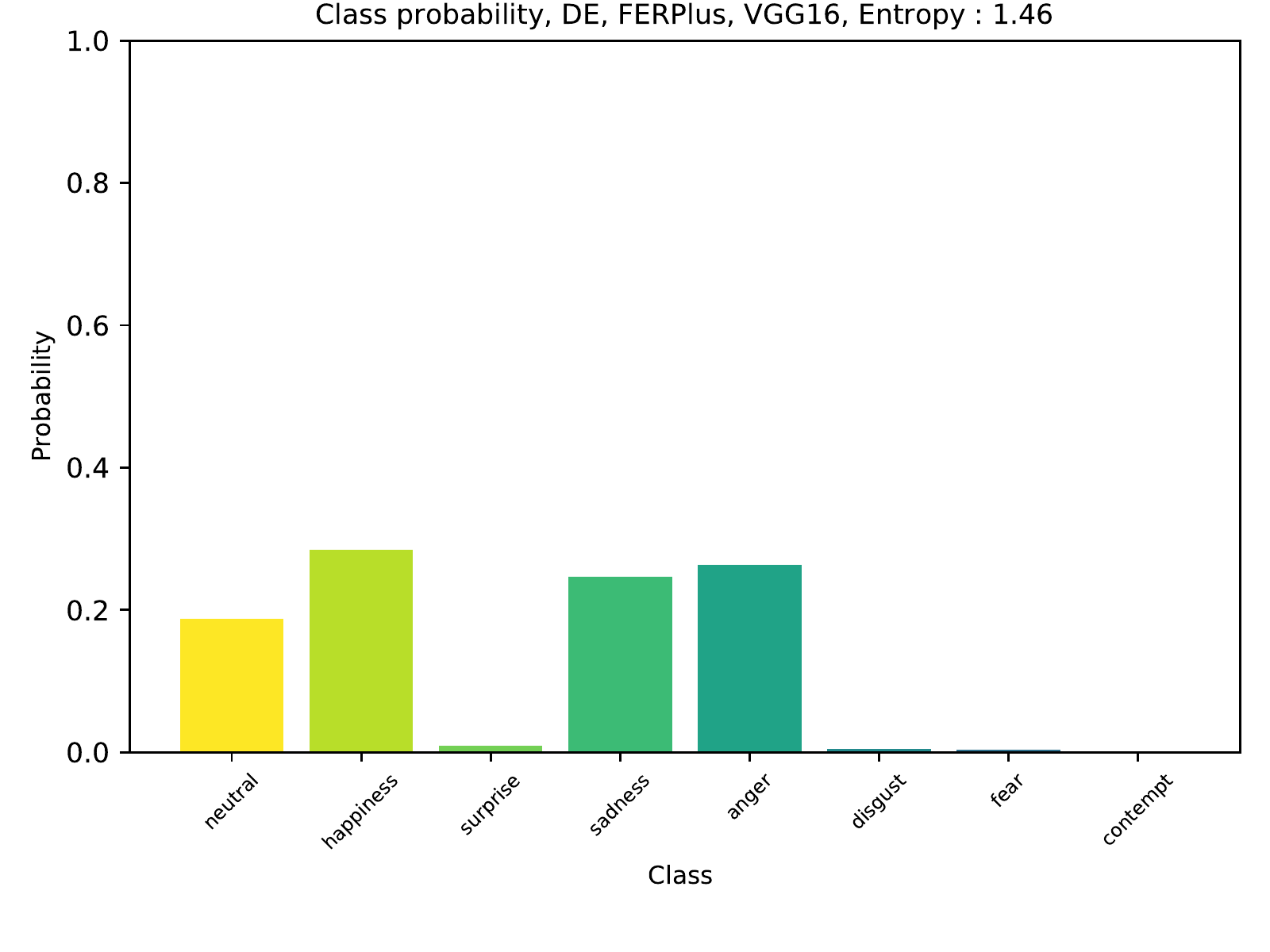}
        \caption*{Happiness}
    \end{subfigure}
    \begin{subfigure}[b]{0.17\linewidth}
        \includegraphics[width=\linewidth]{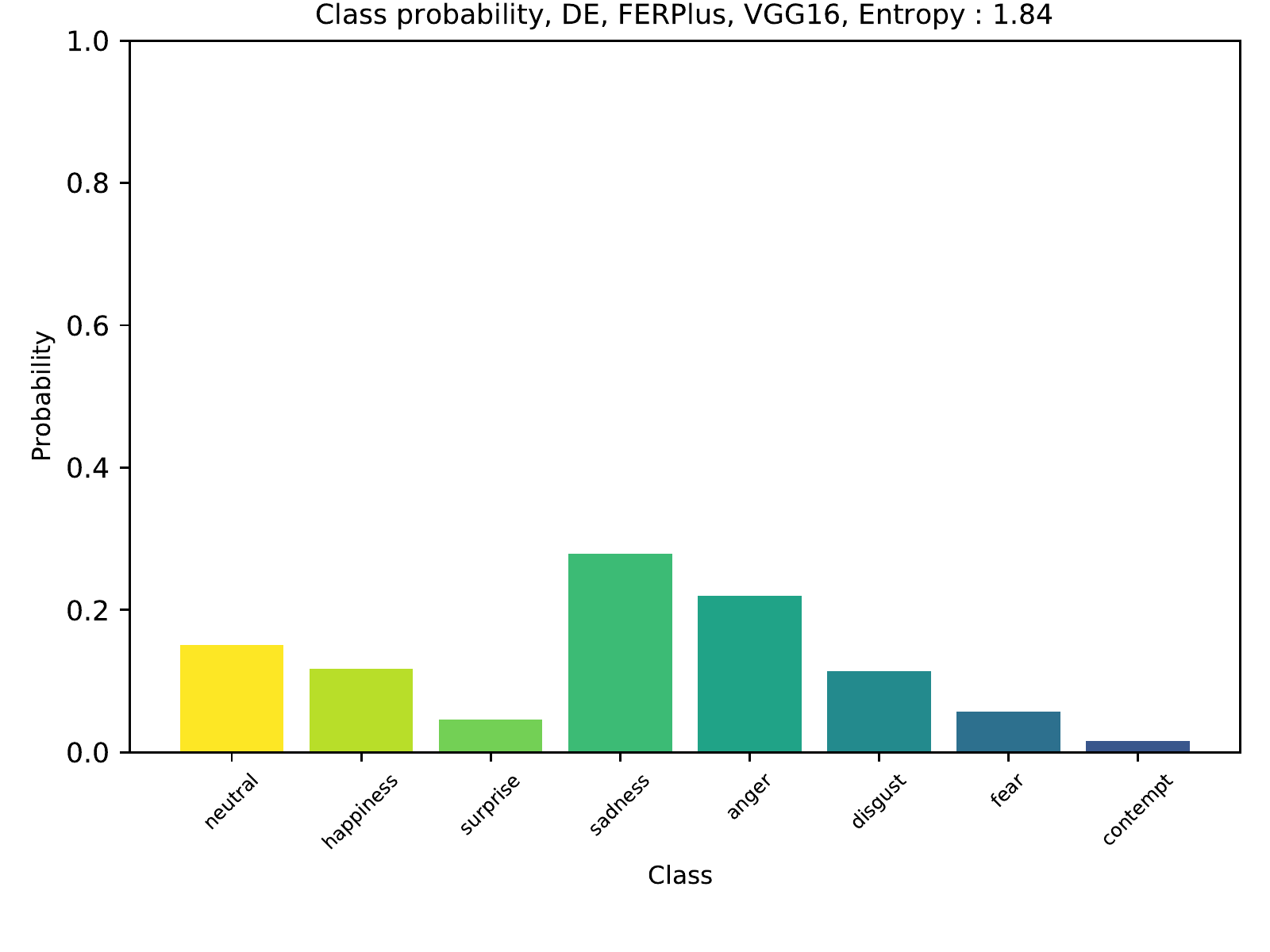}
        \caption*{Sadness}
    \end{subfigure}
    \begin{subfigure}[b]{0.17\linewidth}
        \includegraphics[width=\linewidth]{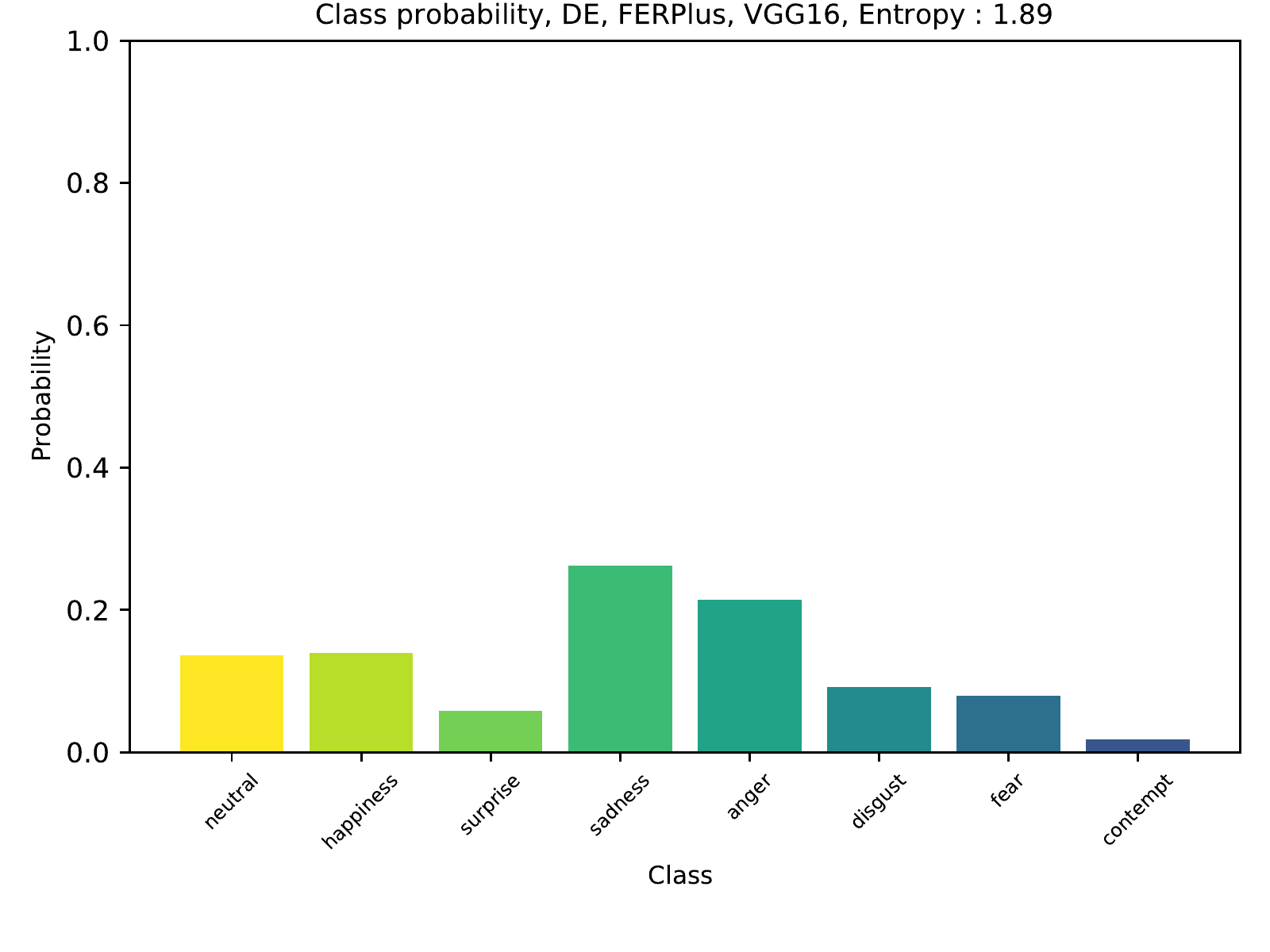}
        \caption*{Sadness}
    \end{subfigure}
    \begin{subfigure}[b]{0.17\linewidth}
        \includegraphics[width=\linewidth]{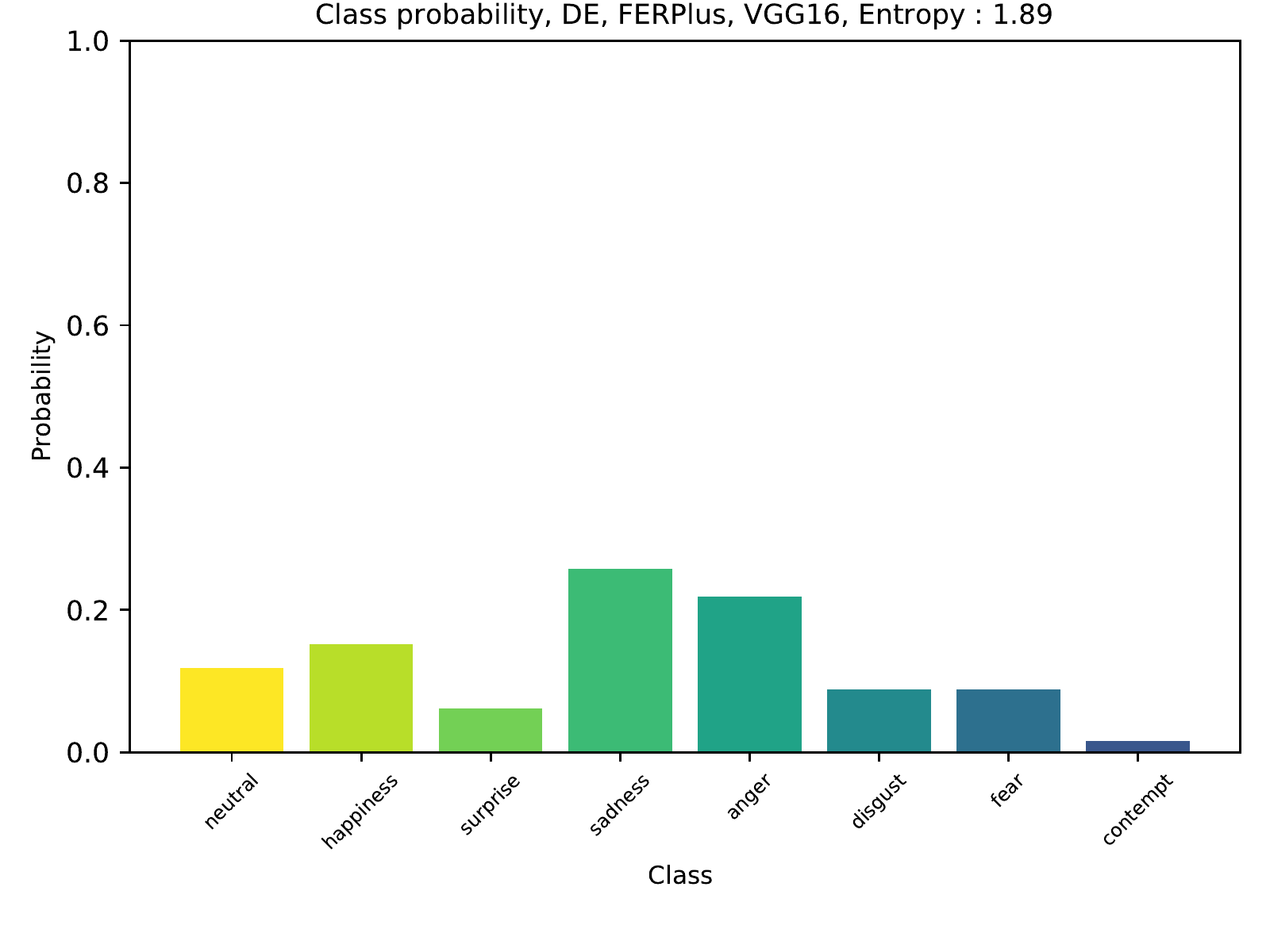}
        \caption*{Sadness}
    \end{subfigure}
    
    \begin{subfigure}[b]{0.09\linewidth}
        \includegraphics[width=\linewidth]{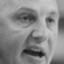}
        \vspace*{0.1em}
        \caption*{Neutral}
    \end{subfigure}
    \begin{subfigure}[b]{0.17\linewidth}
        \includegraphics[width=\linewidth]{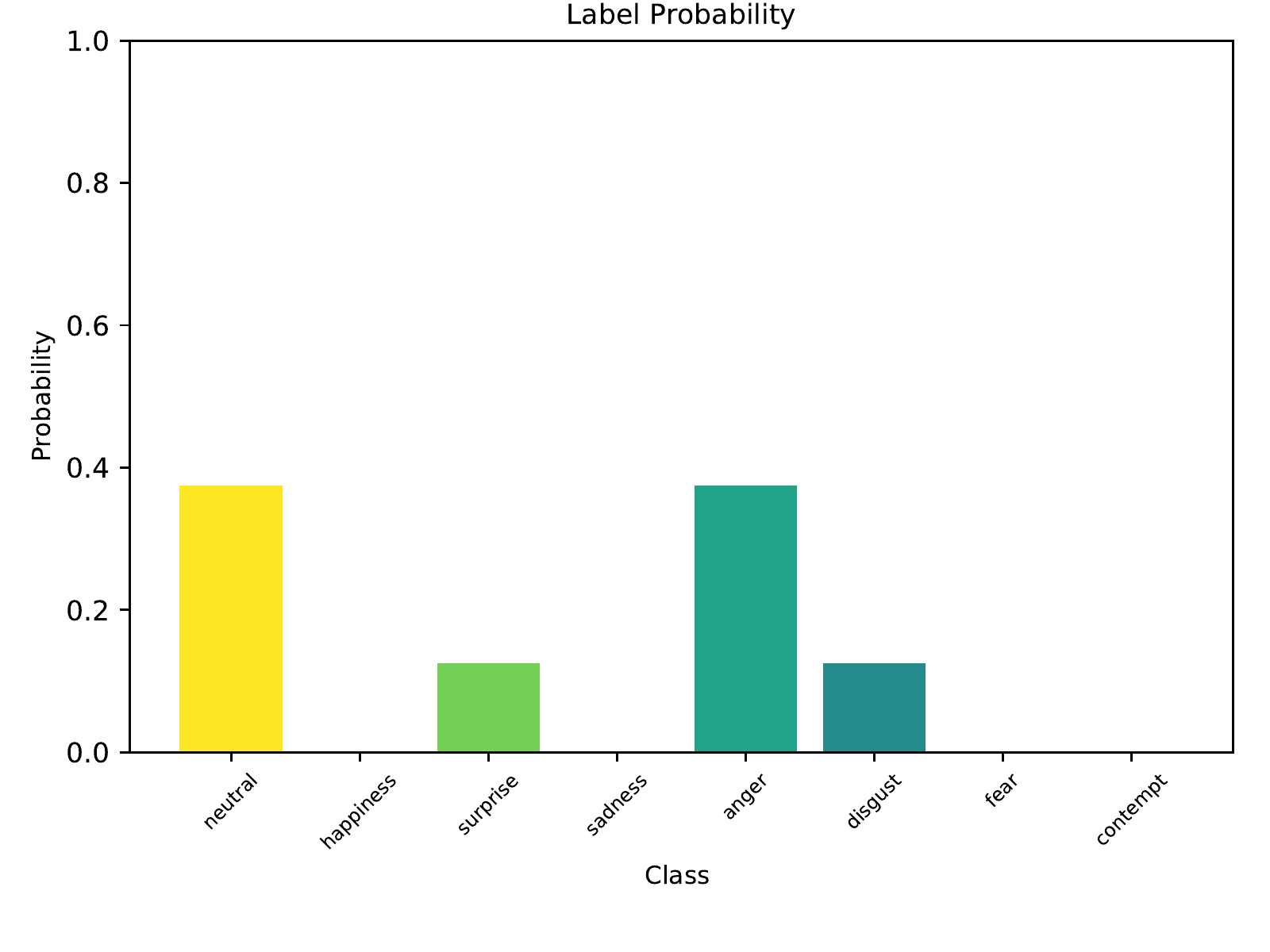}
        \caption*{True Label}
    \end{subfigure}
    \begin{subfigure}[b]{0.17\linewidth}
        \includegraphics[width=\linewidth]{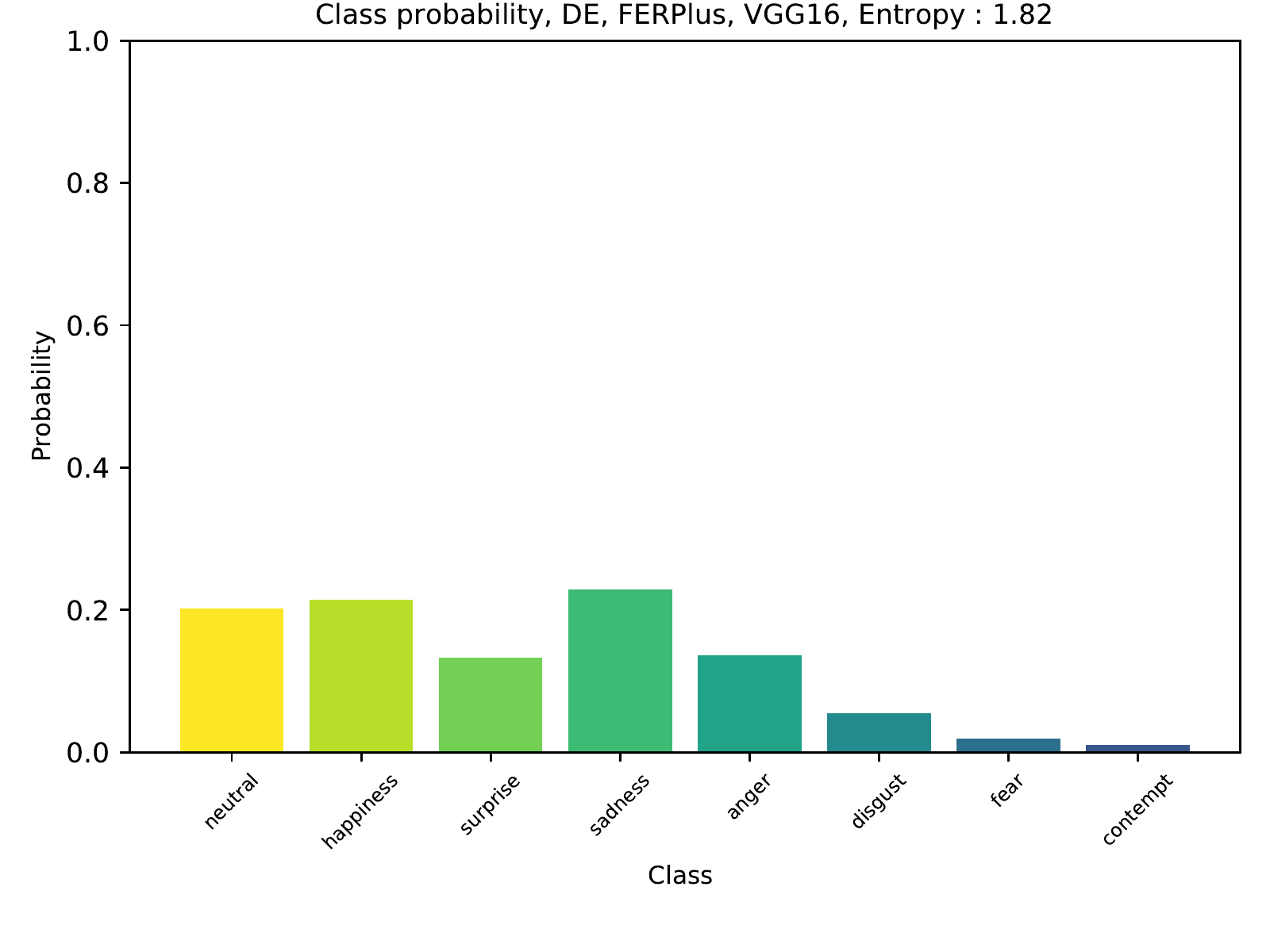}
        \caption*{Sadness}
    \end{subfigure}
    \begin{subfigure}[b]{0.17\linewidth}
        \includegraphics[width=\linewidth]{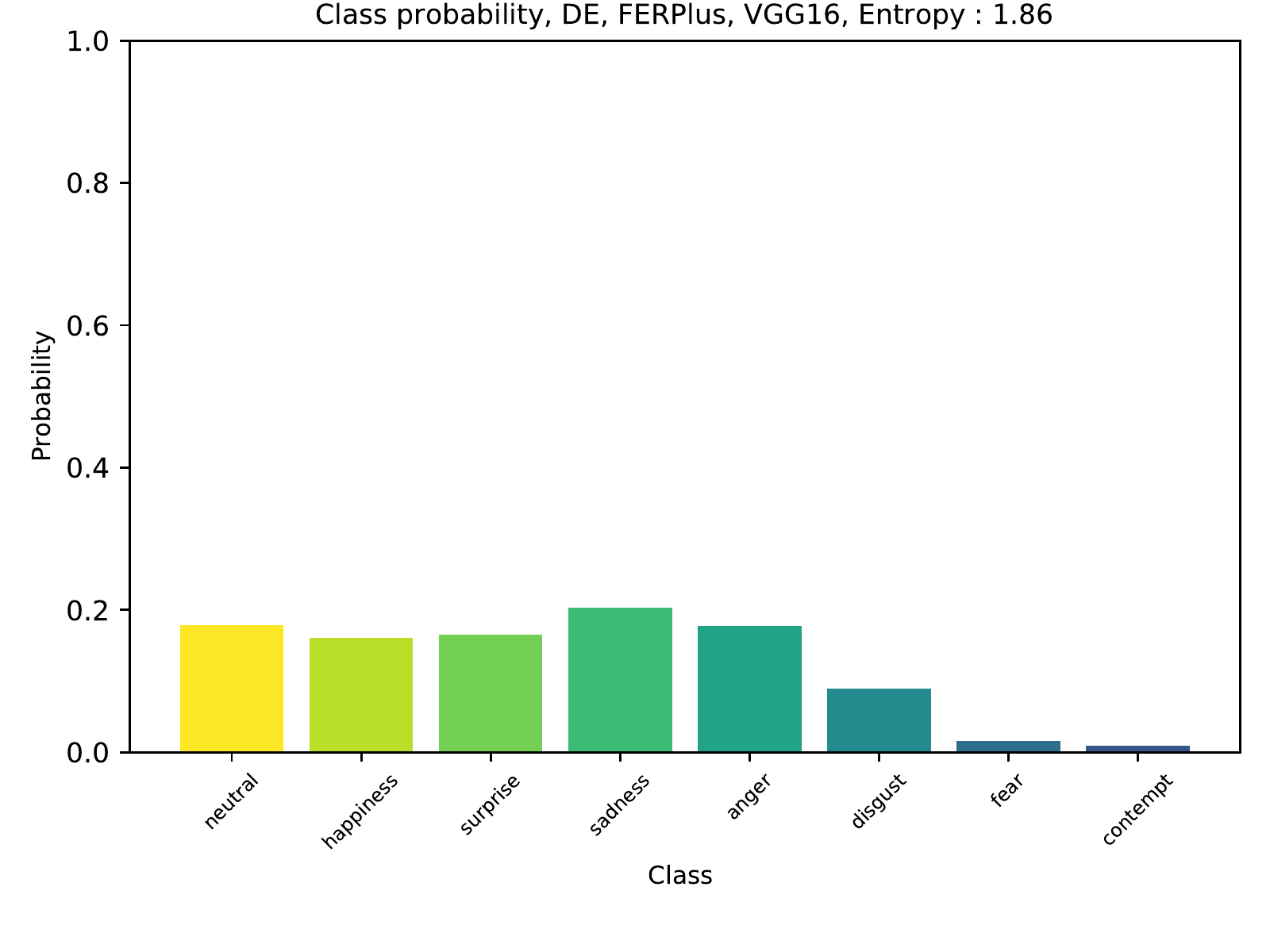}
        \caption*{Sadness}
    \end{subfigure}
    \begin{subfigure}[b]{0.17\linewidth}
        \includegraphics[width=\linewidth]{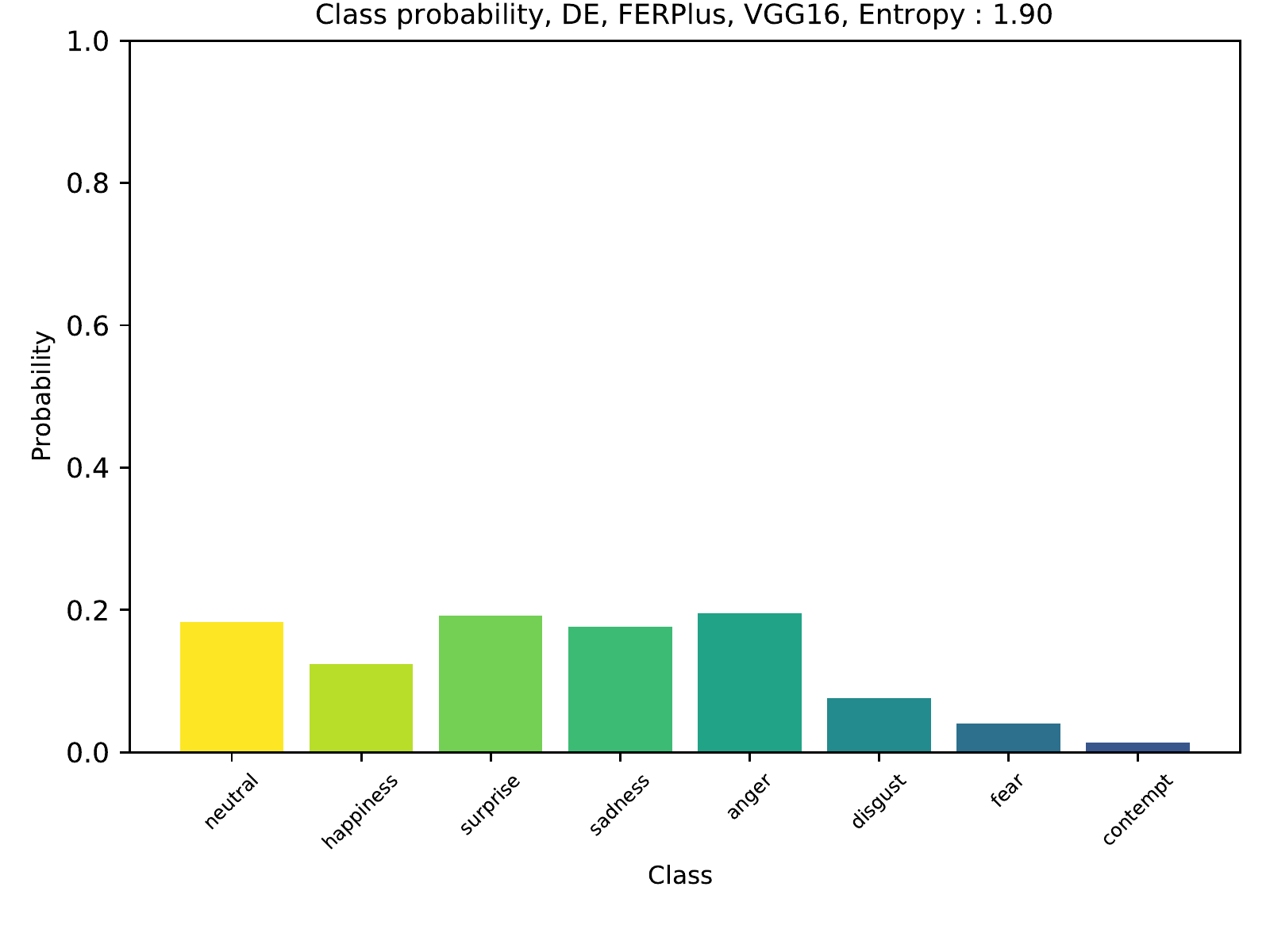}
        \caption*{Anger}
    \end{subfigure}
    \begin{subfigure}[b]{0.17\linewidth}
        \includegraphics[width=\linewidth]{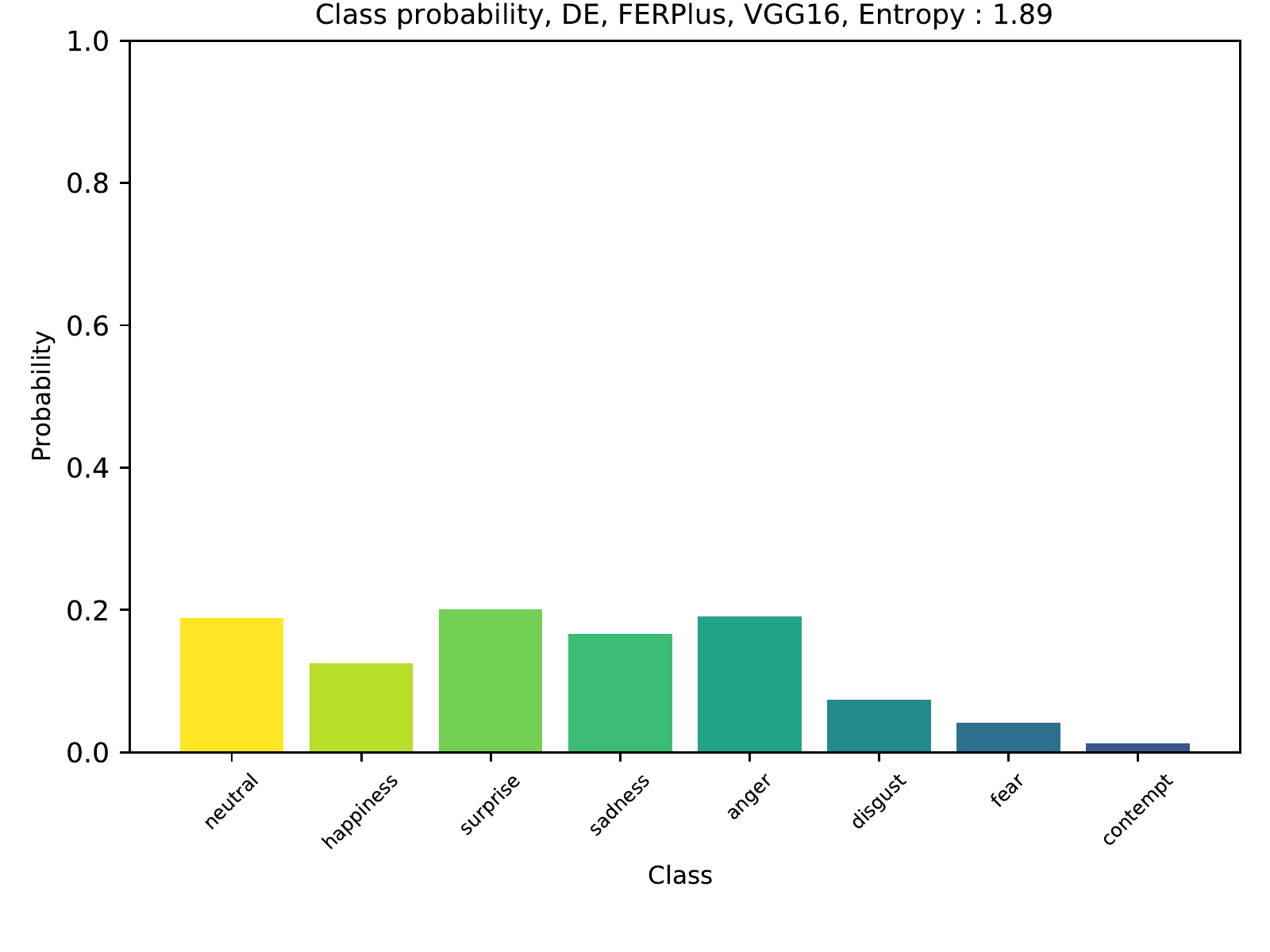}
        \caption*{Surprise}
    \end{subfigure}
    
    \begin{subfigure}[b]{0.09\linewidth}
        \includegraphics[width=\linewidth]{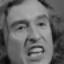}
        \vspace*{0.1em}
        \caption*{Anger}
    \end{subfigure}
    \begin{subfigure}[b]{0.17\linewidth}
        \includegraphics[width=\linewidth]{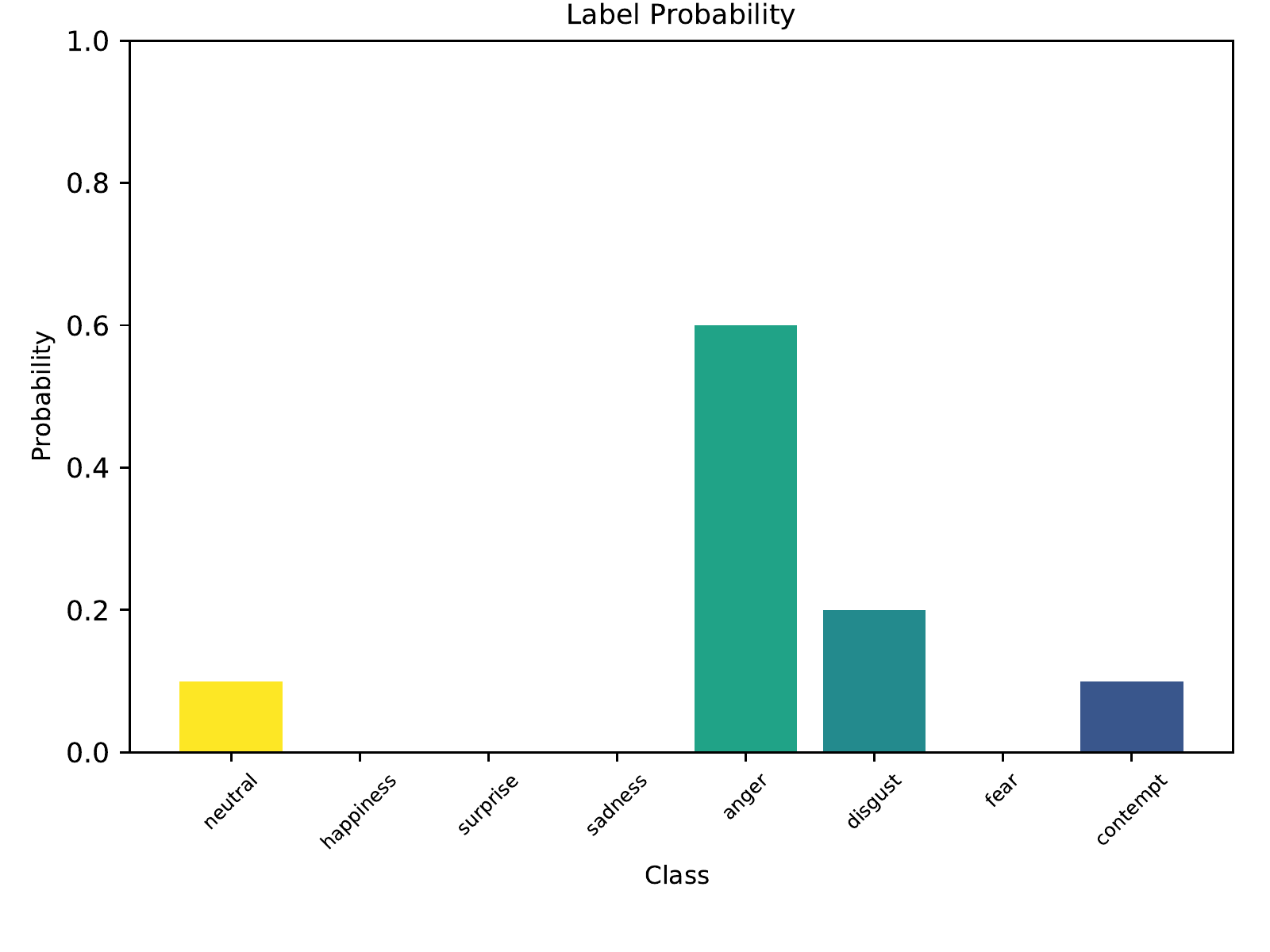}
        \caption*{True Label}
    \end{subfigure}
    \begin{subfigure}[b]{0.17\linewidth}
        \includegraphics[width=\linewidth]{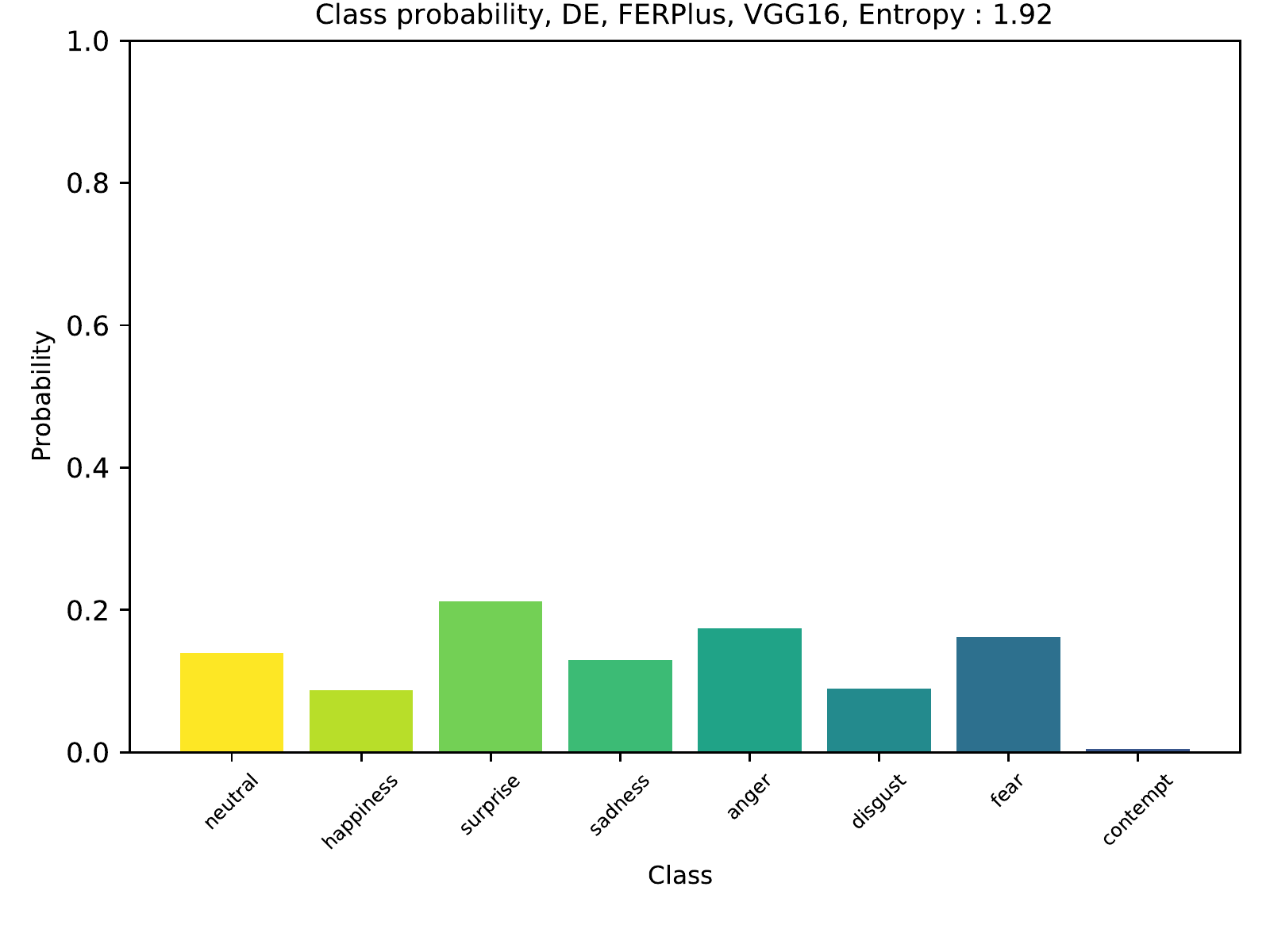}
        \caption*{Surprise}
    \end{subfigure}
    \begin{subfigure}[b]{0.17\linewidth}
        \includegraphics[width=\linewidth]{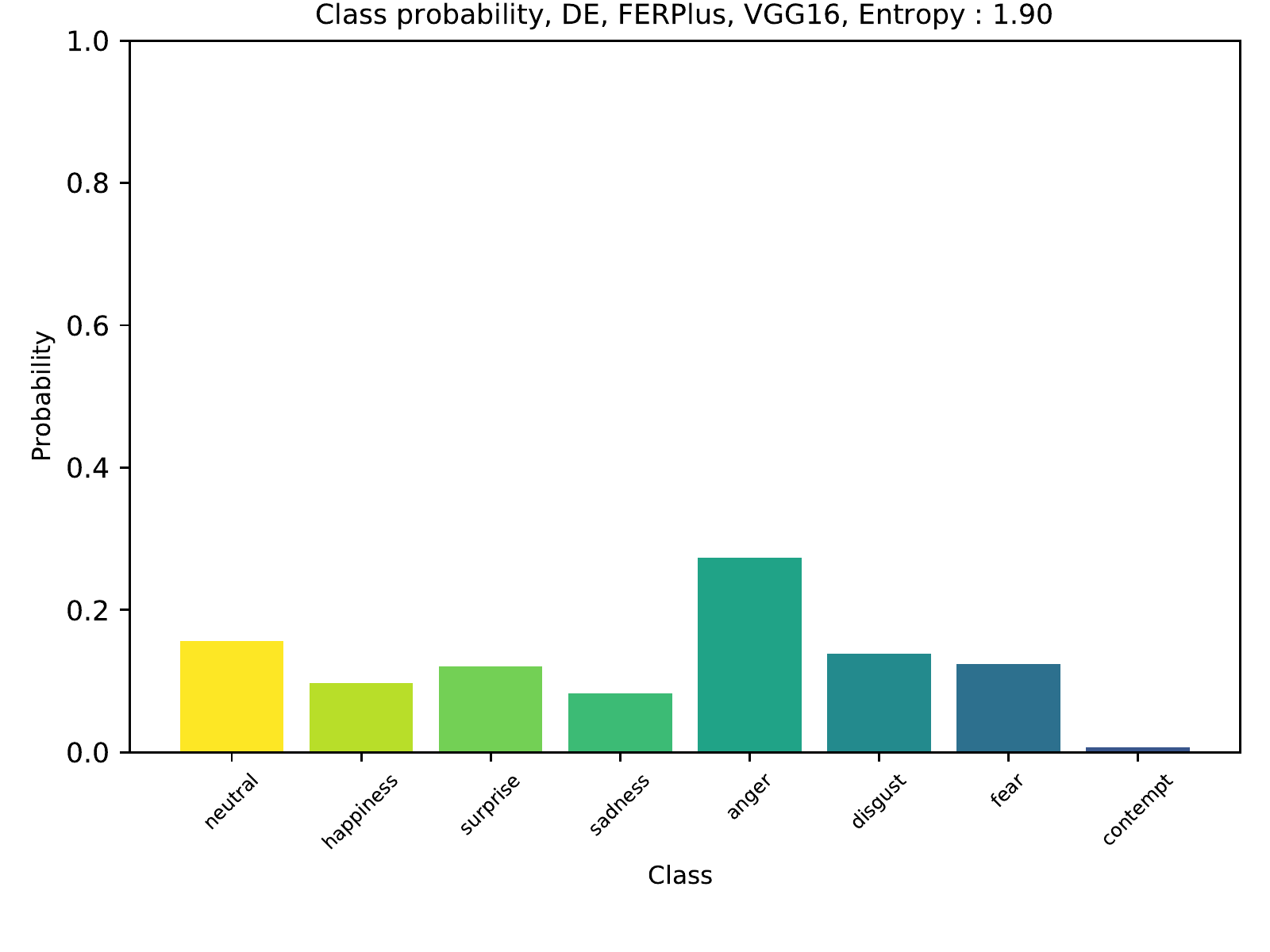}
        \caption*{Anger}
    \end{subfigure}
    \begin{subfigure}[b]{0.17\linewidth}
        \includegraphics[width=\linewidth]{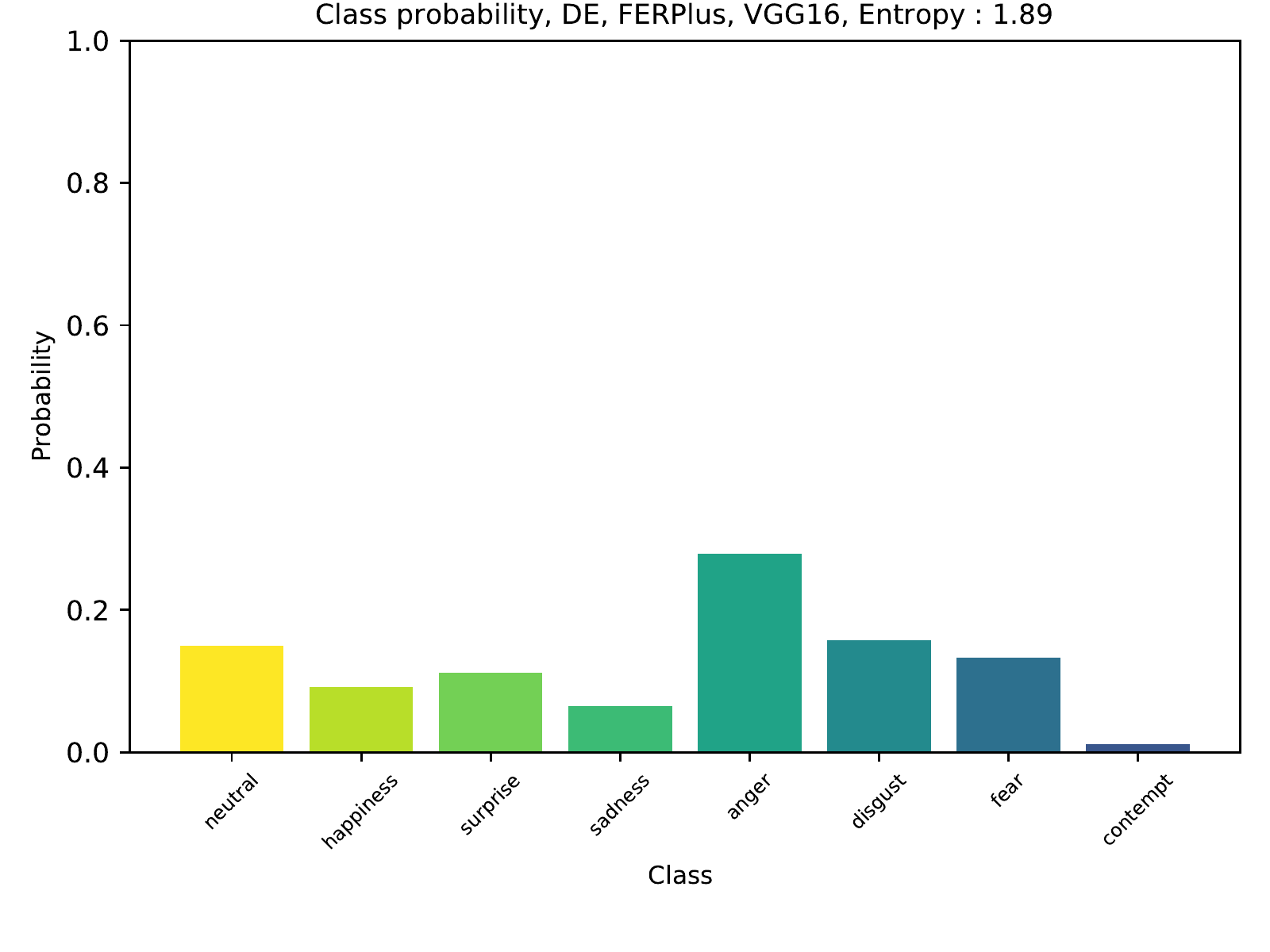}
        \caption*{Anger}
    \end{subfigure}
    \begin{subfigure}[b]{0.17\linewidth}
        \includegraphics[width=\linewidth]{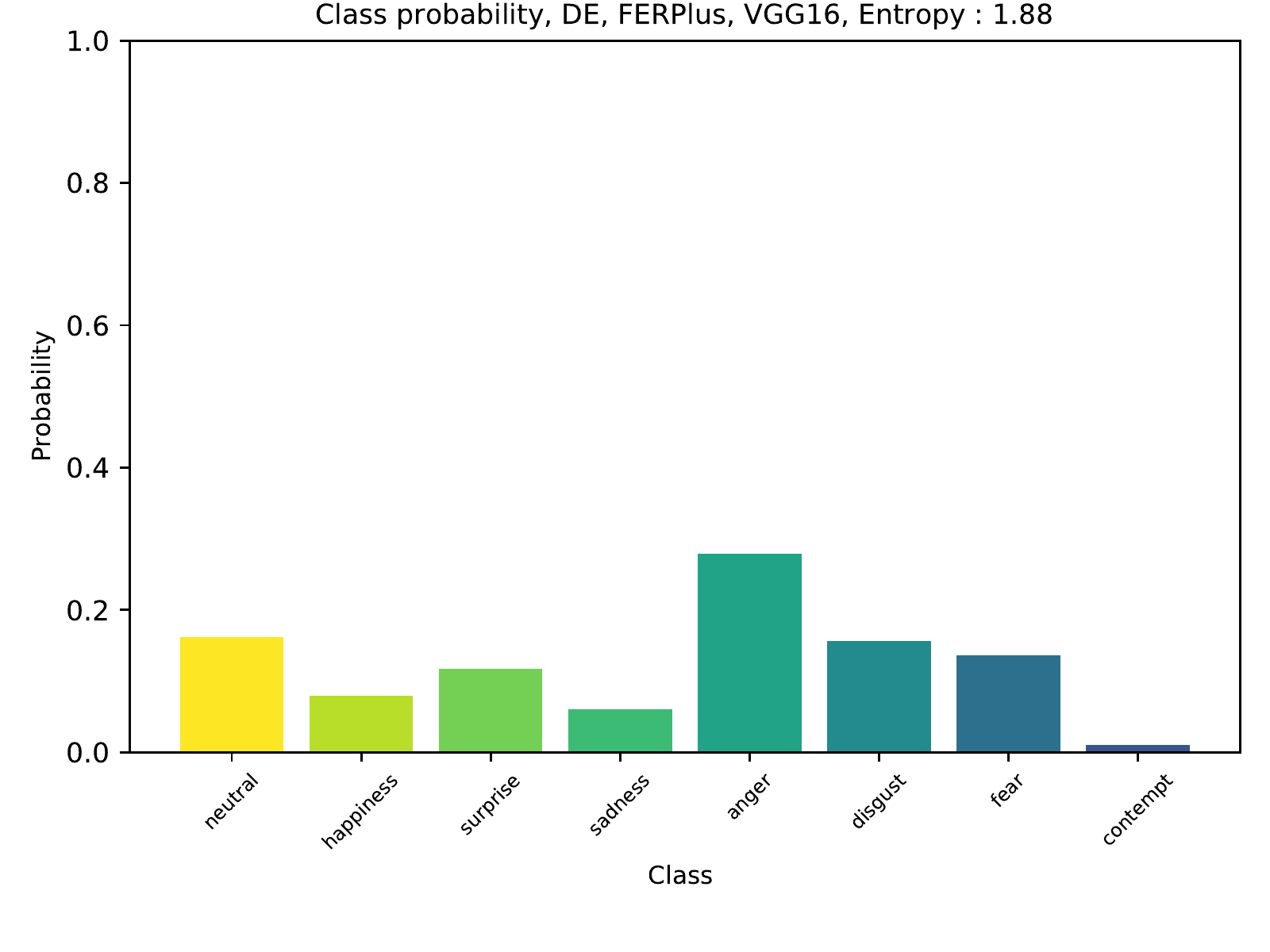}
        \caption*{Anger}
    \end{subfigure}

    \caption{Five most uncertain images based on VGG model and Deep Ensembles with \# of ensembles and a plot of predictive probabilities using 1, 5, 10 and 15 ensembles. The first column represents the image, and the second its ground truth label distribution. Under each probability plot, the predicted class is presented.}       
    \label{fig:ferplus_DE_VGG_probs}
\end{figure}

\begin{figure}[h!]
    \centering
    \begin{subfigure}[b]{0.09\linewidth}
        \caption*{\textbf{Image}}
    \end{subfigure}
    \begin{subfigure}[b]{0.16\linewidth}
        \caption*{\textbf{Labels}}
    \end{subfigure}
    \begin{subfigure}[b]{0.16\linewidth}	
        \caption*{\textbf{1 Ens}}
    \end{subfigure}
    \begin{subfigure}[b]{0.16\linewidth}
        \caption*{\textbf{5 Ens}}
    \end{subfigure}
    \begin{subfigure}[b]{0.16\linewidth}
        \caption*{\textbf{10 Ens}}
    \end{subfigure}
    \begin{subfigure}[b]{0.16\linewidth}
        \caption*{\textbf{15 Ens}}
    \end{subfigure}
    \begin{subfigure}[b]{0.09\linewidth}
        \includegraphics[width=\linewidth]{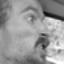}
        \vspace*{0.1em}
        \caption*{Anger}
    \end{subfigure}
    \begin{subfigure}[b]{0.17\linewidth}
        \includegraphics[width=\linewidth]{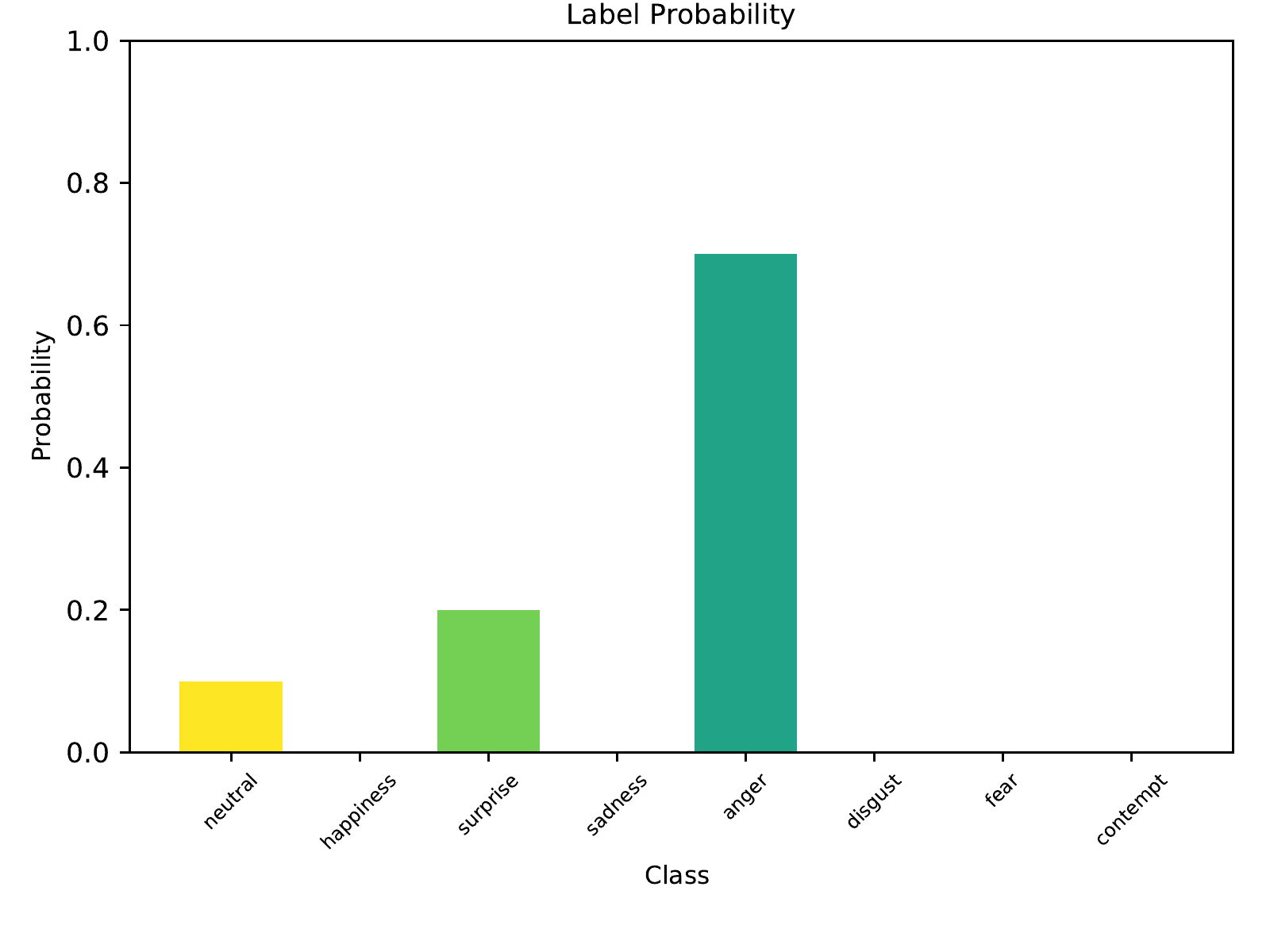}
        \caption*{True Label}
    \end{subfigure}
    \begin{subfigure}[b]{0.17\linewidth}
        \includegraphics[width=\linewidth]{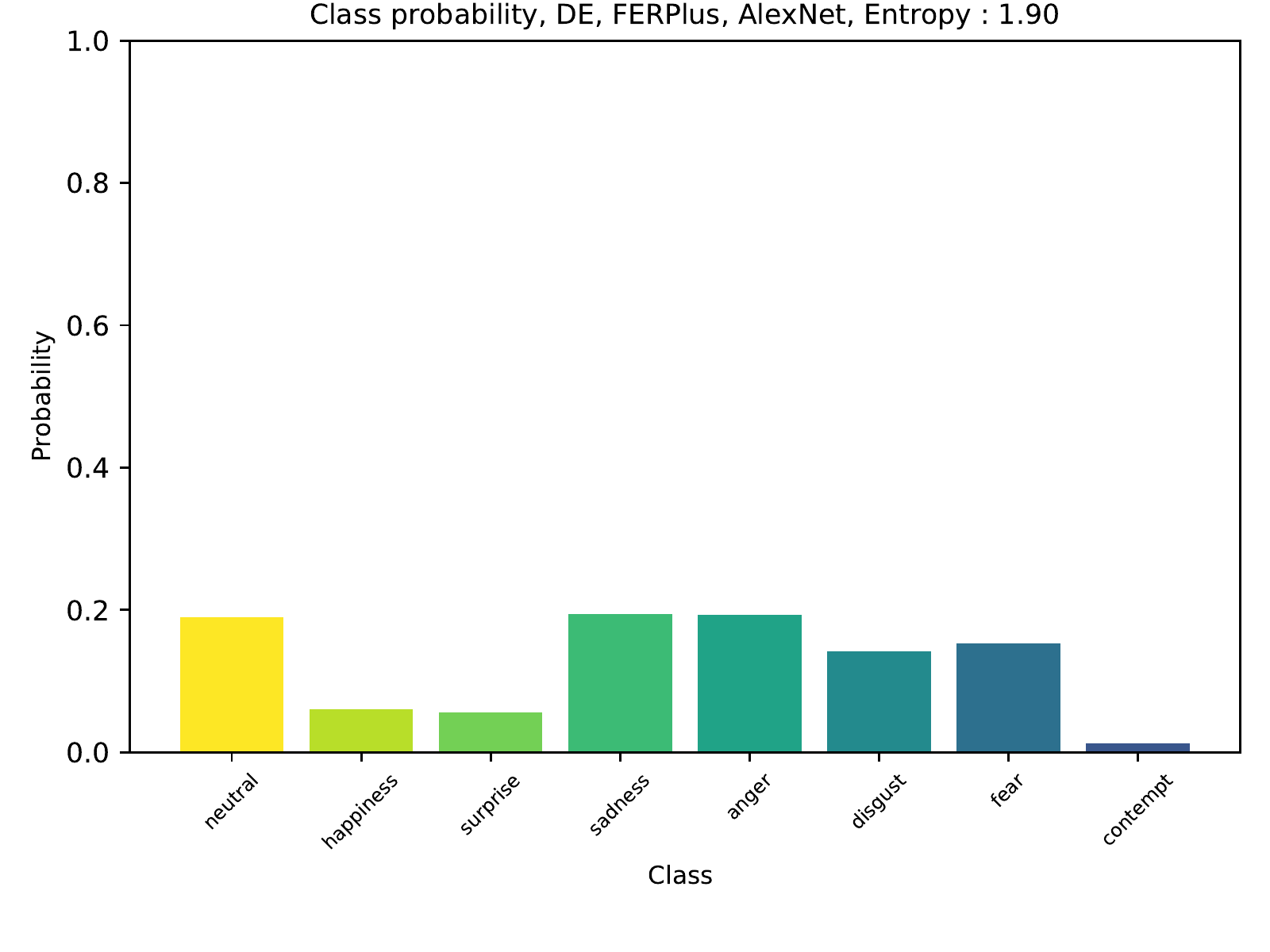}
        \caption*{Sadness}
    \end{subfigure}
    \begin{subfigure}[b]{0.17\linewidth}
        \includegraphics[width=\linewidth]{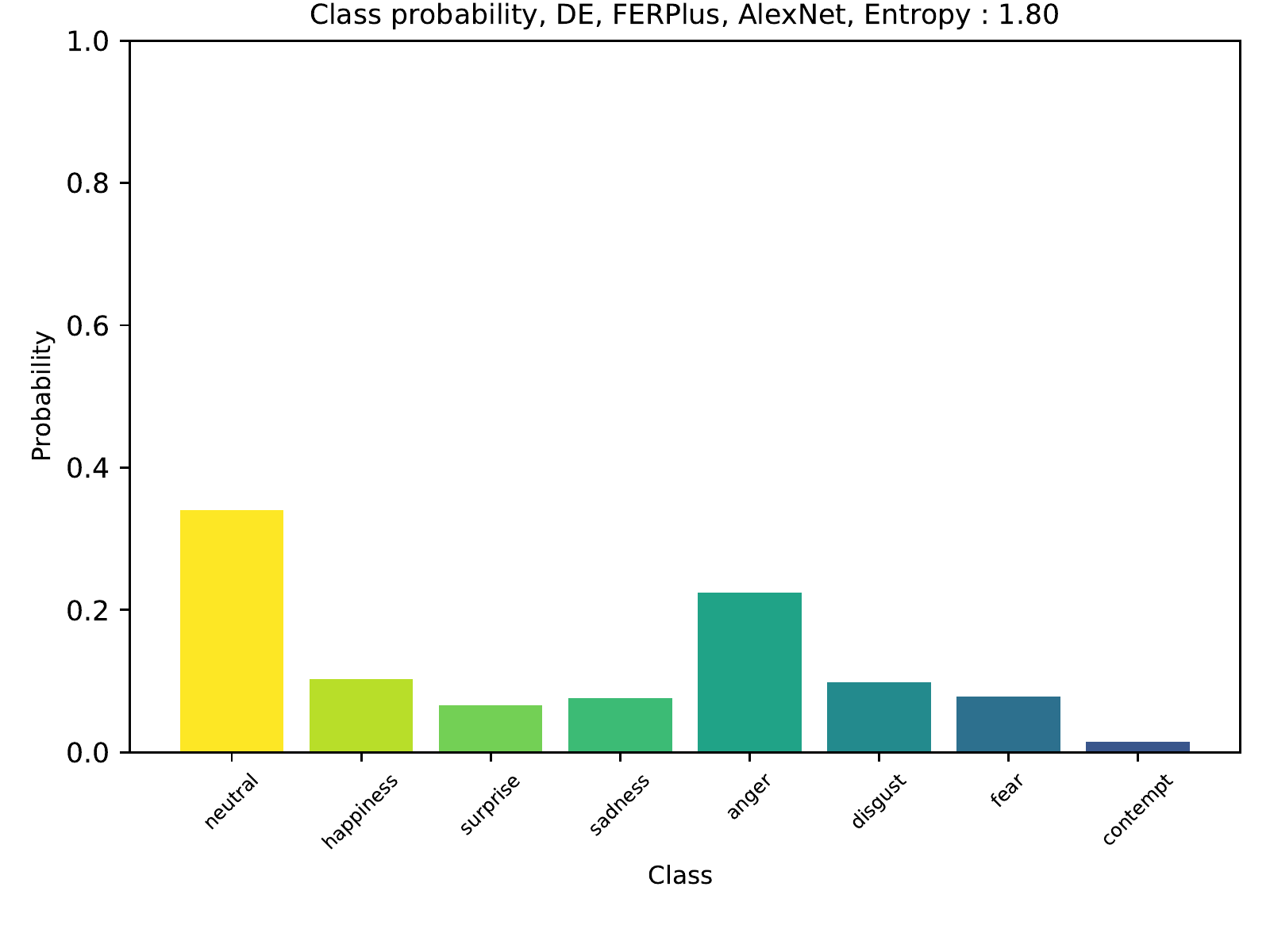}
        \caption*{Neutral}
    \end{subfigure}
    \begin{subfigure}[b]{0.17\linewidth}
        \includegraphics[width=\linewidth]{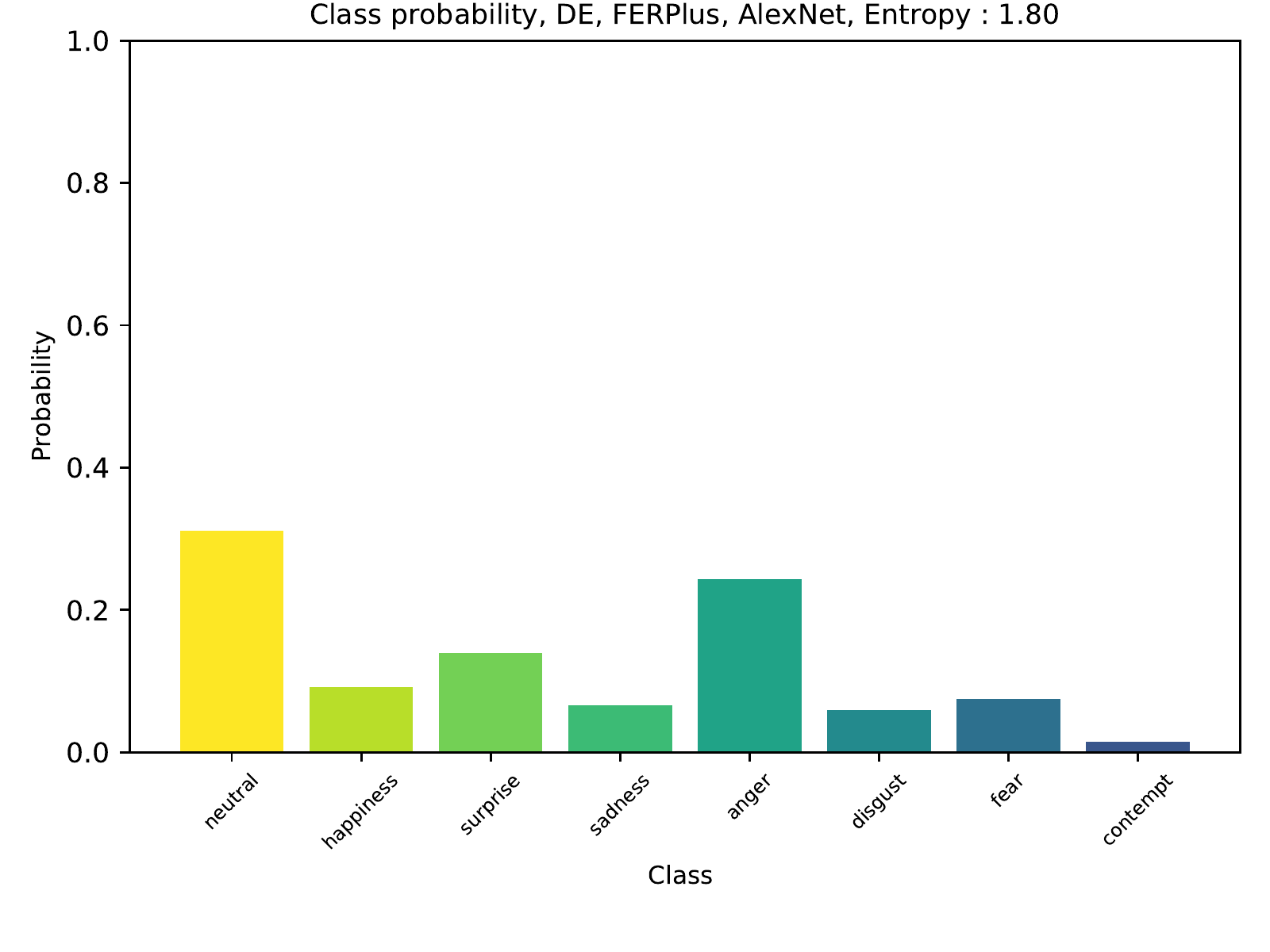}
        \caption*{Neutral}
    \end{subfigure}
    \begin{subfigure}[b]{0.17\linewidth}
        \includegraphics[width=\linewidth]{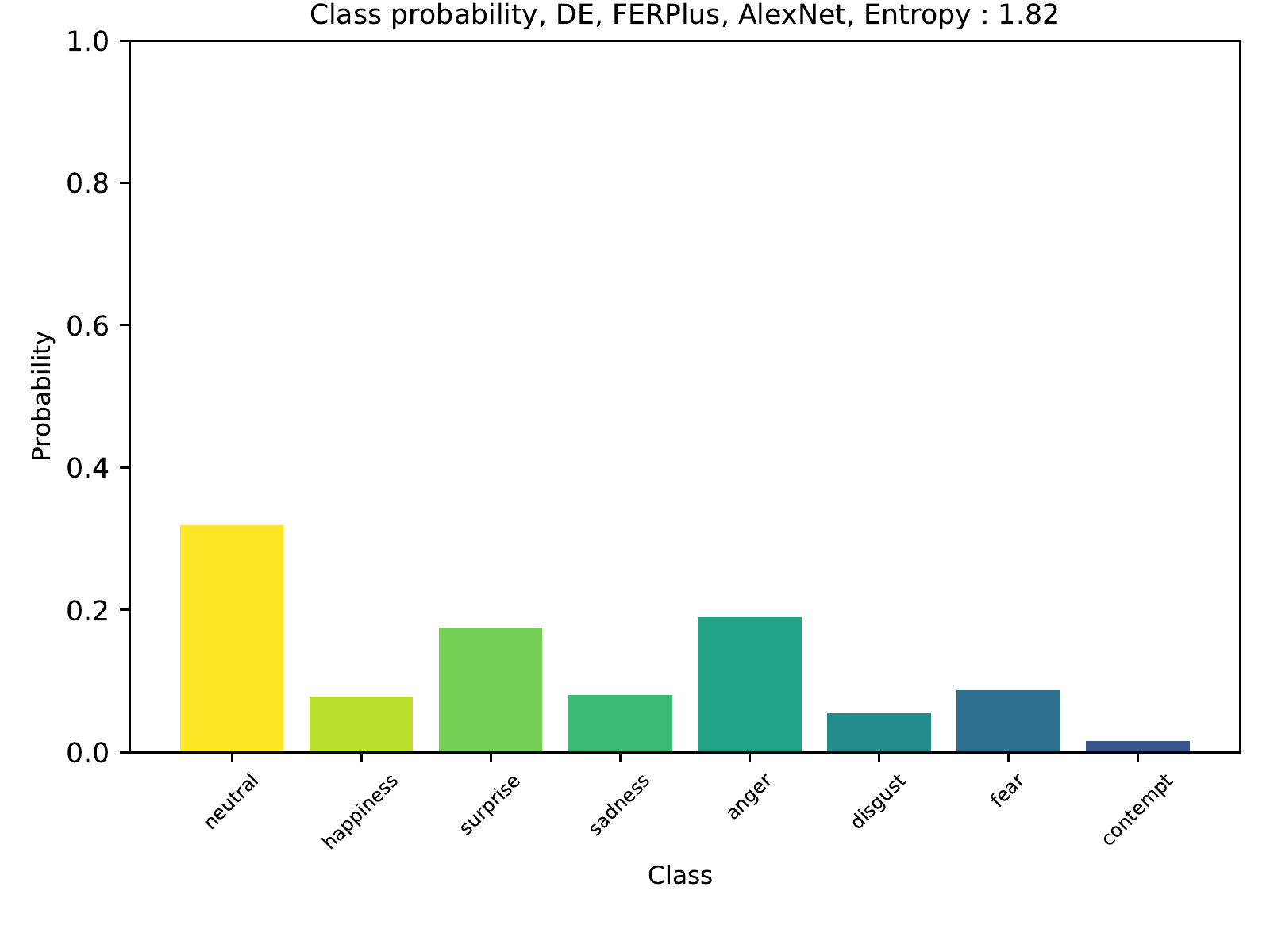}
        \caption*{Neutral}
    \end{subfigure}
    \begin{subfigure}[b]{0.09\linewidth}
        \includegraphics[width=\linewidth]{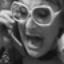}
        \caption*{Surprise}
    \end{subfigure}
    \begin{subfigure}[b]{0.17\linewidth}
        \includegraphics[width=\linewidth]{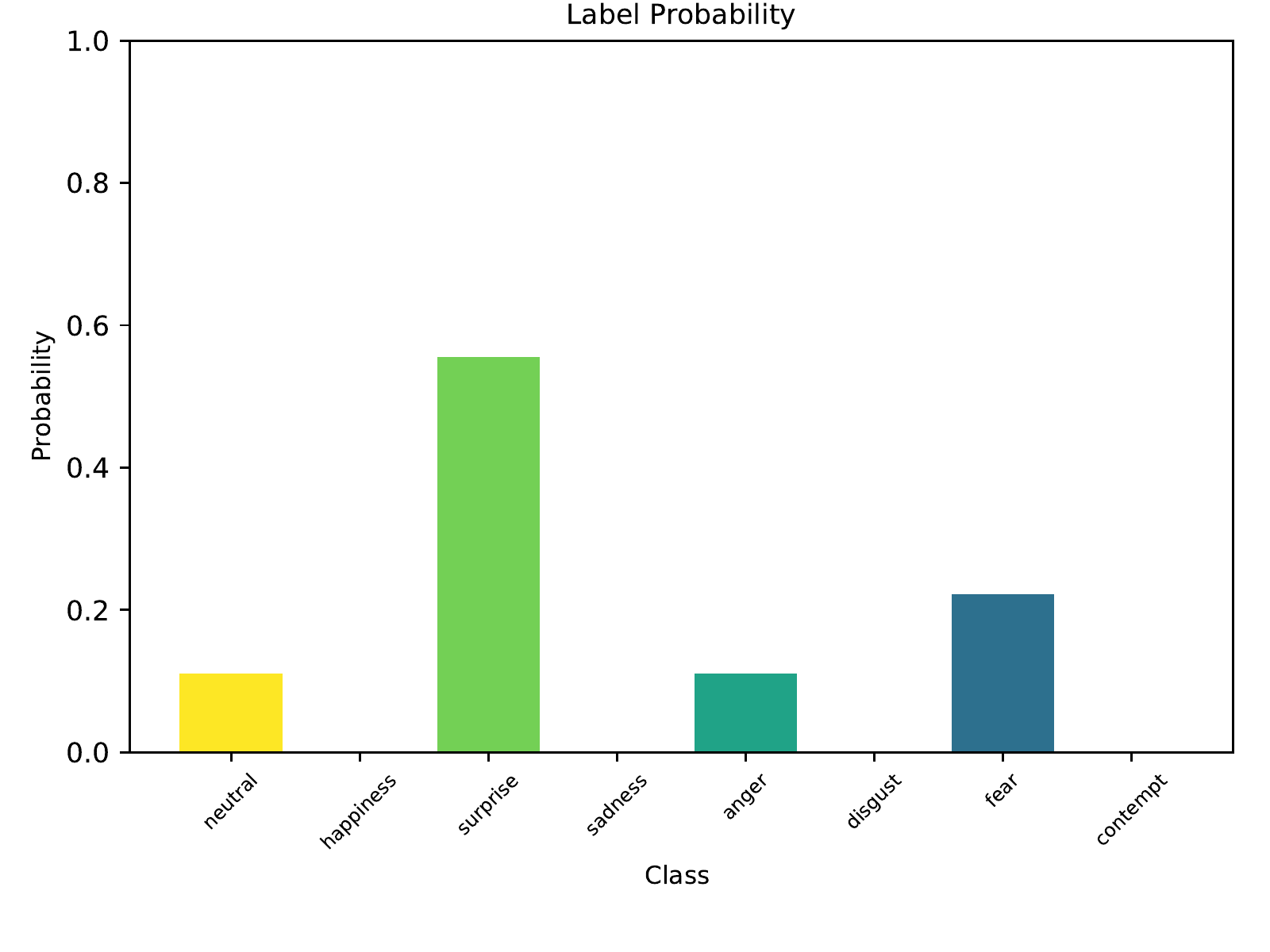}
        \caption*{True Label}
    \end{subfigure}
    \begin{subfigure}[b]{0.17\linewidth}
        \includegraphics[width=\linewidth]{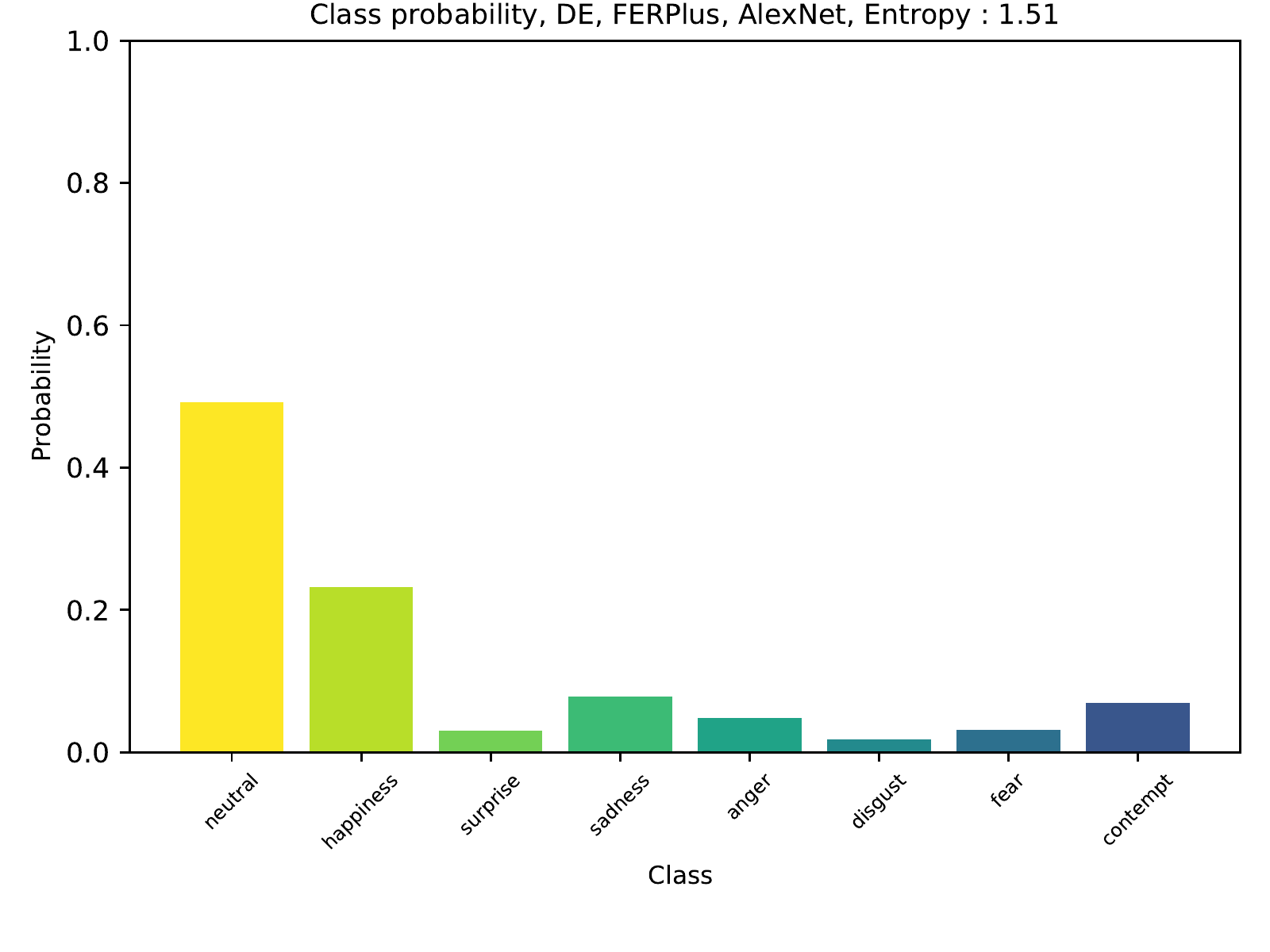}
        \caption*{Neutral}
    \end{subfigure}
    \begin{subfigure}[b]{0.17\linewidth}
        \includegraphics[width=\linewidth]{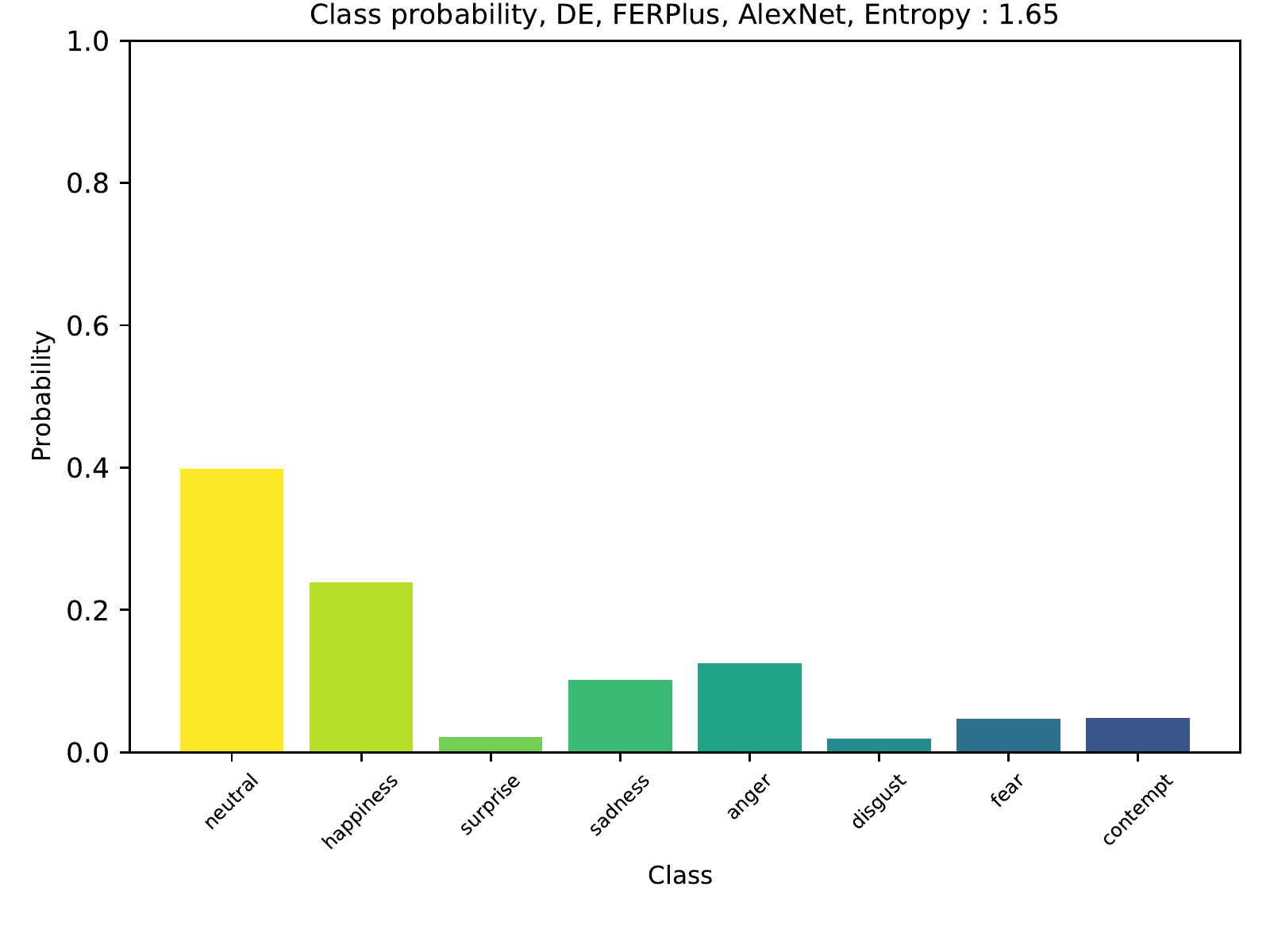}
        \caption*{Neutral}
    \end{subfigure}
    \begin{subfigure}[b]{0.17\linewidth}
        \includegraphics[width=\linewidth]{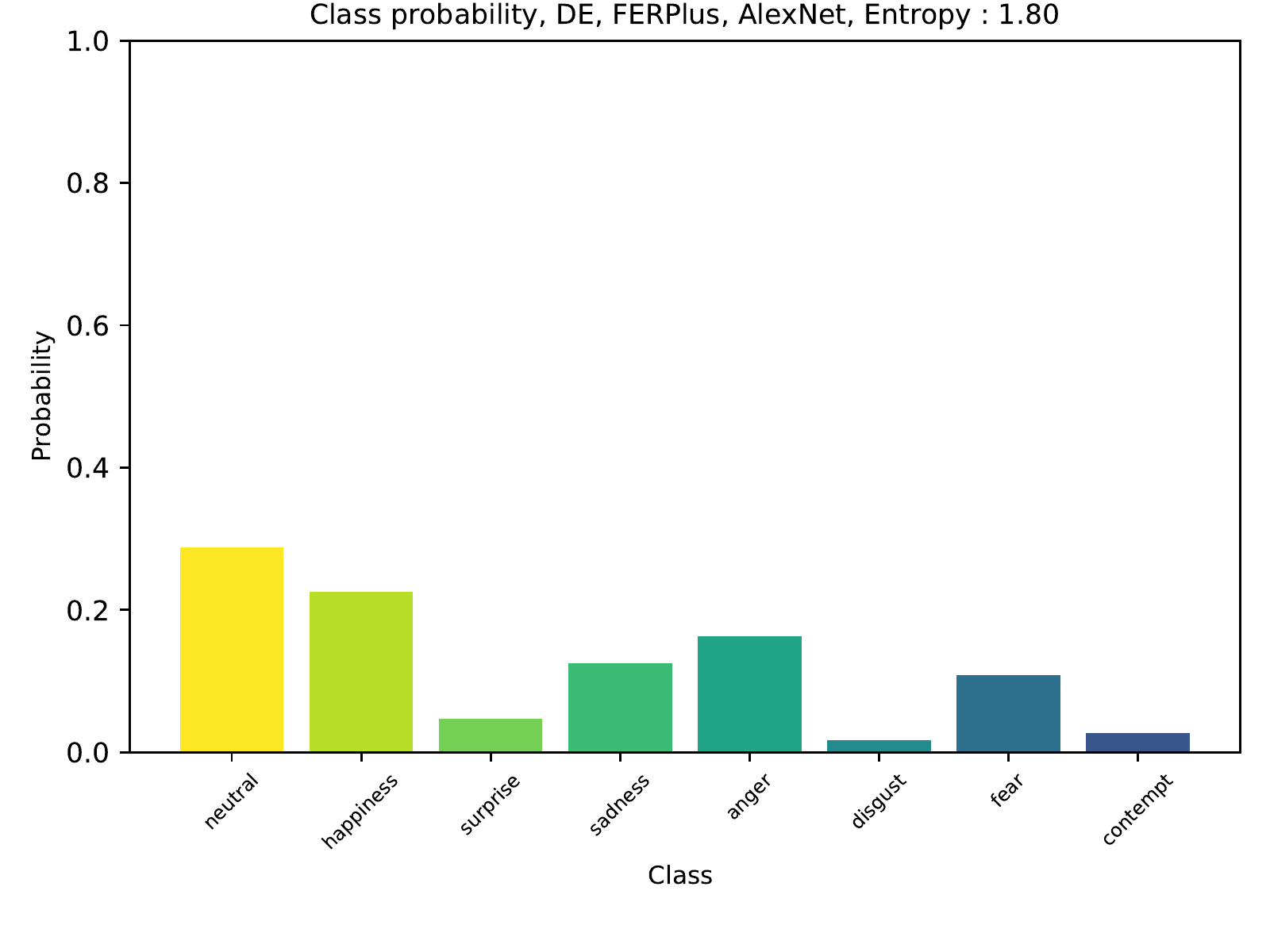}
        \caption*{Neutral}
    \end{subfigure}
    \begin{subfigure}[b]{0.17\linewidth}
        \includegraphics[width=\linewidth]{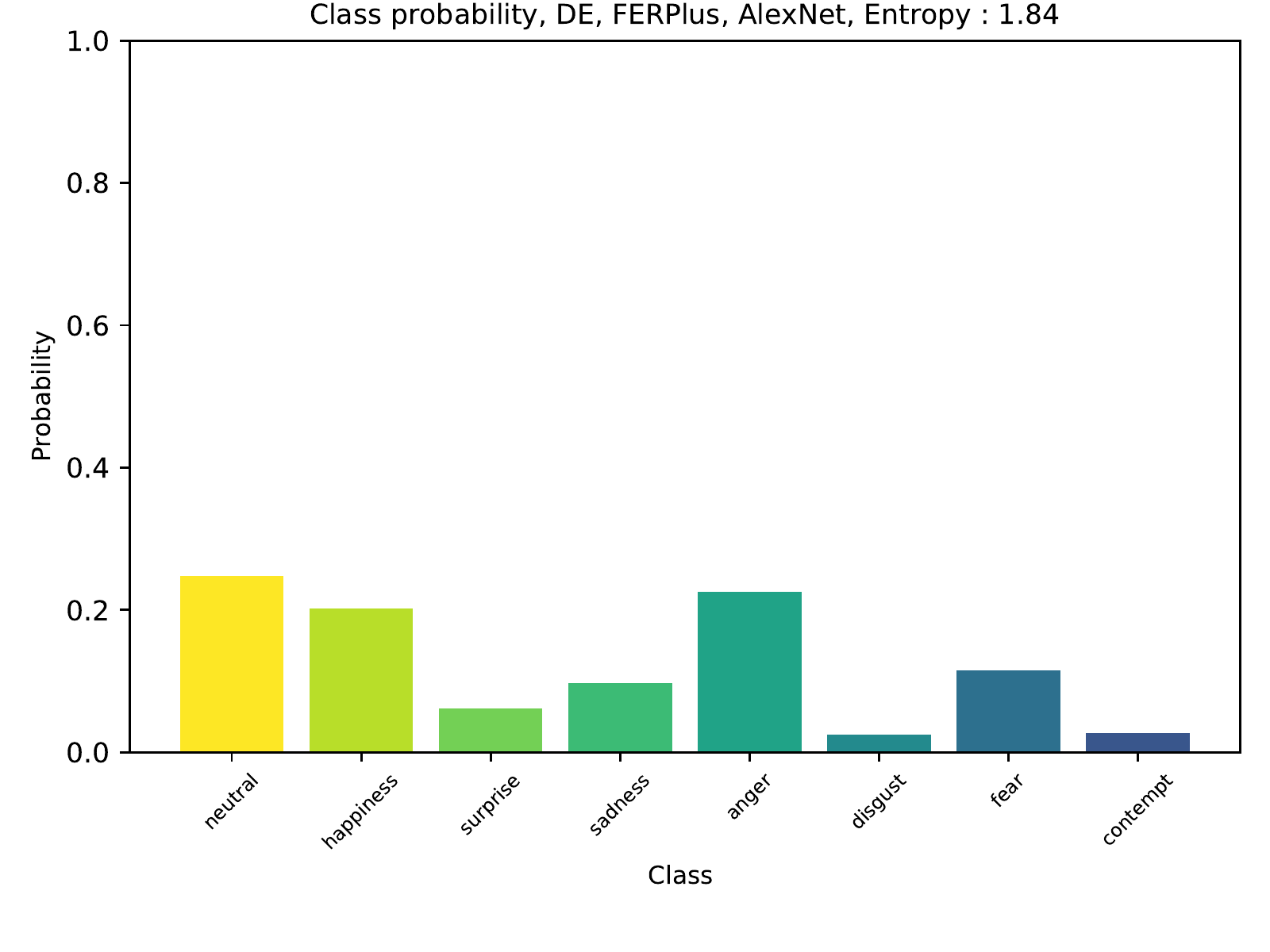}
        \caption*{Neutral}
    \end{subfigure}
    
    \begin{subfigure}[b]{0.09\linewidth}
        \includegraphics[width=\linewidth]{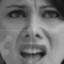}
        \vspace*{0.1em}
        \caption*{Fear}
    \end{subfigure}
    \begin{subfigure}[b]{0.17\linewidth}
        \includegraphics[width=\linewidth]{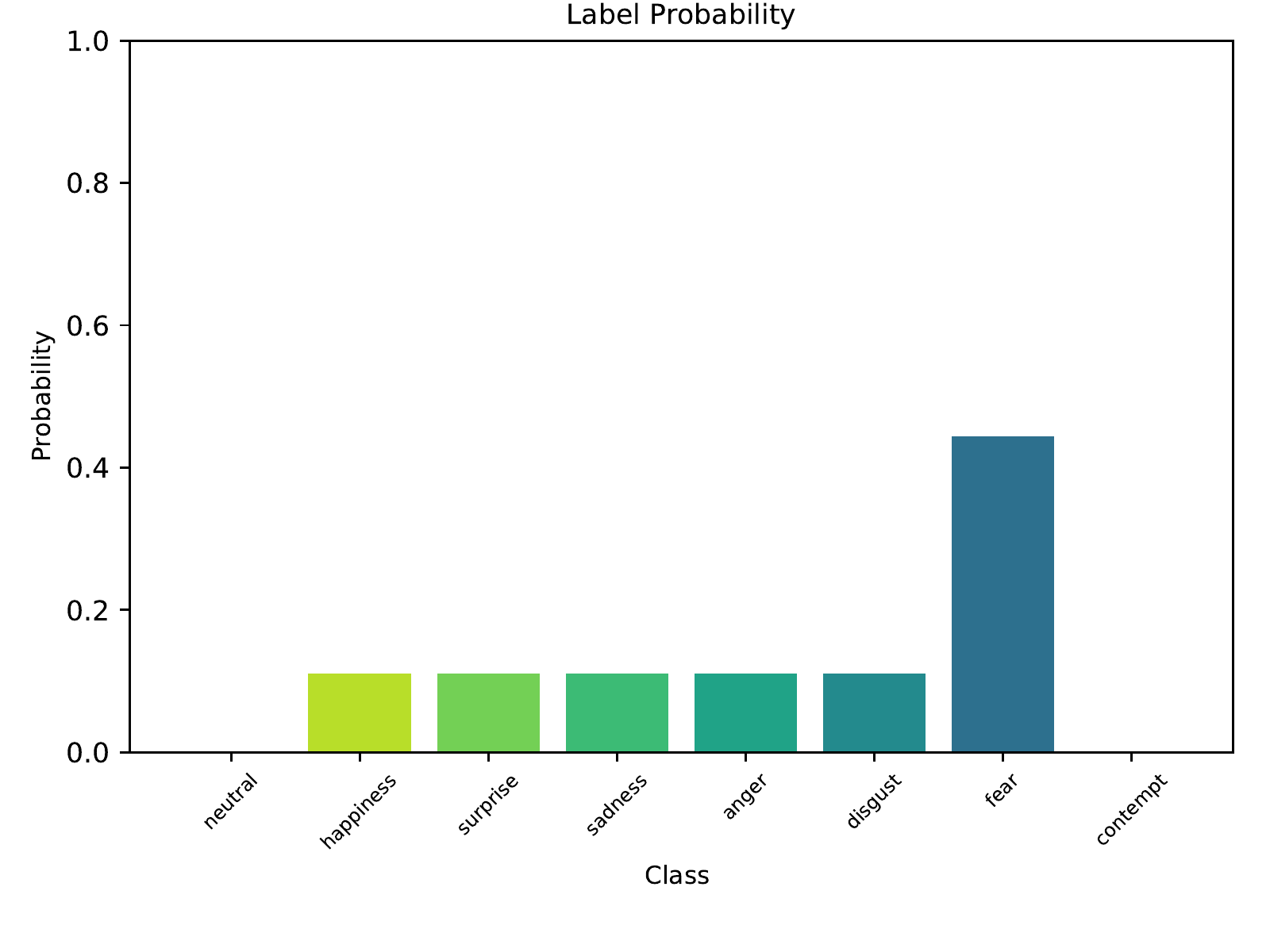}
        \caption*{True Label}
    \end{subfigure}
    \begin{subfigure}[b]{0.17\linewidth}
        \includegraphics[width=\linewidth]{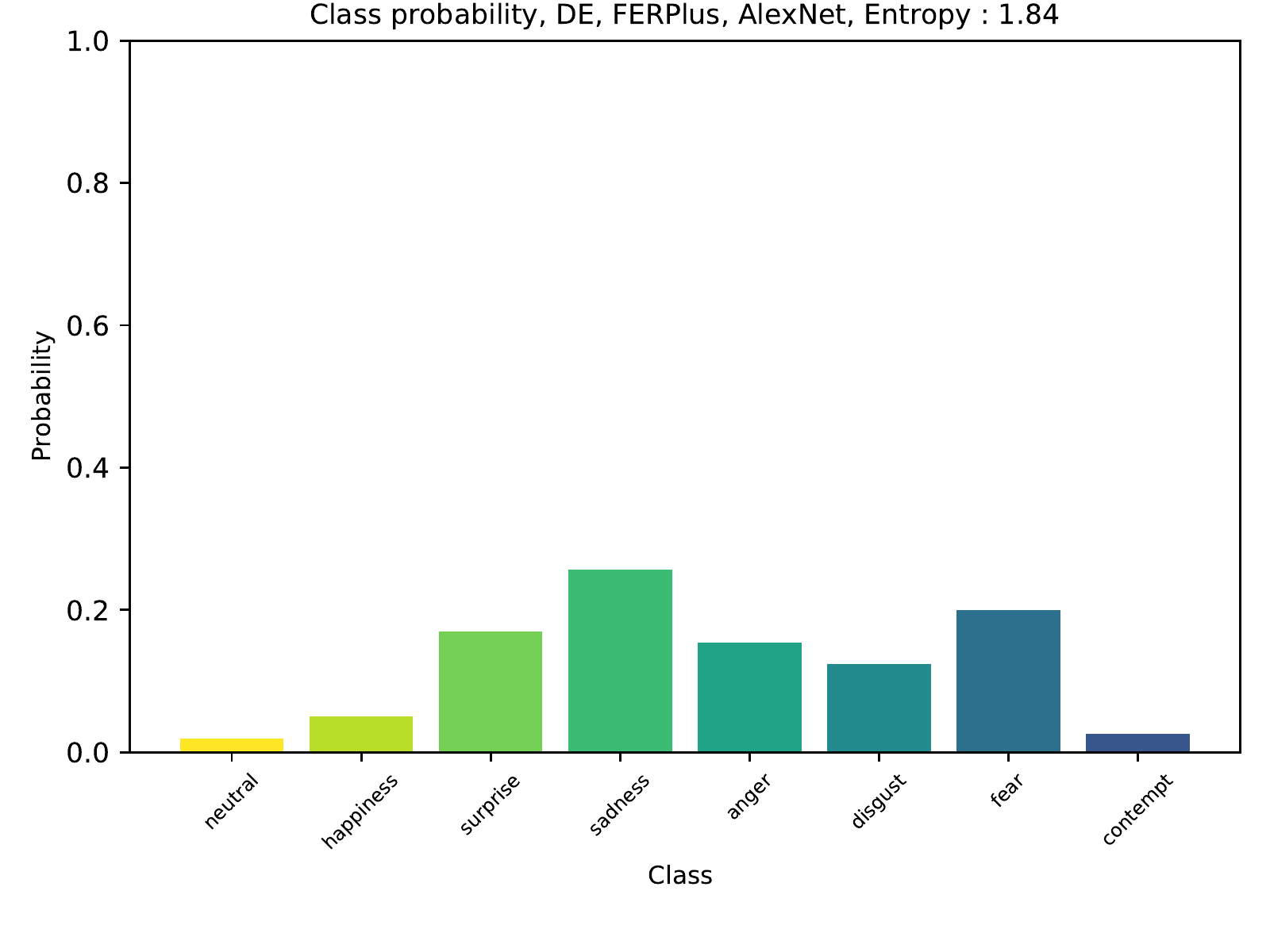}
        \caption*{Sadness}
    \end{subfigure}
    \begin{subfigure}[b]{0.17\linewidth}
        \includegraphics[width=\linewidth]{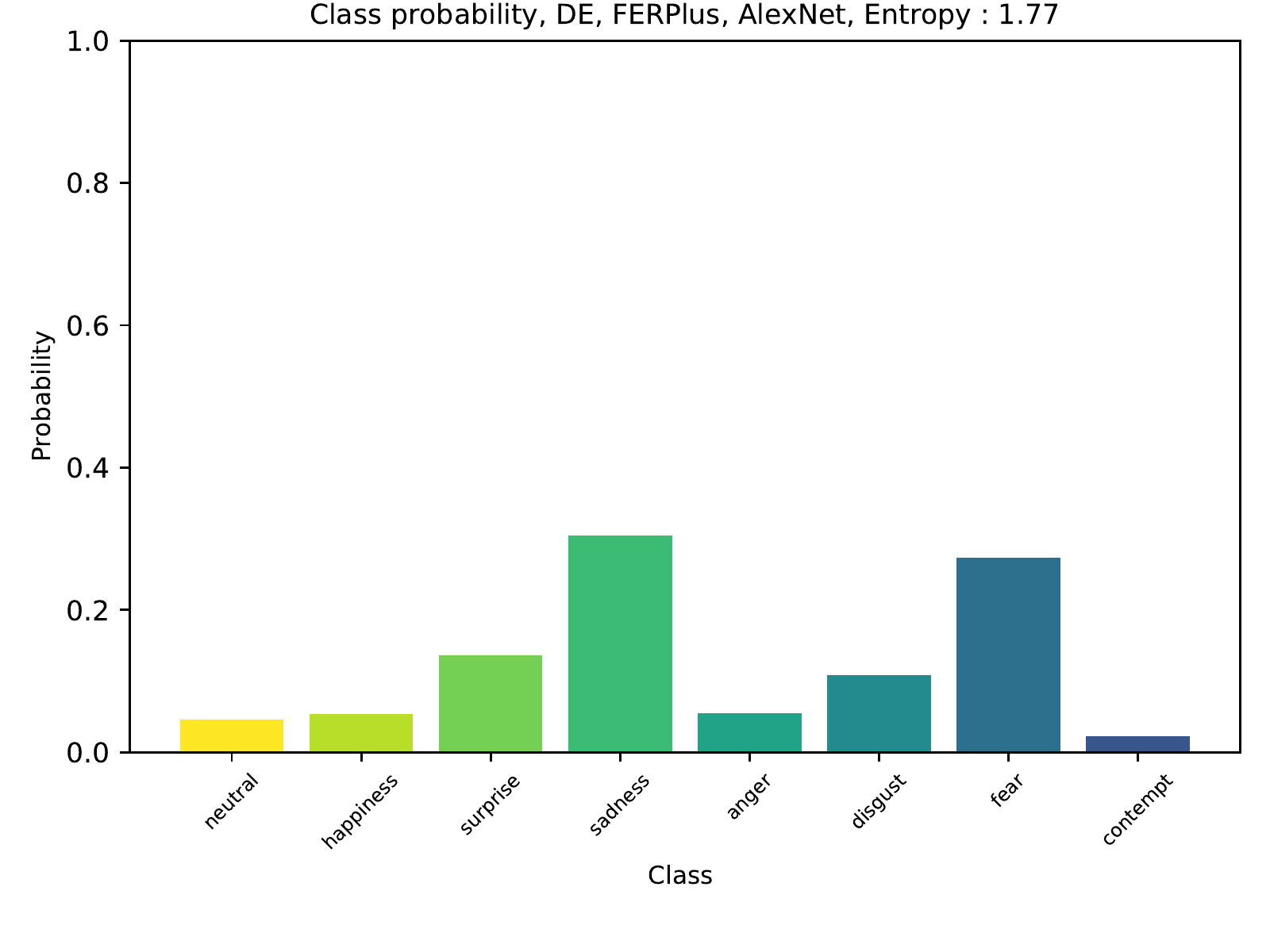}
        \caption*{Sadness}
    \end{subfigure}
    \begin{subfigure}[b]{0.17\linewidth}
        \includegraphics[width=\linewidth]{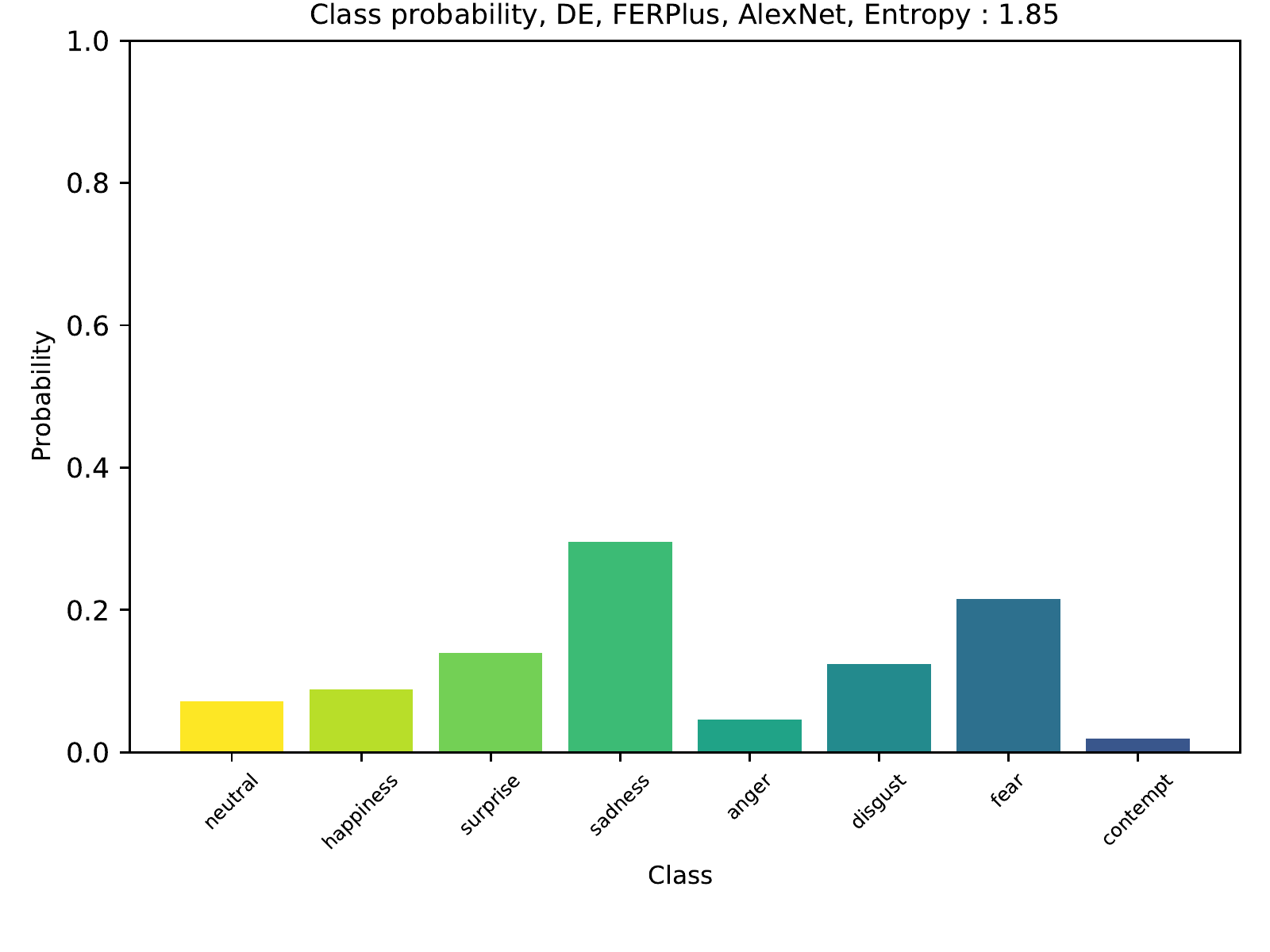}
        \caption*{Sadness}
    \end{subfigure}
    \begin{subfigure}[b]{0.17\linewidth}
        \includegraphics[width=\linewidth]{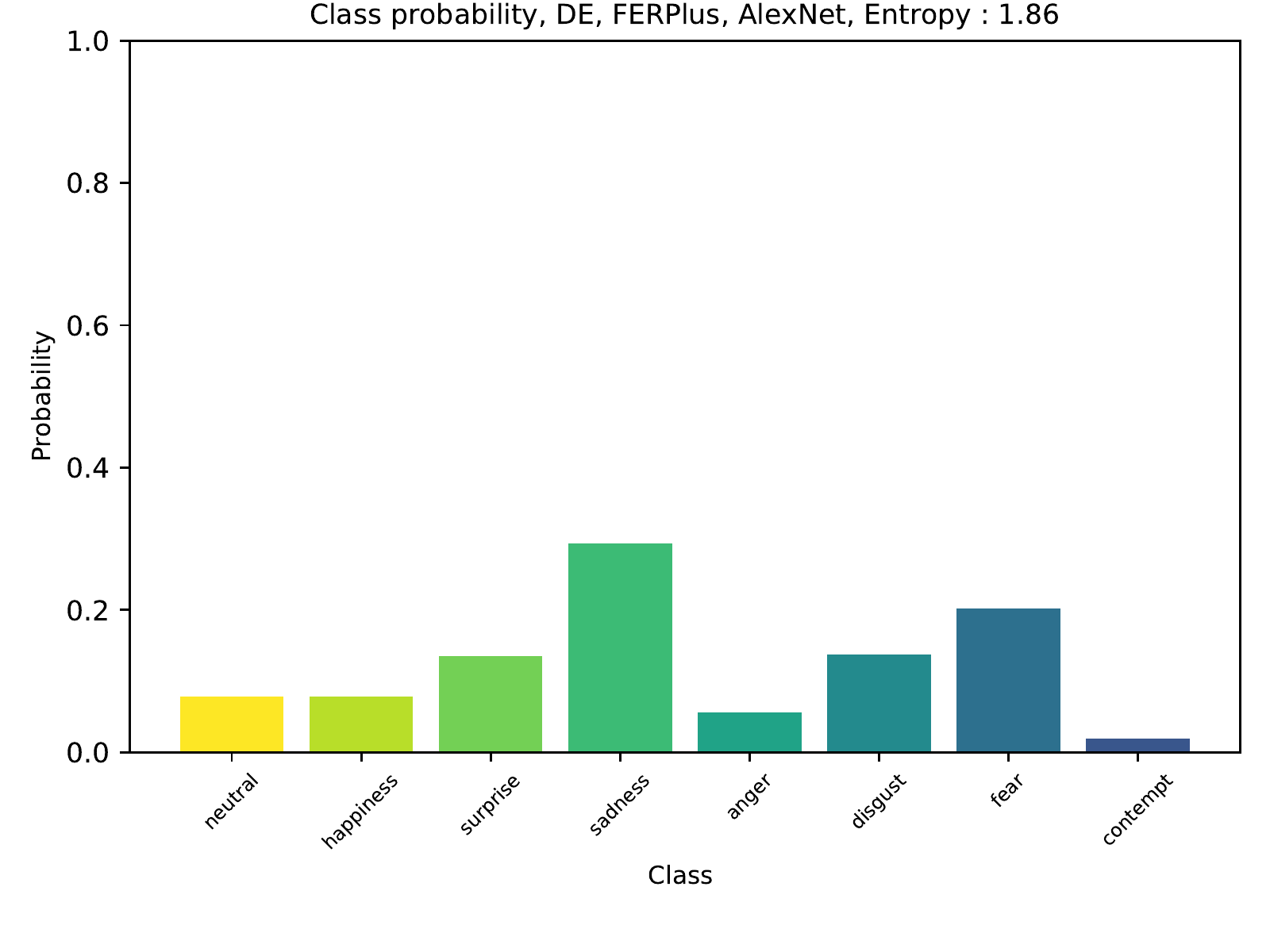}
        \caption*{Sadness}
    \end{subfigure}
    
    \begin{subfigure}[b]{0.09\linewidth}
        \includegraphics[width=\linewidth]{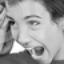}
        \caption*{Happiness}
    \end{subfigure}
    \begin{subfigure}[b]{0.17\linewidth}
        \includegraphics[width=\linewidth]{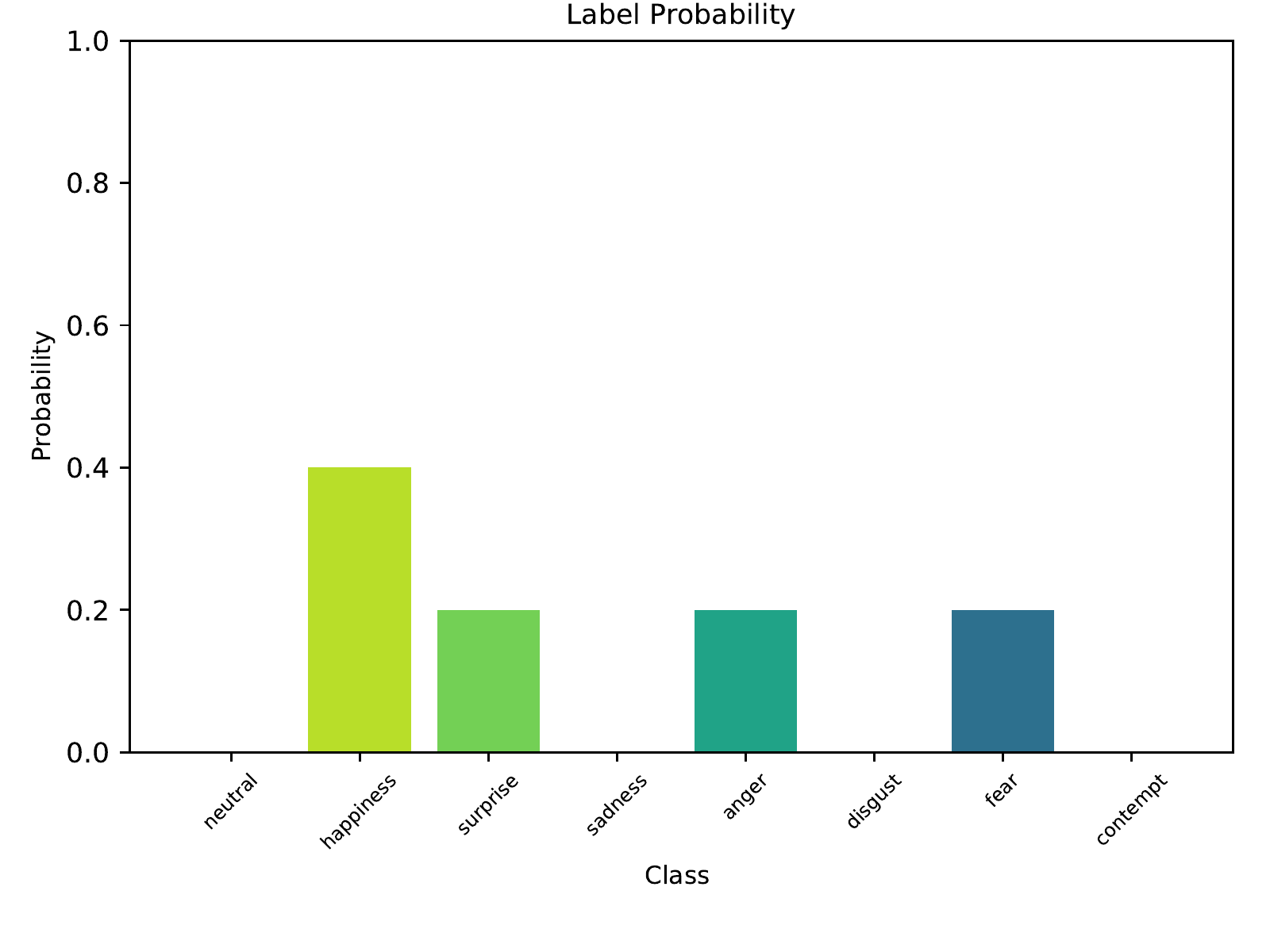}
        \caption*{True Label}
    \end{subfigure}
    \begin{subfigure}[b]{0.17\linewidth}
        \includegraphics[width=\linewidth]{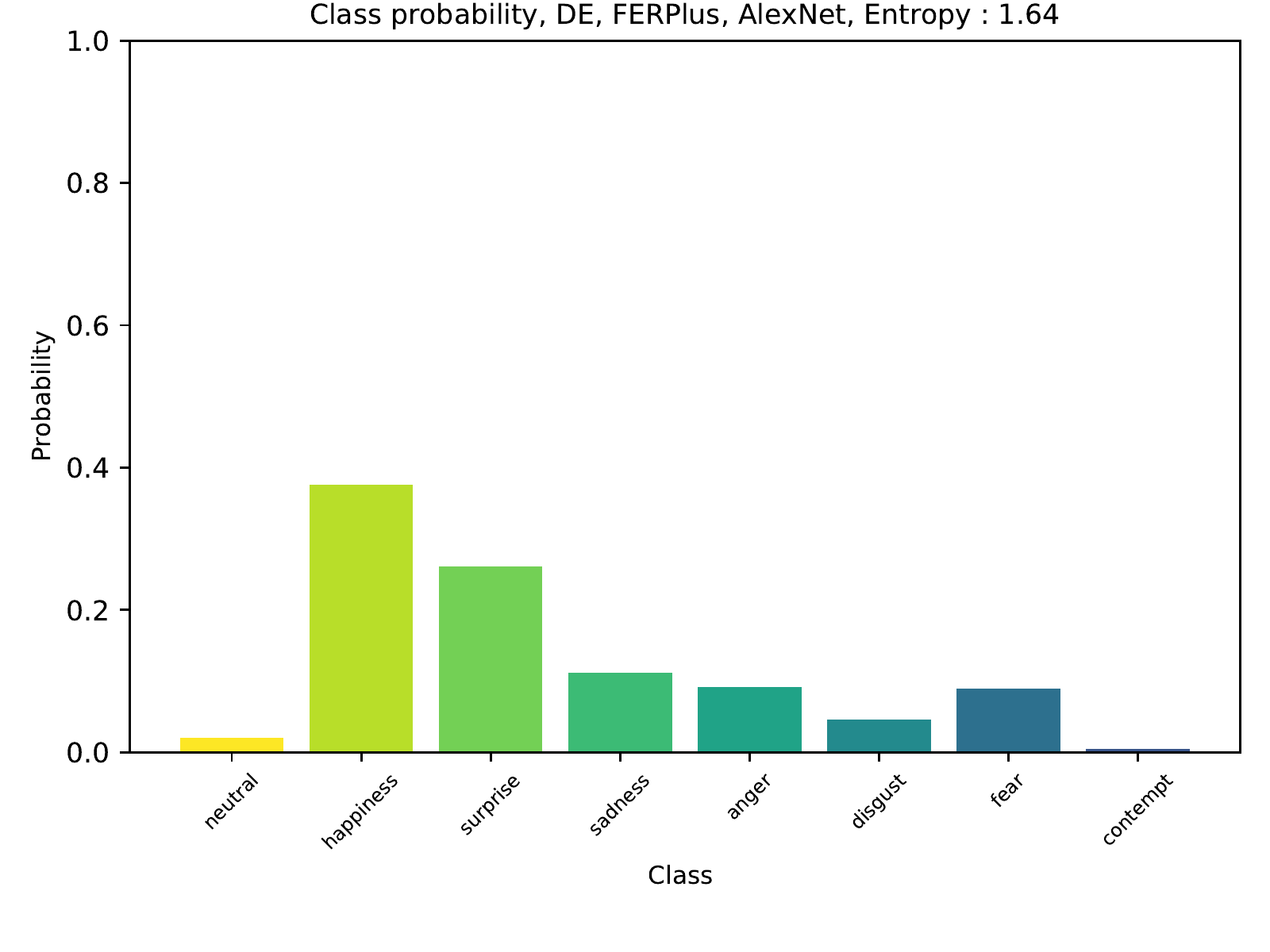}
        \caption*{Happiness}
    \end{subfigure}
    \begin{subfigure}[b]{0.17\linewidth}
        \includegraphics[width=\linewidth]{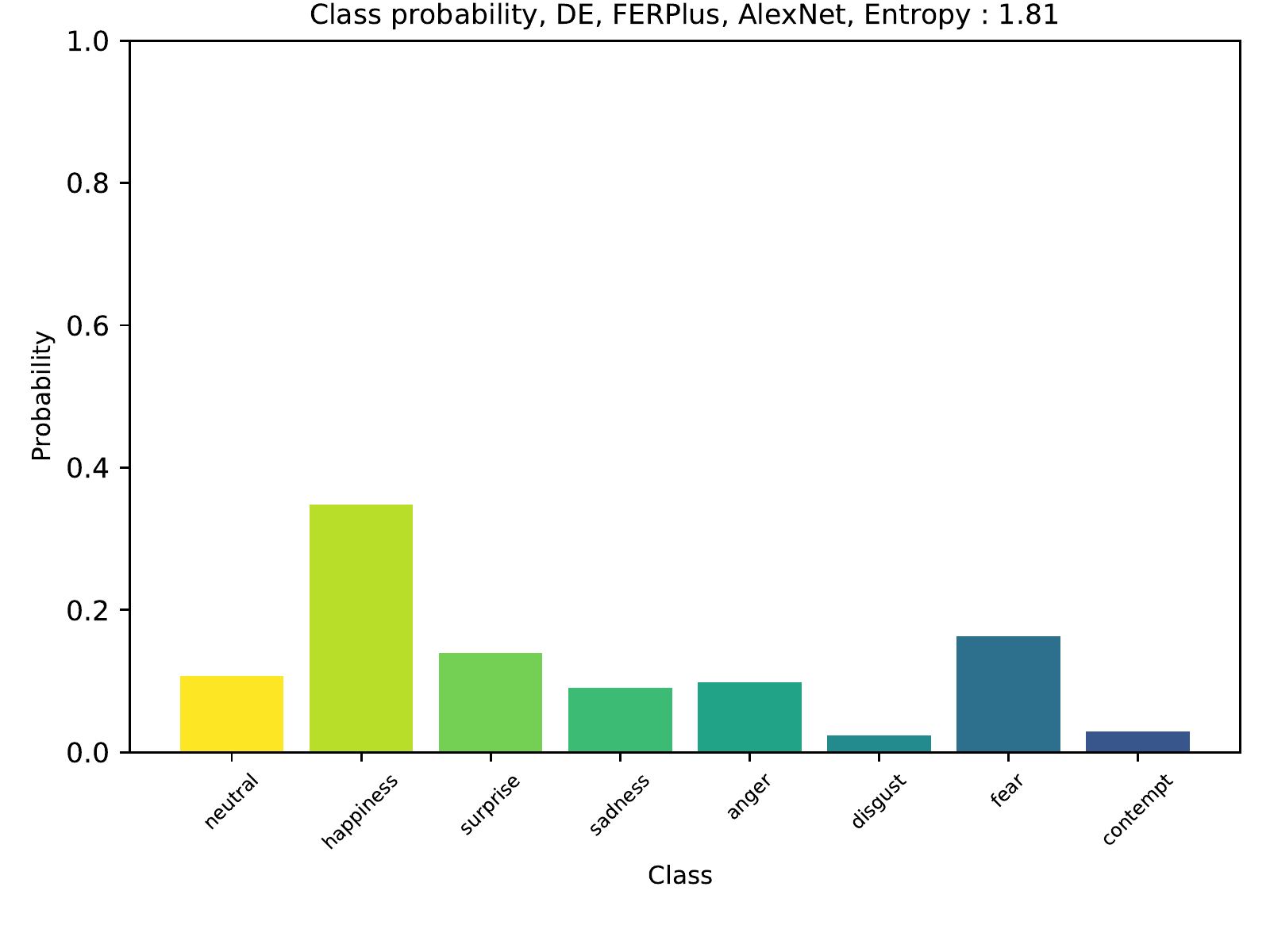}
        \caption*{Happiness}
    \end{subfigure}
    \begin{subfigure}[b]{0.17\linewidth}
        \includegraphics[width=\linewidth]{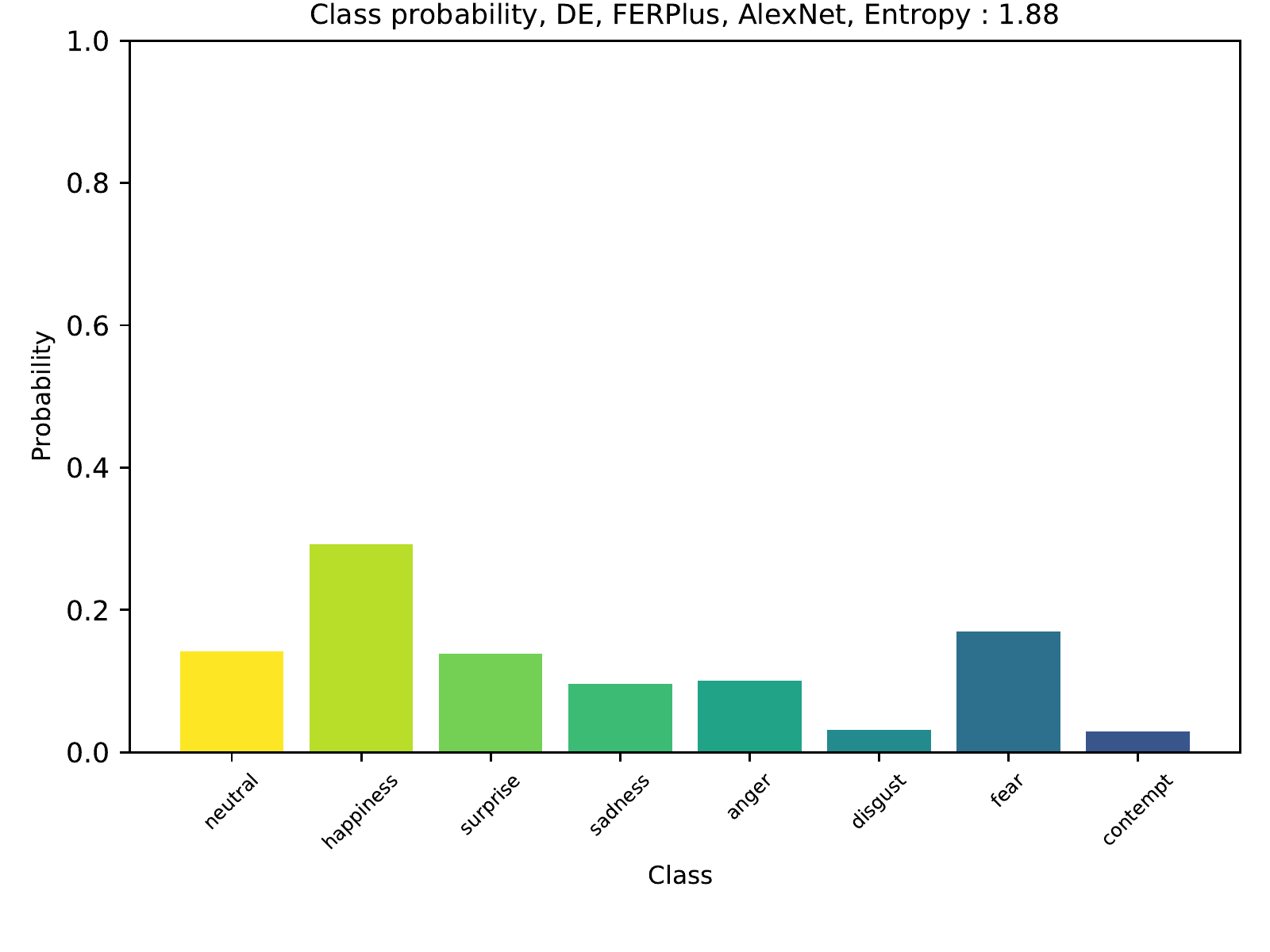}
        \caption*{Happiness}
    \end{subfigure}
    \begin{subfigure}[b]{0.17\linewidth}
        \includegraphics[width=\linewidth]{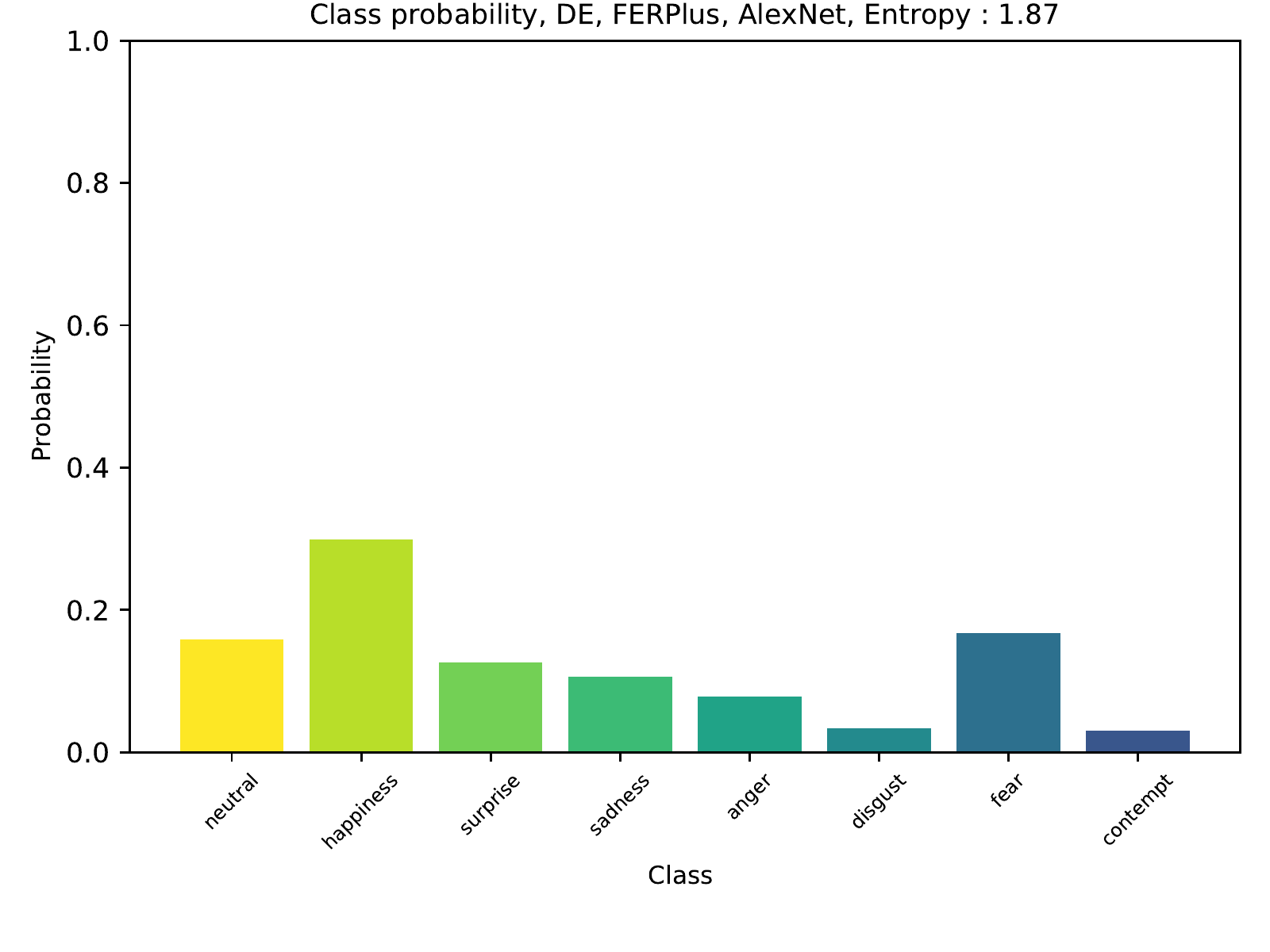}
        \caption*{Happiness}
    \end{subfigure}
    
    \begin{subfigure}[b]{0.09\linewidth}
        \includegraphics[width=\linewidth]{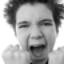}
        \vspace*{0.1em}
        \caption*{Anger}
    \end{subfigure}
    \begin{subfigure}[b]{0.17\linewidth}
        \includegraphics[width=\linewidth]{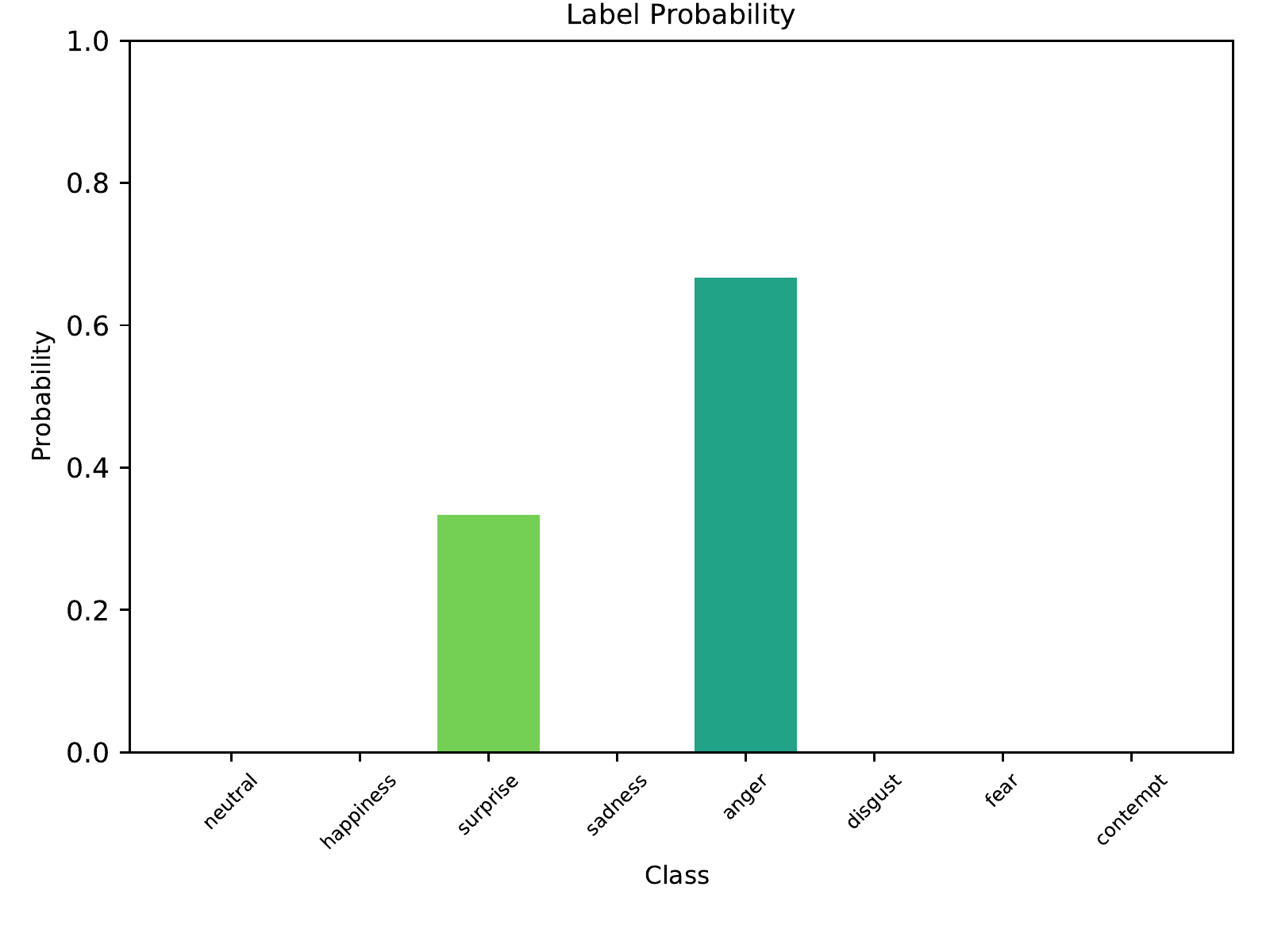}
        \caption*{True Label}
    \end{subfigure}
    \begin{subfigure}[b]{0.17\linewidth}
        \includegraphics[width=\linewidth]{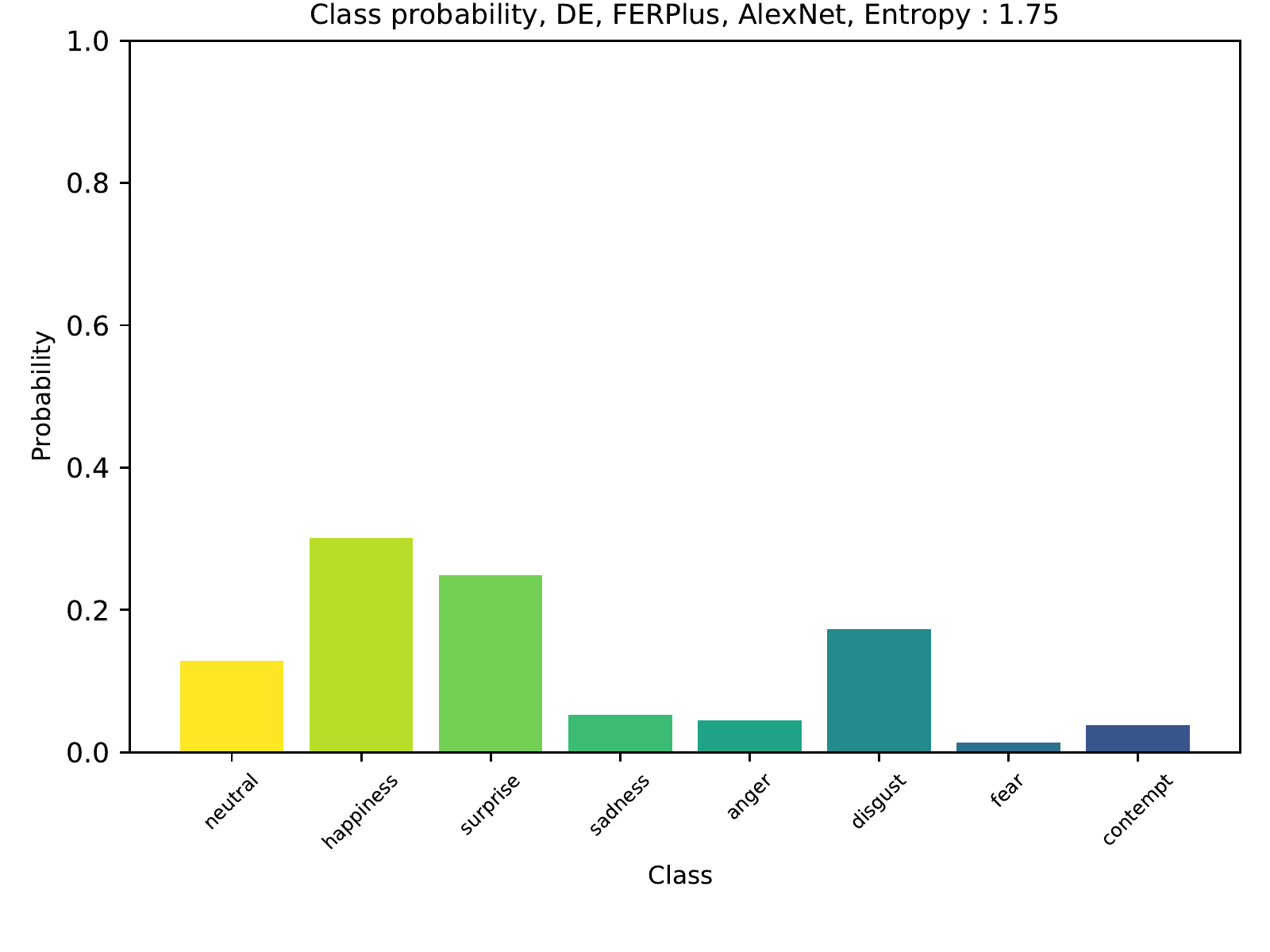}
        \caption*{Happiness}
    \end{subfigure}
    \begin{subfigure}[b]{0.17\linewidth}
        \includegraphics[width=\linewidth]{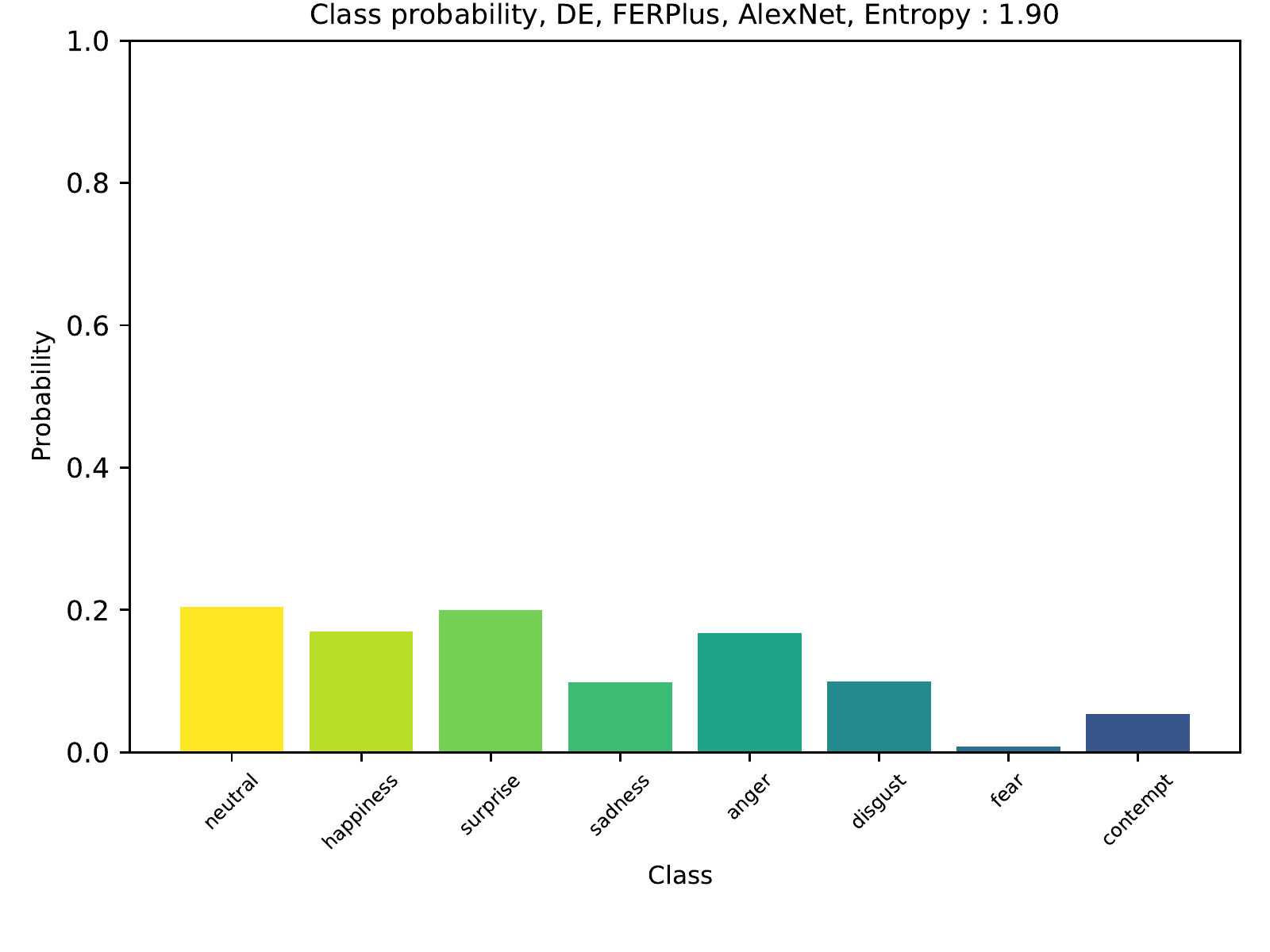}
        \caption*{Neutral}
    \end{subfigure}
    \begin{subfigure}[b]{0.17\linewidth}
        \includegraphics[width=\linewidth]{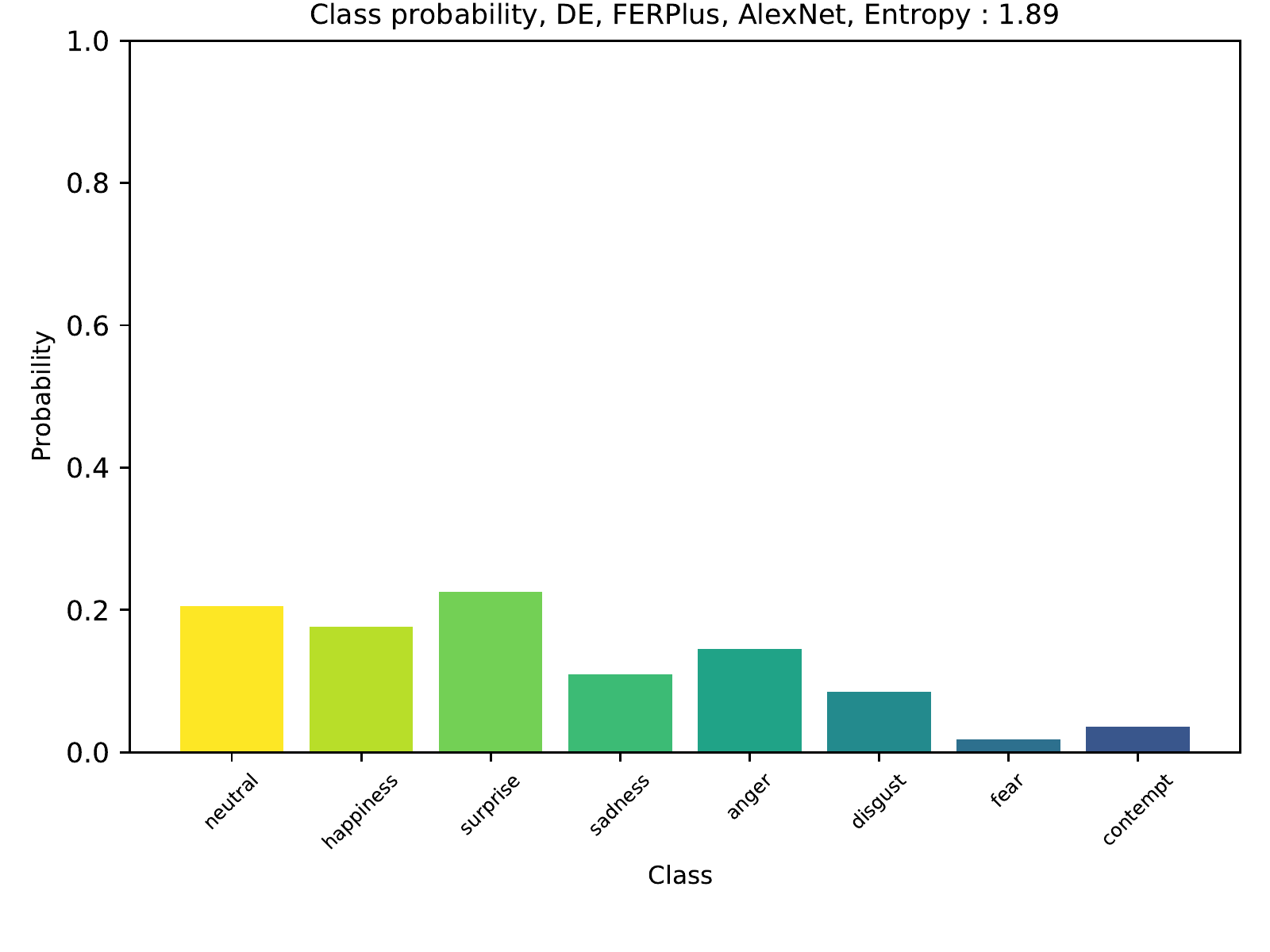}
        \caption*{Surprise}
    \end{subfigure}
    \begin{subfigure}[b]{0.17\linewidth}
        \includegraphics[width=\linewidth]{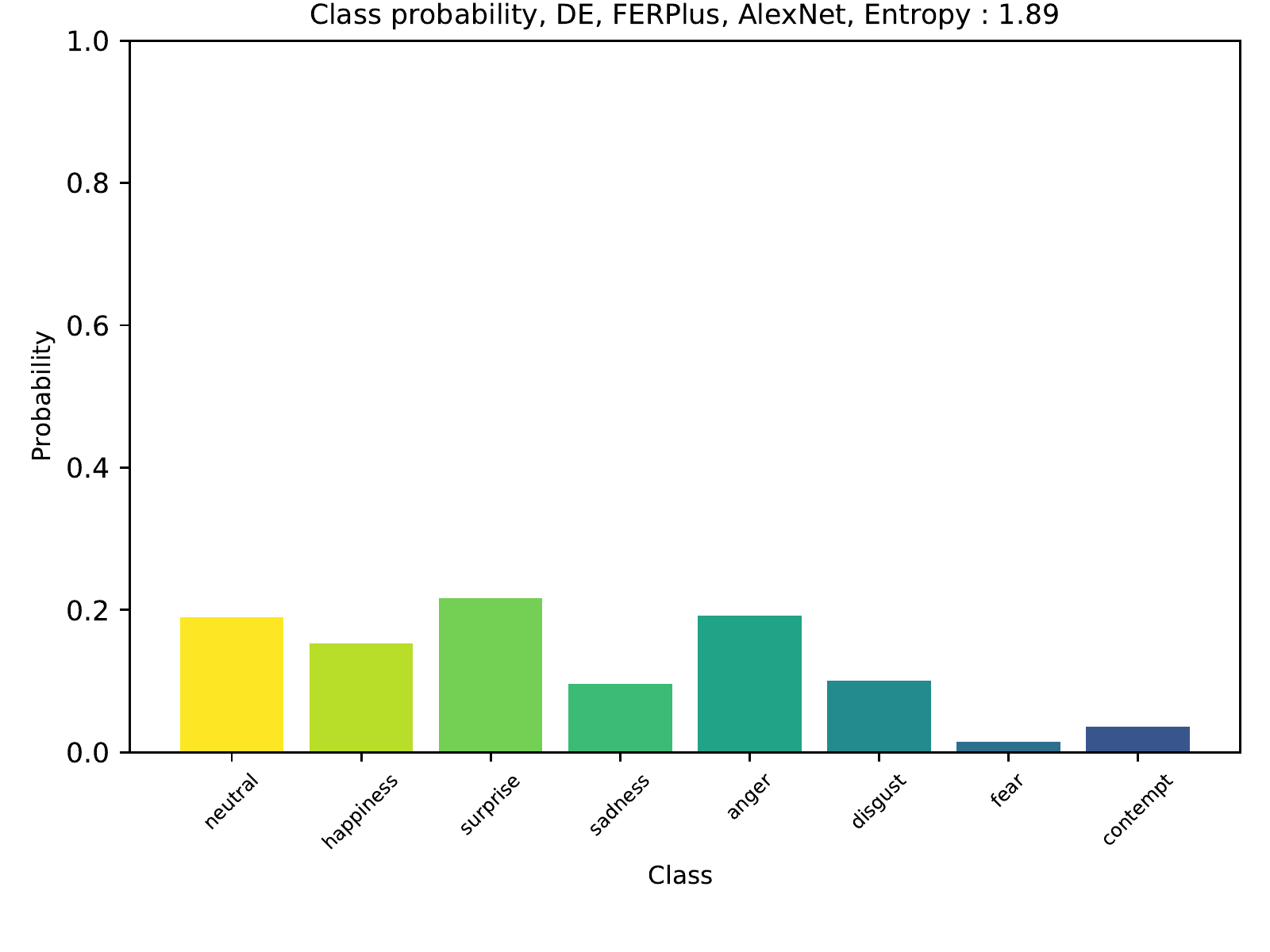}
        \caption*{Surprise}
    \end{subfigure}

    \caption{Five most uncertain images based on AlexNet model and Deep Ensembles with \# of ensembles and a plot of predictive probabilities using 1, 5, 10 and 15 ensembles. The first column represents the image, and the second its ground truth label distribution. Under each probability plot, the predicted class is presented.}
    
    \label{fig:ferplus_DE_Alexnet_probs}
\end{figure}

\end{document}